\documentclass[10pt,twocolumn,letterpaper]{article}

\usepackage[pagenumbers]{cvpr}
\AtEndPreamble{\crefname{appendix}{Sec.}{Secs.}\Crefname{appendix}{Section}{Sections}}

\usepackage[utf8]{inputenc} %
\usepackage[T1]{fontenc}    %
\usepackage[pagebackref]{hyperref}       %
\usepackage{url}            %
\usepackage{booktabs}       %
\usepackage{amsfonts}       %
\usepackage{nicefrac}       %
\usepackage{microtype}      %
\usepackage[dvipsnames,table]{xcolor}
\usepackage{amsmath}
\usepackage{amssymb}
\usepackage{graphicx}
\usepackage{tikz}
\usepackage{pifont}
\usepackage{xspace}       %
\usepackage{tabularx}
\usepackage{array}
\usepackage{placeins}
\usepackage{float}
\usepackage{multirow}

\newcommand{\cmark}{\ding{51}}
\newcommand{\xmark}{\ding{55}}

\definecolor{RankBest}{RGB}{245,145,148}
\definecolor{RankSecond}{RGB}{250,199,145}
\definecolor{RankThird}{RGB}{255,255,143}

\newcommand{\bestcell}[1]{\cellcolor{RankBest}#1}
\newcommand{\secondcell}[1]{\cellcolor{RankSecond}#1}
\newcommand{\thirdcell}[1]{\cellcolor{RankThird}#1}

\title{Topology-Aware 4D Gaussian Splatting for Egocentric Videos}

\author{%
  \textbf{Leonardo Vanni}\textsuperscript{1}
  \qquad
  \textbf{Stefano Gasperini}\textsuperscript{2}
  \\[0.25em]
  \textbf{Federico Tombari}\textsuperscript{2,3}
  \qquad
  \textbf{Chiara Plizzari}\textsuperscript{1}
  \\[0.55em]
  \normalfont
  \textsuperscript{1}Bocconi University
  \\
  \textsuperscript{2}Technical University of Munich
  \\
  \textsuperscript{3}Google
}

\begin{document}

\twocolumn[{%
\renewcommand\twocolumn[1][]{#1}%
\maketitle
\begin{center}
    \centering
    \captionsetup{type=figure}
    \includegraphics[width=1.00\textwidth]{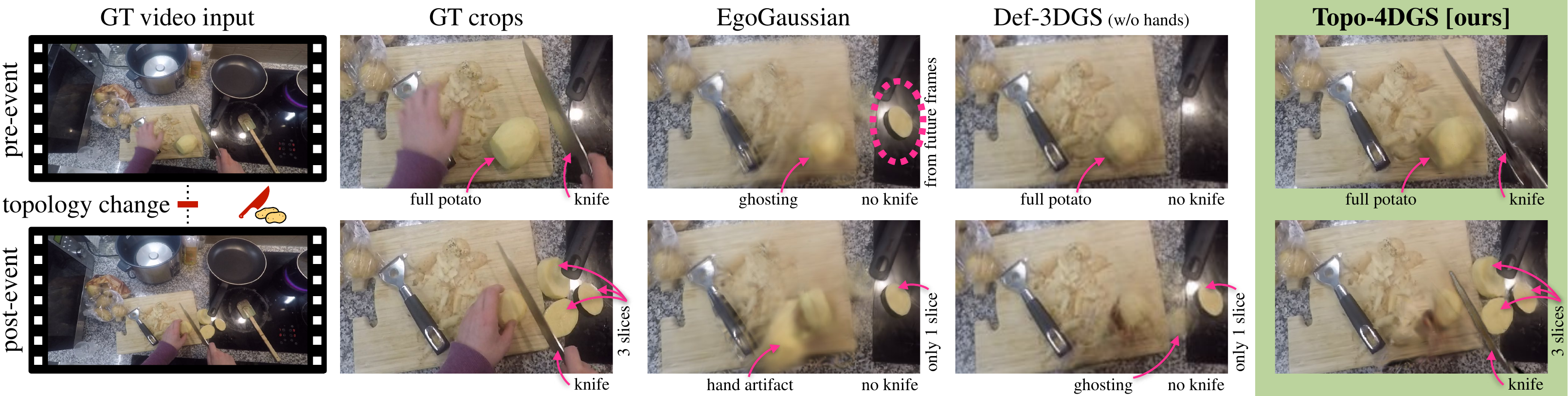}
    \vspace{-15pt}
    \captionof{figure}{
\textbf{Egocentric reconstruction across topology changes.} Common interactions like slicing violate the fixed-topology assumptions of existing 4DGS methods, resulting in severe artifacts (middle). By jointly modeling part-level motion and time-varying surface support, our Topo-4DGS faithfully reconstructs objects throughout topological transitions (right). Example from the proposed TopoEgo benchmark.
}
\vspace{6.8pt}
    \label{fig:teaser}
\end{center}%
}]

\begin{abstract}
Reconstructing dynamic scenes and objects from monocular egocentric video is crucial for embodied AI, yet existing 4D Gaussian Splatting (4DGS) approaches struggle with everyday interactions involving topological changes, such as cutting, assembly, opening, or closing. Since 4DGS methods rely on deforming a fixed representation, events that split, merge, or reveal surfaces cause severe visual artifacts. To address this challenge, we introduce \textbf{TopoEgo}, the first benchmark dedicated to topological changes in egocentric video (covering fission, fusion, open, and close), alongside \textbf{Topo-4DGS}, a topology-aware reconstruction method. Topo-4DGS jointly models part-level motion and time-varying surface support by assigning independent rigid transformations to interacting parts and dynamically activating or deactivating Gaussians as surfaces appear, disappear, or separate. Extensive experiments demonstrate that Topo-4DGS significantly outperforms state-of-the-art 4DGS approaches across topological changes, especially on post-event frames. Code and data will be made publicly available. Project page: \url{https://topoego.github.io}.
\end{abstract}

\section{Introduction}

\begin{figure}[t]
    \centering
    \includegraphics[width=\columnwidth]
        {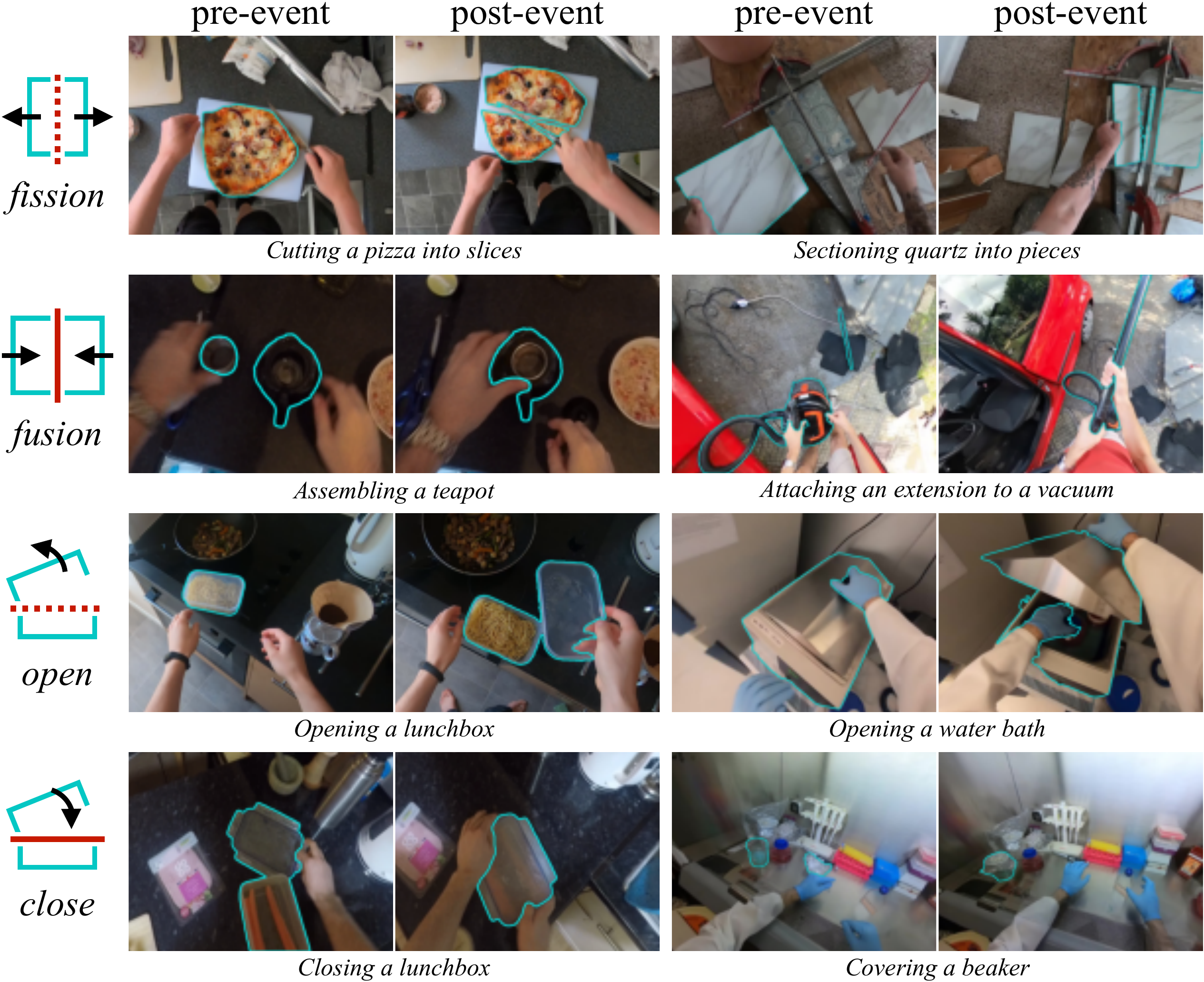}
    \caption{
        \textbf{TopoEgo examples of topological changes} showing pre- and post-event ground-truth frames from EPIC-KITCHENS and Ego4D. Cyan denotes the objects undergoing the changes.
    }
    \vspace{-10pt}
    \label{fig:topoego_examples}
\end{figure}

Dynamic 3D reconstruction from egocentric videos is important for robotics and AR applications, as it directly captures the physics and geometry of human-object interactions~\cite{zhang2024egogaussian,wang2025degauss}. However, everyday manipulation rarely involves purely rigid motion; it inherently induces complex object transformations and topological changes, such as cutting, assembling components, and opening containers (see Fig.~\ref{fig:teaser} and \ref{fig:topoego_examples}).
Because egocentric videos capture daily activities from the actors' viewpoint, such events are ubiquitous. Our analysis identifies them in over $68.8\%$ of Ego4D~\cite{grauman2022ego4d} and over $80.0\%$ of EPIC-KITCHENS~\cite{damen2022epickitchens} annotated videos.
Dynamically reconstructing these interactions in 3D requires more than estimating object motion: a method must determine which parts move together and adapt its geometric representation as surfaces are created, exposed, or concealed.

In particular, topological events violate the persistence assumptions of Gaussian-based monocular 4D reconstruction methods in two ways. First, \textit{motion structure may change}: regions that initially move together can become independent, whereas canonical deformation methods continuously deform fixed geometry~\cite{yang2024deformable3dgs,wu2024fourdgs}, tracking-based methods preserve primitive identities~\cite{luiten2024dynamic3dgs}, and EgoGaussian~\cite{zhang2024egogaussian} assigns a single rigid motion to each object. Second, \textit{surface support}, i.e., the set of 3D
surface locations represented by an object's Gaussian primitives, may change as surfaces are created, exposed, or concealed, causing persistent canonical primitives to produce missing geometry or ghosting ~\cite{yang2024deformable3dgs,wu2024fourdgs,luiten2024dynamic3dgs} (Fig.~\ref{fig:teaser}). 
However, topology-changing reconstruction remains largely underexplored: existing benchmarks rely on controlled capture with synchronized views or depth~\cite{park2021hypernerf}, and none systematically evaluate topological changes in real-world egocentric video.

To enable systematic evaluation of reconstruction across topological events, we introduce \textbf{TopoEgo}, a benchmark curated from EPIC-KITCHENS~\cite{damen2022epickitchens} and Ego4D~\cite{grauman2022ego4d}, two of the largest egocentric video datasets. TopoEgo captures diverse, in-the-wild instances of \textit{fission}, \textit{fusion}, \textit{open}, and \textit{close}, spanning changes in both object connectivity and surface visibility (see Fig.~\ref{fig:topoego_examples}). It contains 24 sequences, with 34 topological events and 36 dynamic objects. Each sequence includes event annotations and topology configurations, together with camera poses and hand-object masks.

To address this challenging benchmark, we introduce \textbf{Topo-4DGS}, a topology-aware reconstruction method that explicitly models the two consequences of topological events: changes in part-level motion and variations in surface support. This leads to Topo-4DGS's two main components. First, \emph{Motion Bifurcation} enables parts that initially move together to follow independent rigid motions after an event. Second, \emph{Topological Boundary Resolution} updates the represented surface support by adding or removing Gaussians as surfaces are created, exposed, or concealed. 

Our main contributions are:
\begin{itemize}
\item \textbf{TopoEgo}, a real-world benchmark for egocentric topological reconstruction, comprising 24 in-the-wild sequences with 36 dynamic objects, 34 annotated events, topology configurations, camera poses, hand-object masks, and standardized evaluation protocols.

\item \textbf{Topo-4DGS}, an event-conditioned representation for topology-aware 4D reconstruction, jointly modeling part-motion bifurcation and evolving surface support.

\item State-of-the-art performance across topological events, with the largest gains after event boundaries, where fixed-topology assumptions fail most severely.
\end{itemize}

\section{Related Work}
\label{sec:related_work}

\paragraph{Dynamic neural fields and canonical 4D representations.}
Dynamic novel-view synthesis models scenes whose geometry and appearance change over time. A common approach maps observations to a canonical representation: D-NeRF~\cite{pumarola2021dnerf} and Nerfies~\cite{park2021nerfies} continuously deform a canonical neural field, while Deformable 3D Gaussians~\cite{yang2024deformable3dgs} and 4DGS~\cite{wu2024fourdgs} apply this principle to Gaussian primitives. Dynamic 3D Gaussians~\cite{luiten2024dynamic3dgs} instead tracks persistent Gaussian identities, and MoRel~\cite{kwak2025morel} introduces locally canonical anchors for improved long-term consistency. Despite their different motion models, these methods preserve a fixed set of scene elements and do not explicitly change an object's part structure at topological events. 

Topo-4DGS instead branches object motion into event-conditioned rigid parts while updating the surface support created, exposed, or concealed by each event.

\paragraph{Egocentric reconstruction and scene decomposition.}
Egocentric reconstruction requires separating manipulated objects from the background and hands while accounting for rapid object and camera motion. EgoGaussian~\cite{zhang2024egogaussian} pioneered dynamic object reconstruction in egocentric videos. It is our closest baseline, representing each object in a canonical space and tracking it with a single rigid transformation. DeGauss~\cite{wang2025degauss} decomposes static and dynamic scene content to reduce distractors, while DP-DeGauss~\cite{chen2026dpdegaussdynamicprobabilisticgaussian} routes Gaussians into semantic background, hand, and object branches. These methods separate scene-level components but do not decompose one manipulated object into independently moving parts or update its surface support across an event. 

Topo-4DGS fills this gap by extending EgoGaussian's rigid prior to event-conditioned part-level motions.

\paragraph{Topology-changing and temporally adaptive surfaces.}
Reconstructing topological events requires a representation that can change both object connectivity and surface support over time. HyperNeRF~\cite{park2021hypernerf} addresses changing connectivity by embedding a canonical scene in a higher-dimensional space, but is designed for more controlled capture than monocular egocentric video. Other topology-aware approaches obtain stronger geometric constraints from RGB-D observations~\cite{li2020topology,cai2022neural} or synchronized and sparse multi-view capture~\cite{hartley2024neural,chen2025atgs,zheng2025gaustar,wu2026fourdsurf}. These observations are unavailable in egocentric video, where topology changes must be inferred despite camera motion, limited viewpoints, and hand occlusion. A complementary line of work changes surface support over time. TED-4DGS~\cite{ho2025ted4dgs} learns temporal activation windows that allow individual Gaussians to appear and disappear. We adopt this parameterization but explicitly condition it on structural events: our Topological Boundary Resolution activates Gaussians for created or exposed surfaces and deactivates them when surfaces become concealed. 
Moreover, we address topology-changing interactions using only a single moving RGB camera, without synchronized views or measured depth.

\paragraph{Benchmarks for dynamic and egocentric reconstruction.}
Existing datasets cover either dynamic reconstruction or egocentric scene understanding, but not their intersection under topological change. D-NeRF~\cite{pumarola2021dnerf}, Neural 3D Video~\cite{li2022neu3d}, and HyperNeRF~\cite{park2021hypernerf} evaluate dynamic scenes under controlled capture conditions. EPIC-Diff~\cite{tschernezki2021neuraldiff} benchmarks dynamic-object segmentation and novel-view synthesis in egocentric videos, labeling objects by motion rather than topological change. EPIC-Fields~\cite{tschernezki2023epicfields}, VISOR~\cite{darkhalil2022visor}, and Ego4D~\cite{grauman2022ego4d} provide egocentric camera geometry, object segmentation, and interaction annotations, but lack event-level topology labels and reconstruction protocols. Consequently, no existing benchmark systematically evaluates \textit{fission}, \textit{fusion}, \textit{open}, and \textit{close} in monocular egocentric video. TopoEgo fills this gap with manually annotated events and standardized 4D reconstruction evaluation.

\section{TopoEgo Benchmark}
\label{sec:dataset}

We define a \emph{topological event} as a temporally localized manipulation that changes an object's part connectivity, motion structure, or surface support. Each event separates two intervals: \emph{pre-event} and \emph{post-event}, to depict the object before and after the transformation, respectively. Reconstructing both intervals requires accounting for parts that begin or cease to move independently and surfaces exposed, concealed, created, or removed. The TopoEgo benchmark captures these transitions in unscripted, close-range interactions with camera motion, hand occlusion, and limited views (Fig.~\ref{fig:topoego_examples}).

\paragraph{Benchmark construction.}
TopoEgo is curated from EPIC-KITCHENS~\cite{damen2022epickitchens} and Ego4D~\cite{grauman2022ego4d}. The two datasets provide different forms of temporal action annotation, which we use only to retrieve candidate clips. Ego4D provides annotations for interactions that change an object's state. Each annotation specifies a state-transition type and the estimated frame at which the transition becomes irreversible. We map deconstruct transitions to \textit{fission}, construct to \textit{fusion}, and relevant activate/deactivate to \textit{open} and \textit{close}. %

Instead, EPIC-KITCHENS provides temporally bounded action segments with verb and noun labels. We retrieve segments whose verbs are associated with our taxonomy: cut, chop, slice, peel, grate, crack, break, snap, and disassemble for \textit{fission}; attach, screw, and tie for \textit{fusion}; open and uncover for \textit{open}; and close and cover for \textit{close}. 
\\
We then manually review every candidate at its native frame rate. We retain a clip only when the mapped event is clearly visible, the manipulated object is identifiable, and sufficient pre- and post-event context frames are available. We reject candidates in which the event occurs off-screen, is fully occluded, is not realized within the annotated interaction, or does not change object connectivity or surface support. 

We select three clips for each event-family--source pair, yielding six clips per event family and 12 clips from each source dataset. 
TopoEgo expands the scale of dynamic, egocentric novel-view synthesis benchmarks. It contains 24 clips, 4,618 frames, 36 annotated dynamic object instances, and 34 topological events, providing $1.5\times$--$6\times$ more evaluation sequences than existing benchmarks~\cite{zhang2024egogaussian,tschernezki2021neuraldiff,wang2025degauss,park2021hypernerf}. This scale enables comprehensive evaluation across diverse scenes, interactions, and topological transitions.

\paragraph{Annotations.}
For each topological event, we manually annotate its \emph{event time}, defined as the first frame in which the post-event state is visible, and its \emph{event family} (e.g., \textit{fission}). For each manipulated object, we record its \emph{terminal-piece count} $K$, defined as the number of leaf pieces after all events are resolved in the separation direction. Finally, an \emph{event tree} organizes the terminal pieces hierarchically by linking each parent part to its children, with the event time specifying when the children are released to move independently. 

We obtain camera intrinsics and extrinsics from EPIC Fields~\cite{tschernezki2023epicfields} where available and estimate them using COLMAP~\cite{schoenberger2016colmap} otherwise. Object and hand masks are generated using SAM~3 \cite{carion2026sam3segmentconcepts}. We will release these reconstruction inputs together with clip identifiers, benchmark manifests, preprocessing code, and evaluation scripts. %

\section{Preliminaries}
\label{sec:preliminaries}

\paragraph{3D Gaussian Splatting.}
3D Gaussian Splatting (3DGS)~\cite{kerbl20233dgs} represents a scene as an explicit set of anisotropic Gaussian primitives. Each primitive $G_i$ has a center $\boldsymbol{\mu}_i \in \mathbb{R}^3$, opacity $\alpha_i \in [0,1]$, view-dependent color $\mathbf{c}_i$, and covariance $\boldsymbol{\Sigma}_i \in \mathbb{R}^{3\times3}$, parameterized as
\begin{equation}
\begin{array}{c}
\boldsymbol{\Sigma}_i = \mathbf{R}_i \operatorname{diag}(\mathbf{s}_i^2) \mathbf{R}_i^\top,
\end{array}
\end{equation}
where $\mathbf{R}_i \in \mathrm{SO}(3)$ is a rotation, and $\mathbf{s}_i \in \mathbb{R}_{>0}^3$ contains the three principal scales.
To render pixel $\mathbf{u}$, the $N(\mathbf{u})$ Gaussians overlapping it are depth-ordered and alpha-composited front-to-back via differentiable \emph{rasterization}~\cite{kerbl20233dgs}:
\begin{equation}
\label{eq:splatting}
\begin{array}{r@{{}={}}l}
\mathbf{C}(\mathbf{u}) & \sum_{i=1}^{N(\mathbf{u})} \mathbf{c}_i \, \tilde{\alpha}_i(\mathbf{u}) \prod_{\ell<i} \left(1-\tilde{\alpha}_\ell(\mathbf{u})\right), \\[6pt]
\tilde{\alpha}_i(\mathbf{u}) & \alpha_i \hat{G}_i(\mathbf{u}),
\end{array}
\end{equation}
where $\mathbf{C}(\mathbf{u}) \in \mathbb{R}^3$ is the rendered pixel color and $\hat{G}_i(\mathbf{u})$ is the projected 2D footprint of $G_i$ at $\mathbf{u}$~\cite{zwicker2001ewa}.

A common class of dynamic neural and Gaussian methods~\cite{pumarola2021dnerf,park2021nerfies,yang2024deformable3dgs,wu2024fourdgs} uses \textit{canonical-space deformation}, representing motion by deforming a canonical scene—a single, time-independent reference configuration from which each frame is generated—over time. With
$\boldsymbol{\mu}^{\mathrm{can}}_i$ for a canonical Gaussian center, a typical
formulation predicts
\begin{equation}
\label{eq:canonical_deformation}
\begin{array}{c}
\boldsymbol{\mu}_i(t)=
\boldsymbol{\mu}^{\mathrm{can}}_i+
f_\theta(\boldsymbol{\mu}^{\mathrm{can}}_i,t),
\end{array}
\end{equation}
where $f_\theta(\boldsymbol{\mu}^{\mathrm{can}}_i,t)\in\mathbb{R}^3$ is a learned deformation field that predicts the displacement of Gaussian $i$ from its canonical position at time $t$. Related methods may also predict time-varying covariance, opacity, or color~\cite{yang2024deformable3dgs,li2024spacetime}.

\begin{figure}[t]
  \centering
  \includegraphics[width=\linewidth]{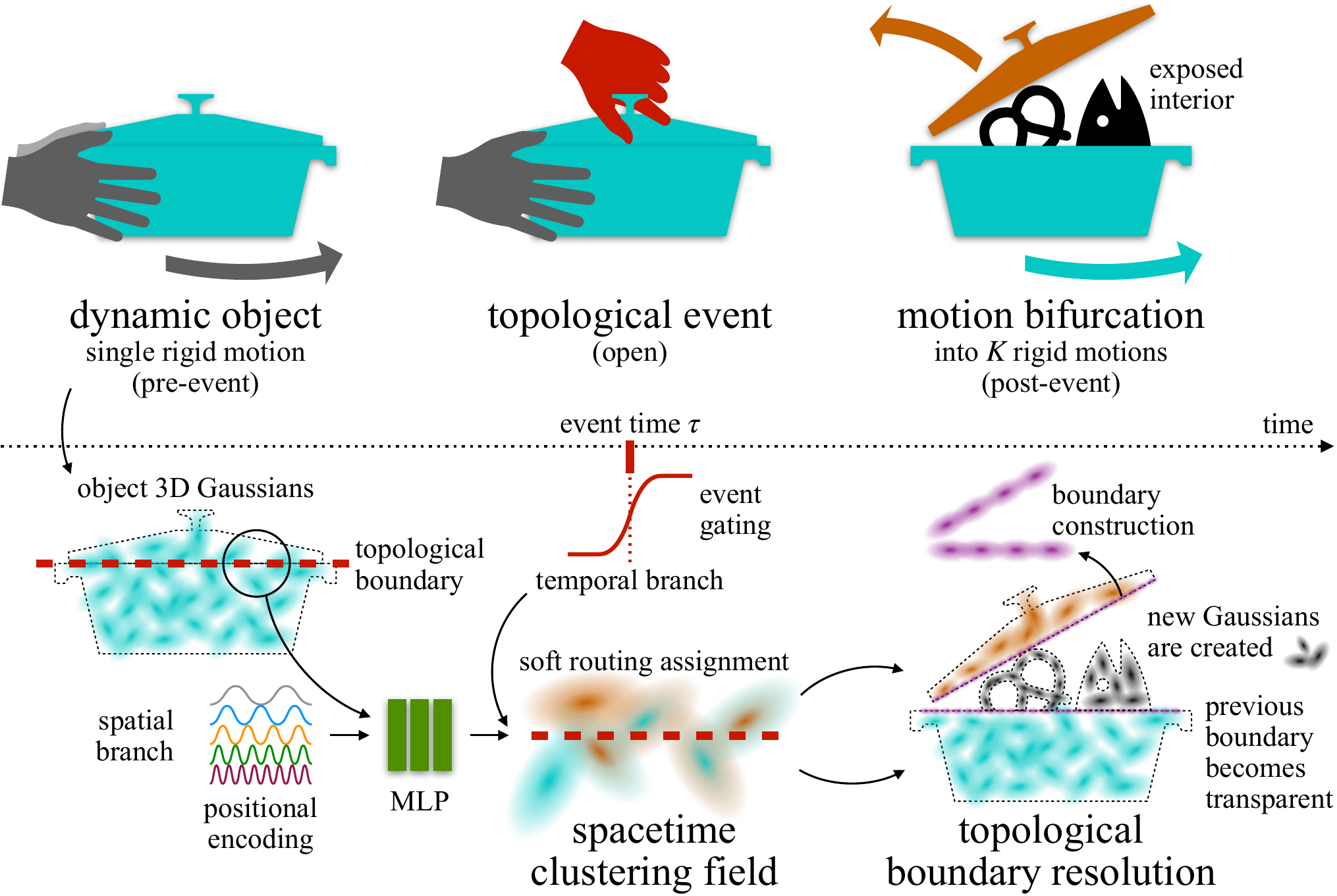}
  \caption{\textbf{Method overview of Topo-4DGS.}
  Top: at event time $\tau$, Motion Bifurcation transitions the object from a single rigid motion to $K$ piecewise-rigid trajectories. Bottom: the Spacetime Clustering Field combines the learned Spatial Branch with the event-gated Temporal Branch to softly route the object Gaussians to the corresponding motion branches. Topological Boundary Resolution creates support for newly formed or exposed surfaces and temporally deactivates obsolete or concealed support.}
  \label{fig:method}
\end{figure}

\section{Topo-4DGS: Topology-aware 4DGS}
\label{sec:method}
\label{sec:method_overview}

Given a monocular egocentric video, calibrated cameras, object and hand masks, and an event specification, Topo-4DGS jointly reconstructs a static background and multiple dynamic objects (Fig.~\ref{fig:method}). \textbf{Motion Bifurcation} (Sec.~\ref{sec:branching}) routes object Gaussians to event-dependent piecewise-rigid trajectories, while \textbf{Topological Boundary Resolution} (Sec.~\ref{sec:topo_resolution}) activates newly created or exposed surface support and deactivates support that becomes concealed. %

\subsection{Motion Bifurcation}
\label{sec:branching}

A scene may contain multiple dynamic objects (e.g., vegetable and knife), indexed by
$\mathcal{J}=\{1,\ldots,J\}$. For clarity, we formulate the method for one
object and omit its index $j$. Let $\mathcal{K}=\{1,\ldots,K\}$ index
its terminal pieces, defined as the leaves of its event tree under the
separation convention, and let
$\mathcal{E}=(\tau_1,\ldots,\tau_M)$ denote its ordered event times. A rooted event tree $\mathcal{T}$ groups terminal pieces constrained to share a rigid transform. At each $\tau_m$, a parent splits into child groups that may move independently thereafter. Below, we discuss the \textit{fission}/\textit{open} case. For \textit{fusion}/\textit{close}, we reverse the temporal order of the frames, camera poses, masks, and event times. Consequently, a sequence in which $K$ separate pieces merge into one is processed as one object separating into $K$ pieces. After reconstruction, we map the outputs back to the original temporal order.

\paragraph{Soft routing.}
Each terminal piece $k\in\mathcal{K}$ follows a rigid trajectory $\mathbf{T}_k(t)\in\mathrm{SE}(3)$ at frame $t\in\{1,\ldots,N\}$, with rotation $\mathbf{R}_k(t)\in\mathrm{SO}(3)$ and translation $\mathbf{d}_k(t)\in\mathbb{R}^3$. Each canonical Gaussian $\mathcal{G}_i$ has a time-dependent routing distribution $\{P_{i,k}(t)\}_{k=1}^{K}$ over the terminal pieces, where $P_{i,k}(t)\geq0$ and $\sum_{k=1}^{K}P_{i,k}(t)=1$. Its deformed center is obtained by blending the corresponding candidate motions:
\begin{equation}
\label{eq:softroute}
\begin{array}{c}
\boldsymbol{\mu}_i(t)=\sum_{k=1}^{K}P_{i,k}(t)\left(\mathbf{R}_k(t)\boldsymbol{\mu}^{\mathrm{can}}_i+\mathbf{d}_k(t)\right).
\end{array}
\end{equation}
The same weights blend the rotated canonical covariance factors: $\mathbf{L}_i(t)=\sum_k P_{i,k}(t)\mathbf{R}_k(t)\mathbf{L}^{\mathrm{can}}_i$, with $\boldsymbol{\Sigma}_i(t)=\mathbf{L}_i(t)\mathbf{L}_i(t)^\top$. 
When the distribution is one-hot, $\mathcal{G}_i$ is effectively hard-assigned to one terminal piece; otherwise, it is softly assigned across multiple pieces. 
We next describe how the routing weights $P_{i,k}(t)$ in Eq.~\eqref{eq:softroute} are computed.

\paragraph{Spatial branch.}
\label{spatial_branch}
This branch assigns with logits to the $K$ pieces each Gaussian $\mathcal{G}_i$ with canonical center $\bar{\boldsymbol{\mu}}^{\mathrm{can}}_i$:
\begin{equation}
\label{eq:spatial_logits}
\mathbf{z}^{\mathrm{sp}}_i=h_\phi\!\left(\gamma(\bar{\boldsymbol{\mu}}^{\mathrm{can}}_i)\right)\in\mathbb{R}^{K},
\end{equation}
where $h_\phi$ is an MLP and $\gamma$ is a low-frequency positional encoding that encourages spatially coherent assignments. Because these logits are time-independent, they prevent unrestricted frame-by-frame changes in piece assignment. \label{Kmeans} We use a short $K$-means warm start to break permutation symmetry. Further normalization, initialization, and regularization details are provided in the Supp.

\paragraph{Temporal branch.}
The temporal branch encodes whether each annotated event has occurred using a smooth event gate. Let $\bar{t}$ and $\bar{\tau}_m$ denote the normalized frame and event times, respectively. For event $m$, we define
\begin{equation}
\label{eq:event_gate}
g_m(\bar{t})=\sigma\!\left(\frac{\bar{t}-\bar{\tau}_m}{\beta}\right),
\qquad
\sigma(u)=\frac{1}{1+e^{-u}},
\end{equation}
where $\beta>0$ controls the temporal width of the transition. We fix $\beta=0.25$ for all events and experiments. The gate satisfies $g_m(\bar{t})\approx0$ before the event, $g_m(\bar{\tau}_m)=0.5$ at the event time, and $g_m(\bar{t})\approx1$ after the event. We collect the gates into $\mathbf{g}(\bar{t})=[g_1(\bar{t}),\ldots,g_M(\bar{t})]^\top\in\mathbb{R}^{M}$.

\paragraph{Spacetime Clustering Field.}
The Spacetime Clustering Field combines the spatial logits of Gaussian $\mathcal{G}_i$ with the temporal event gates to produce its routing weights:
\begin{equation}
\label{eq:routing}
P_{i,k}(t)=
\left[
\operatorname{Softmax}\!\left(
\frac{\mathbf{z}^{\mathrm{sp}}_i+\mathbf{B}\mathbf{g}(\bar{t})}{\lambda}
\right)
\right]_k,
\qquad k\in\mathcal{K},
\end{equation}
where $[\cdot]_k$ denotes the $k$-th component of the resulting $K$-dimensional routing distribution, $\bar{t}$ is the normalized timestamp for frame $t$, $\mathbf{B}\in\mathbb{R}^{K\times M}$ is a learned event-bias matrix, and $\lambda>0$ is the routing temperature. The $m$-th column of $\mathbf{B}$ determines how event $m$ modifies the routing logits over the $K$ terminal pieces. We initialize $\mathbf{B}$ to zero, so the initial routing depends only on the spatial logits. \label{UTMLP} Because time enters only through the event gates, routing remains stable between events. In our ablation, an MLP-based temporal branch replaces $\mathbf{B}\mathbf{g}(\bar{t})$, allowing frame-to-frame assignment changes that may induce artificial motion.

\subsection{Topological Boundary Resolution}
\label{sec:topo_resolution}

Motion Bifurcation (Sec.~\ref{sec:branching}) changes how Gaussians move but cannot create or deactivate surface support. Instead, Topological Boundary Resolution introduces surfaces exposed by \textit{fission} or \textit{open} and retires obsolete or concealed support at \textit{fusion} or \textit{close} events. As in Sec.~\ref{sec:branching}, \textit{fusion} and \textit{close} are handled by reversing the time formulation.

\paragraph{Per-piece boundary construction.}
\label{PC}
When two terminal pieces $p$ and $q$ separate at time $t_{\mathrm{sep}}(p,q)$, the newly exposed boundary contains one face attached to $p$ and another attached to $q$. We therefore construct the two faces independently in their respective canonical frames. For each piece, we identify the contact region using mutual nearest-neighbor correspondences with the other piece, fit a plane to the matched Gaussian centers, project the points onto the plane, and fill their 2D convex hull with new Gaussians. The new Gaussians inherit their initial appearance and geometry from nearby Gaussians and are bound to the motion branch of their corresponding piece.
Further construction and branch-binding details are provided in the Supp.

\paragraph{Event-conditioned temporal activation.}
\label{temporal_gate}
We associate each Gaussian $\mathcal{G}_i$ with a differentiable activation window $\mathcal{W}_i(t)$ and modulate its opacity as $\alpha_i^t=\alpha_i\mathcal{W}_i(t)$. Following TED-4DGS~\cite{ho2025ted4dgs}, we parameterize the window using an activation boundary $a_{s,i}$, a deactivation boundary $a_{f,i}$, and transition widths $b_{s,i},b_{f,i}>0$:
\begin{equation}
\label{eq:activation_window}
\mathcal{W}_i(t)=
\begin{cases}
\exp\!\left[-\left(\dfrac{t-a_{s,i}}{b_{s,i}}\right)^2\right], & t<a_{s,i},\\[4pt]
1, & a_{s,i}\leq t\leq a_{f,i},\\[4pt]
\exp\!\left[-\left(\dfrac{t-a_{f,i}}{b_{f,i}}\right)^2\right], & t>a_{f,i}.
\end{cases}
\end{equation}
Thus, $\mathcal{G}_i$ is fully active between $a_{s,i}$ and $a_{f,i}$ and smoothly fades outside this interval. Unlike~\cite{ho2025ted4dgs}, though, which learns temporal support without explicit structural supervision, we initialize these boundaries from topological events. Gaussians representing newly created or exposed surfaces receive an activation boundary at the corresponding event time, whereas Gaussians representing obsolete or concealed surfaces receive a deactivation boundary. Gaussians unaffected by an event remain active throughout the sequence.

\paragraph{Time refinement.}
\label{sec:learned_timers}
During final refinement, the activation and deactivation boundaries are adjusted to better match the observed images, subject to $a_{f,i}\geq a_{s,i}$. This permits small per-Gaussian timing corrections while leaving the annotated event times unchanged. Further details are in the Supp.

\subsection{Optimization}
\label{sec:objective}
\label{sec:interp}

To support multiple manipulated objects, we apply EgoGaussian's single-object pose-estimation procedure~\cite{zhang2024egogaussian} independently to each object $j\in\mathcal{J}$, obtaining separate per-object trajectories by estimating coarse poses at valid frames and interpolating missing poses (see Supp.). We then jointly refine the object's canonical Gaussian representation and motion, additionally optimizing the piece trajectories and Spacetime Clustering Field. The optimization uses a weighted combination of $L_1$ and D-SSIM image-reconstruction losses, denoted by $\mathcal{L}_{\mathrm{rgb}}$. Pixels occluded by the hands are excluded from the loss using the provided hand masks.

Let $\mathbf{C}^{(j)}$, $A^{(j)}$, and $S_j$ denote the rendered RGB image, rendered opacity, and binary mask of object $j$, respectively, and let $\mathbf{C}^{\mathrm{gt}}$ denote the observed frame. The objective is
\begin{equation}
\label{eq:obj_objective}
\mathcal{L}^{(j)}_{\mathrm{obj}}
=
\mathcal{L}_{\mathrm{rgb}}
\bigl(\mathbf{C}^{(j)},S_j\odot\mathbf{C}^{\mathrm{gt}}\bigr)
+
\lambda_{\alpha}\lVert A^{(j)}-S_j\rVert_2^2
+
\lambda_{\mathrm{sp}}\mathcal{L}_{\mathrm{sp}},
\end{equation}
Here, $\lambda_{\alpha}$ and $\lambda_{\mathrm{sp}}$ weight the silhouette and spatial-coherence terms, respectively. For objects with $K>1$, we sample $\boldsymbol{\epsilon}\sim \mathcal{N}(\mathbf{0},\sigma^2\mathbf{I}_3)$ and let $\mathbf{P}_{i,\boldsymbol{\epsilon}}(t)$ denote the routing distribution obtained by evaluating the spatial branch at $\bar{\boldsymbol{\mu}}^{\mathrm{can}}_i+\boldsymbol{\epsilon}$ instead of $\bar{\boldsymbol{\mu}}^{\mathrm{can}}_i$. We define 
\begin{equation} 
\label{eq:spatial_regularizer} 
\mathcal{L}_{\mathrm{sp}} = \mathbb{E}_{i,t,\boldsymbol{\epsilon}} \left[ \left\| \mathbf{P}_i(t) - \mathbf{P}_{i,\boldsymbol{\epsilon}}(t) \right\|_1 \right]. 
\end{equation} 
This regularizer encourages nearby canonical positions to have similar routing distributions, discouraging fragmented assignments without requiring piece labels. %

\begin{table*}[t]
\centering

\vspace{0pt}
\centering

\caption{\textbf{Reconstruction comparison} on TopoEgo. Object-region PSNR$\uparrow$ is reported for each event family and source dataset. Hand-masked baselines are added for a fair comparison. The last three columns report performance on clips from~\cite{damen2022epickitchens,grauman2022ego4d} without topological changes.
}
\label{tab:final_primary}

{\small %
\setlength{\tabcolsep}{6.2pt}   %

\begin{tabular}{l*{8}{c}|*{3}{c}}

\toprule
& \multicolumn{4}{c}{TopoEgo (EPIC-KITCHENS)}
& \multicolumn{4}{c}{TopoEgo (Ego4D)}
& \multicolumn{3}{|c}{Rigid Objects Control} \\
\cmidrule(lr){2-5}
\cmidrule(lr){6-9}
\cmidrule(lr){10-12}
Method
& \textit{fission} & \textit{fusion} & \textit{open} & \textit{close}
& \textit{fission} & \textit{fusion} & \textit{open} & \textit{close} 
& PSNR$\uparrow$
& SSIM$\uparrow$
& LPIPS$\downarrow$ \\
\midrule

EgoGaussian~\cite{zhang2024egogaussian}
& 20.48 & 25.17 & 18.32 & 22.20
& 22.97 & 20.76 & 17.70 & 19.69
& \secondcell{27.34}
& \secondcell{0.804}
& \secondcell{0.186} \\

DeGauss~\cite{wang2025degauss}
& 21.39 & 25.60 & 18.58 & 23.24
& 24.37 & 22.23 & \thirdcell{19.51} & 20.99
& 25.10 & 0.742 & 0.232 \\

DeGauss (w/o hands)
& 20.15 & 21.70 & 16.70 & 19.40
& 21.03 & 19.92 & 17.20 & 18.64
& 23.07 & 0.651 & 0.317 \\

Def-3DGS~\cite{yang2024deformable3dgs}
& 21.93 & \thirdcell{26.03}
& \thirdcell{18.93} & \secondcell{23.58}
& \secondcell{24.64} & 22.54
& \secondcell{19.67} & 21.31
& 24.37 & 0.732 & 0.243 \\

Def-3DGS (w/o hands)
& \secondcell{22.70} & \secondcell{26.70}
& \secondcell{19.05} & \thirdcell{23.35}
& \thirdcell{24.45} & \thirdcell{23.10}
& 19.40 & \secondcell{22.05}
& \thirdcell{26.44}
& \thirdcell{0.764}
& \thirdcell{0.197} \\

4DGS~\cite{wu2024fourdgs}
& 20.36 & 24.97 & 17.75 & 22.28
& 22.79 & 20.41 & 17.40 & 19.03
& 23.91 & 0.676 & 0.288 \\

4DGS (w/o hands)
& \thirdcell{22.10} & 25.10 & 18.20 & 22.70
& 23.20 & \secondcell{23.46}
& 18.52 & \thirdcell{21.72}
& 25.26 & 0.693 & 0.270 \\

\textbf{Topo-4DGS [ours]}
& \bestcell{23.84} & \bestcell{27.33}
& \bestcell{20.31} & \bestcell{24.48}
& \bestcell{25.33} & \bestcell{23.82}
& \bestcell{20.80} & \bestcell{22.23}
& \bestcell{27.92}
& \bestcell{0.833}
& \bestcell{0.174} \\

\bottomrule
\end{tabular}

}

\end{table*}

\begin{table*}[t]
\centering
\caption{\textbf{Additional comparison} on TopoEgo.
Results are averaged over all 24 clips for the full scene, the
topology-changing object over all frames (reported per event type and dataset in Tab.~\ref{tab:final_primary}) and post-event frames, and other
dynamic objects (i.e., manipulated, rigid objects).}
\label{tab:final_complementary}
\small
\setlength{\tabcolsep}{5.5pt}

\begin{tabular}{l ccc ccc ccc ccc}
\toprule
& \multicolumn{3}{c}{Full Scene}
& \multicolumn{3}{c}{Topo. Object}
& \multicolumn{3}{c}{Post-Event Object}
& \multicolumn{3}{c}{Other Dyn. Objects} \\
\cmidrule(lr){2-4}
\cmidrule(lr){5-7}
\cmidrule(lr){8-10}
\cmidrule(lr){11-13}

Method
& PSNR & SSIM & LPIPS
& PSNR & SSIM & LPIPS
& PSNR & SSIM & LPIPS
& PSNR & SSIM & LPIPS \\
\midrule

EgoGaussian~\cite{zhang2024egogaussian}
& 23.82 & 0.729 & \secondcell{0.249}
& 20.91 & 0.638 & \secondcell{0.247}
& 16.79 & 0.464 & \secondcell{0.300}
& 20.51 & 0.602 & 0.302 \\

DeGauss~\cite{wang2025degauss}
& 26.64 & 0.837 & 0.264
& 21.99 & 0.698 & 0.257
& 18.76 & 0.591 & \thirdcell{0.310}
& 23.65 & 0.754 & \thirdcell{0.224} \\

DeGauss (w/o hands)
& 23.44 & 0.844 & 0.271
& 19.34 & 0.704 & 0.264
& 16.55 & 0.598 & 0.318
& 20.81 & \thirdcell{0.760} & 0.231 \\

Def-3DGS~\cite{yang2024deformable3dgs}
& 27.15 & \thirdcell{0.846} & \thirdcell{0.256}
& \thirdcell{22.33} & \thirdcell{0.705} & \thirdcell{0.256}
& \thirdcell{19.31} & \thirdcell{0.600} & 0.312
& \thirdcell{23.97} & 0.752 & \secondcell{0.223} \\

Def-3DGS (w/o hands)
& \secondcell{27.80} & \secondcell{0.858} & 0.263
& \secondcell{22.60} & \secondcell{0.711} & 0.266
& \secondcell{19.83} & \secondcell{0.611} & 0.320
& \secondcell{24.69} & \secondcell{0.764} & 0.230 \\

4DGS~\cite{wu2024fourdgs}
& 25.79 & 0.756 & 0.353
& 20.62 & 0.622 & 0.310
& 16.62 & 0.456 & 0.376
& 21.06 & 0.635 & 0.284 \\

4DGS (w/o hands)
& \thirdcell{27.20} & 0.770 & 0.362
& 21.88 & 0.635 & 0.319
& 17.59 & 0.469 & 0.387
& 22.32 & 0.648 & 0.293 \\

\textbf{Topo-4DGS [ours]}
& \bestcell{27.95} & \bestcell{0.885} & \bestcell{0.179}
& \bestcell{23.52} & \bestcell{0.740} & \bestcell{0.212}
& \bestcell{24.30} & \bestcell{0.742} & \bestcell{0.227}
& \bestcell{25.90} & \bestcell{0.786} & \bestcell{0.189} \\

\bottomrule
\end{tabular}
\end{table*}

\paragraph{Joint scene refinement.}
\label{sec:joint_scene}
Finally, we jointly render the resolved objects and static background using the same photometric loss $\mathcal{L}_{\mathrm{rgb}}$ defined above, again excluding hand pixels from the supervision. This stage refines the background and each object's appearance, Gaussian opacity, scale, and rotation. Gaussian centers, recovered poses, routing assignments, and the event tree remain fixed, and densification is disabled. The time refinement step updates only the activation and deactivation
boundaries through this joint objective.

\section{Experiments and Results}
\label{sec:experiments}

\paragraph{Benchmark and split.}
We evaluate Topo-4DGS on the 24 TopoEgo sequences (Sec.~\ref{sec:dataset}). Every clip retains its native frame rate and is processed at one quarter of its resolution. Frames alternate between training and evaluation in a $1{:}1$ split, following EgoGaussian's~\cite{zhang2024egogaussian} evaluation protocol. All methods use the same held-out frames, camera poses, and evaluation annotations. The experiments on TopoEgo use the provided manual piece counts, event trees, and event times. We additionally evaluate our model on a separate six-clip control set without topological changes, referred to as the \textit{Rigid Objects Control} set. The set comprises four clips from the original EgoGaussian benchmark and two randomly sampled Ego4D clips. This evaluation verifies that our method preserves EgoGaussian's performance on dynamic objects in the absence of topological events. Due to computational time constraints, the ablations in Tab.~\ref{tab:abl_method} and Tab.~\ref{tab:abl_alternatives} use an 8-clip subset of TopoEgo, containing one clip for each event-family--dataset pair.  

\paragraph{Evaluation.}
We report PSNR, SSIM~\cite{wang2004ssim}, and LPIPS~\cite{zhang2018lpips} on full frames and object regions. Full-scene metrics evaluate the composite render. For object regions, PSNR is computed over the ground-truth mask pixels, SSIM over the masked pixels of a padded image crop, and LPIPS over its bounding-box crop. We evaluate the topology-changing object and other dynamic objects separately. This composite protocol supports all baselines. Following EgoGaussian, hand pixels are zeroed in both prediction and ground truth only during metric computation. %

\begin{figure*}[t]
    \centering
    \footnotesize
    \setlength{\tabcolsep}{0.35pt}
    \renewcommand{\arraystretch}{1.02}

    \newcommand{\qualimg}[1]{%
      \includegraphics[width=0.161\textwidth]{#1}%
    }

    \begin{tabular}{cc c c c c c c}
      &&
      GT pre-event &
      GT post-event &
      EgoGaussian &
      Def-3DGS (w/o hands) &
      4DGS (w/o hands) &
      \textbf{Topo-4DGS [ours]}
      \\[0.2pt]

      \rotatebox{90}{\textit{fission}} &&
      \qualimg{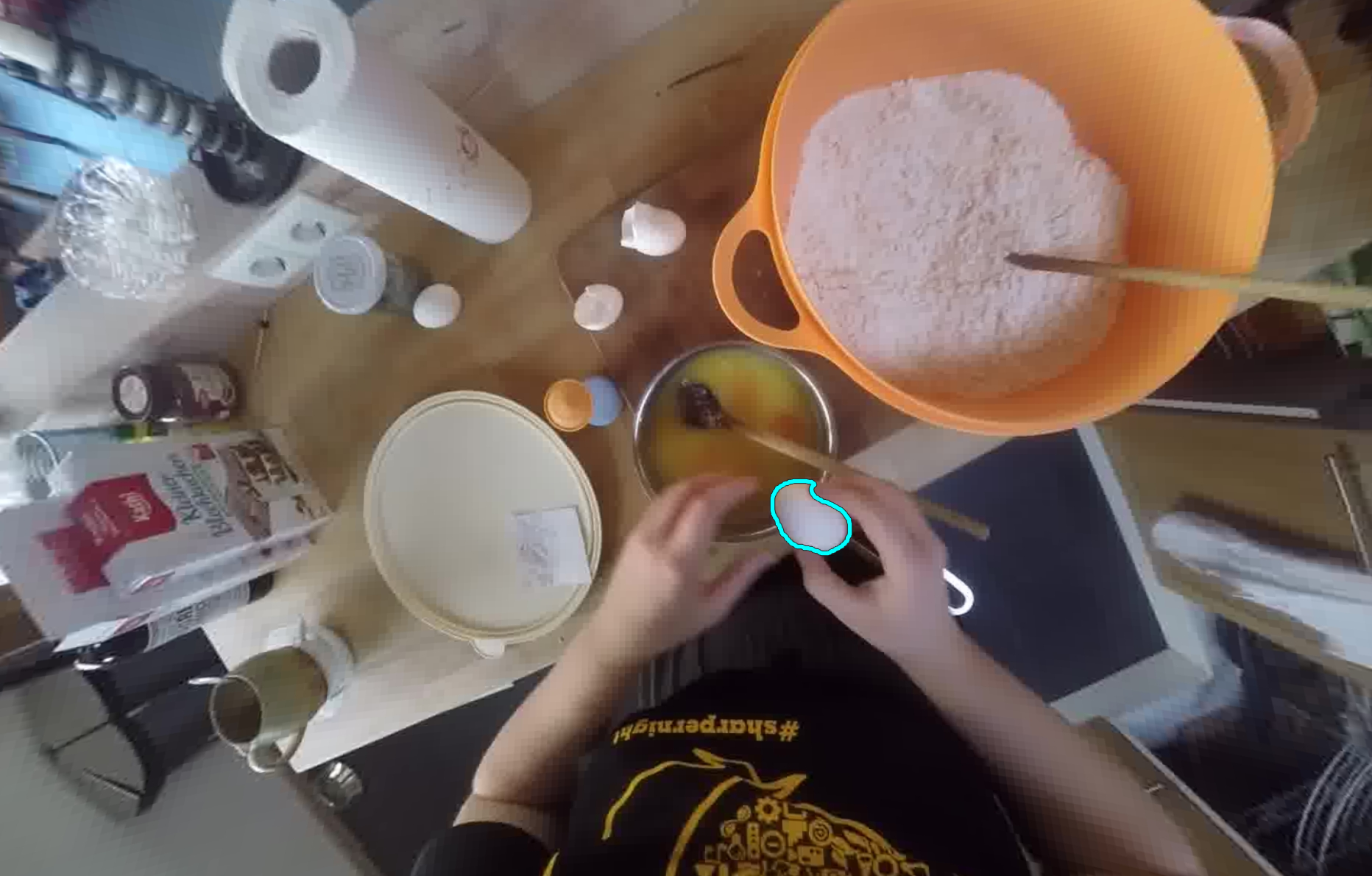} &
      \qualimg{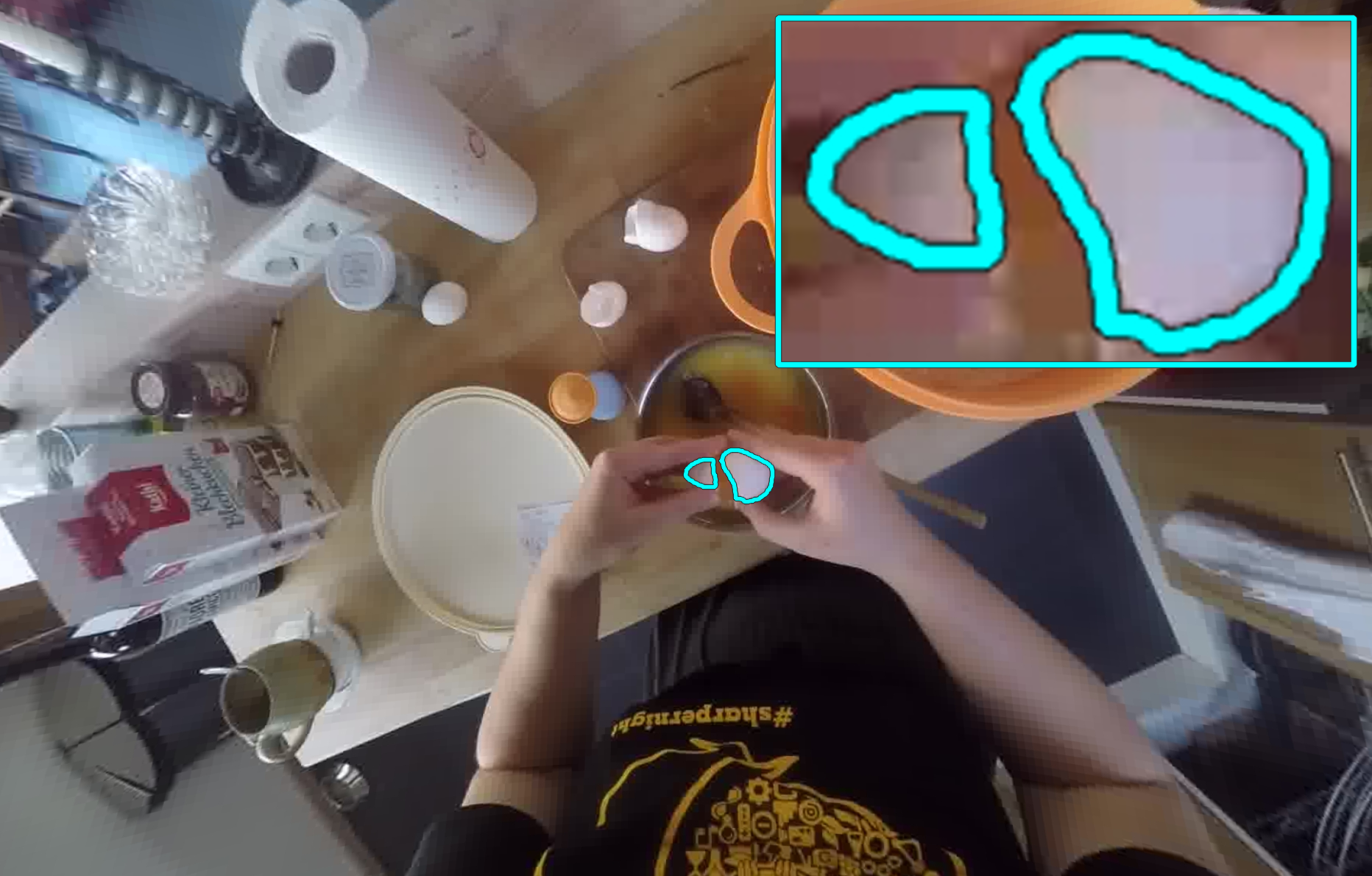} &
      \qualimg{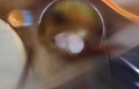} &
      \qualimg{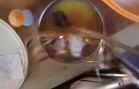} &
      \qualimg{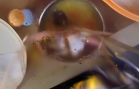} &
      \qualimg{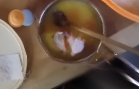}
      \\[-2.5pt]
      
      \rotatebox{90}{\textit{fusion}} &&
      \qualimg{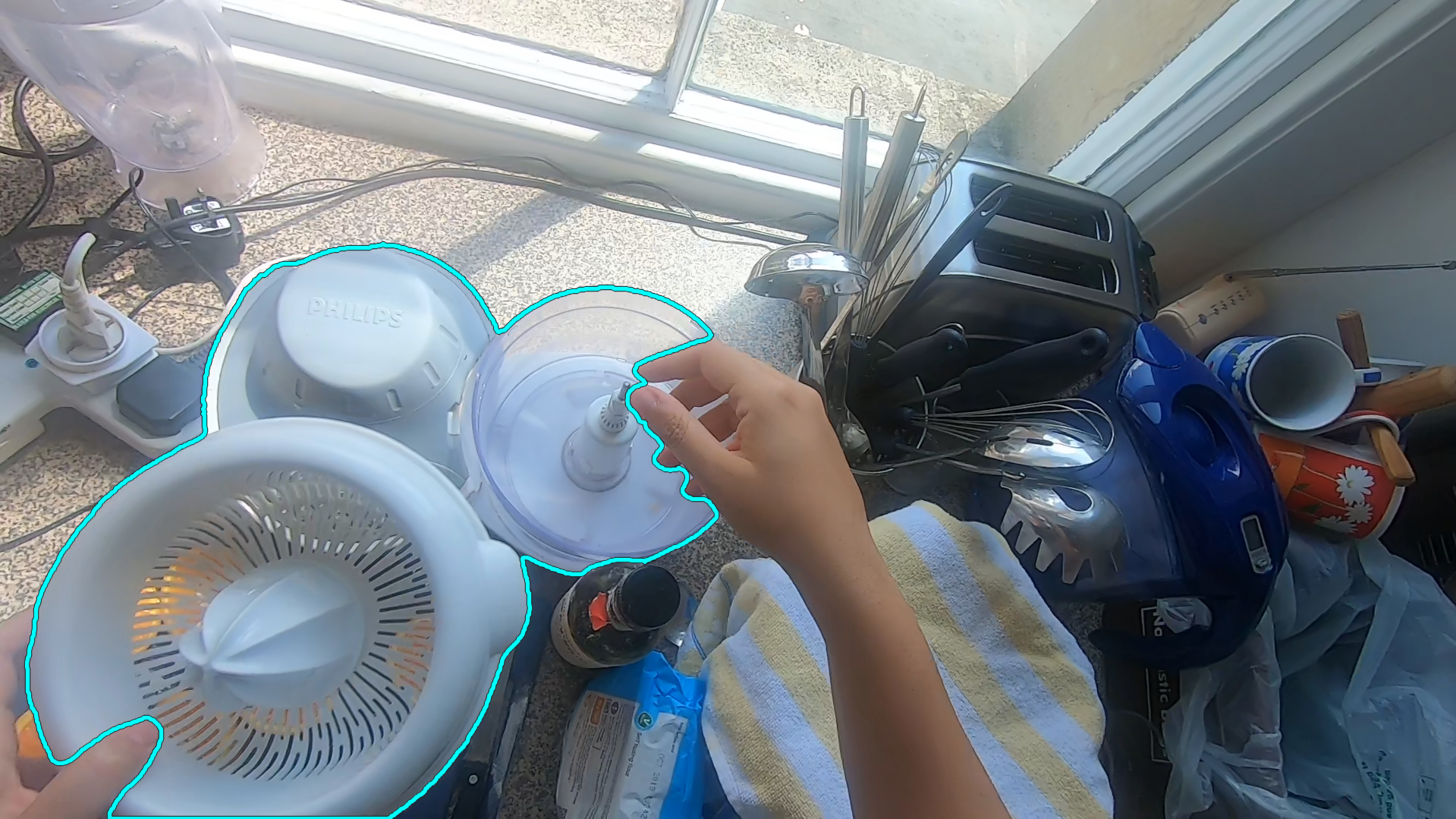} &
      \qualimg{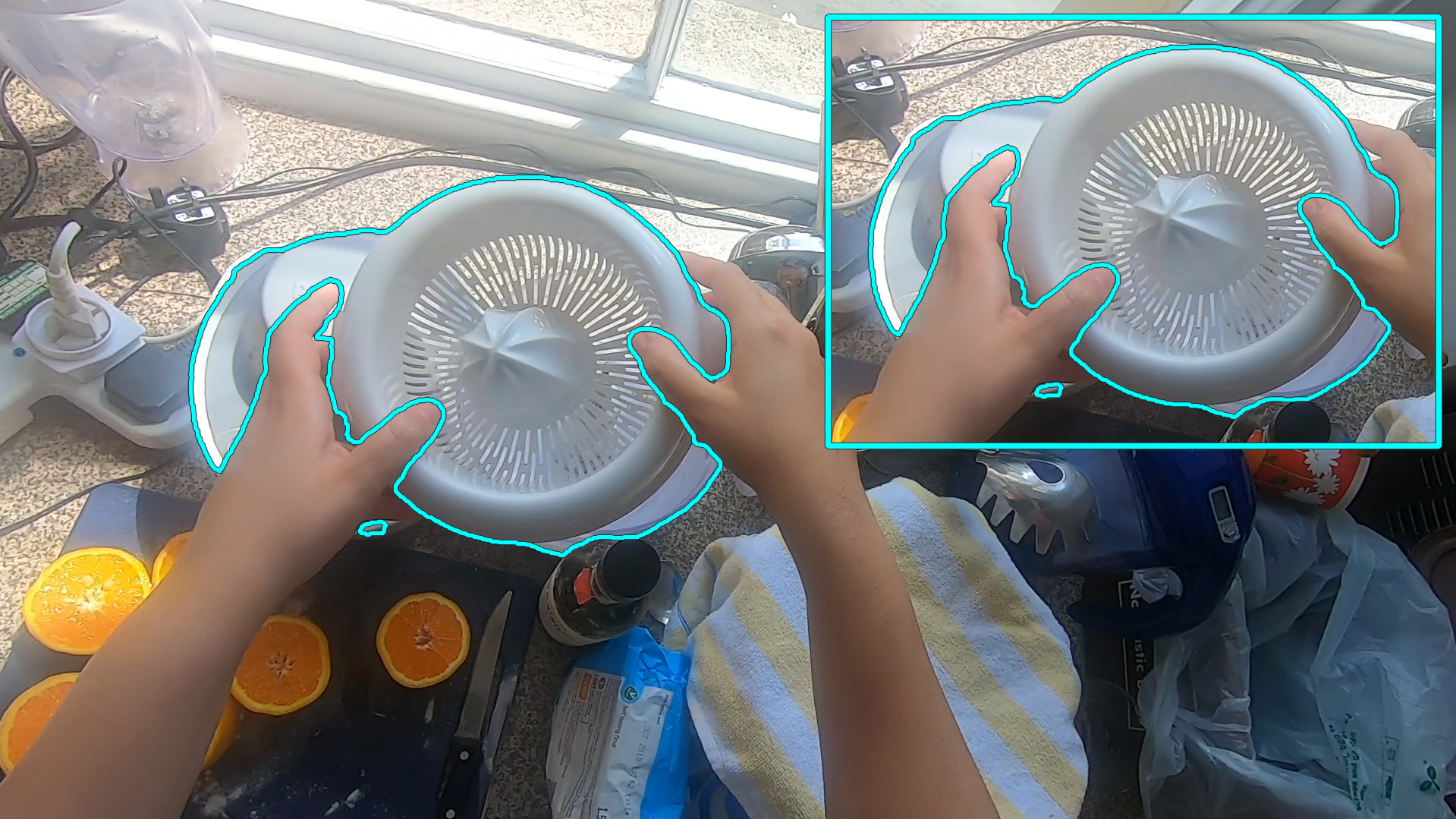} &
      \qualimg{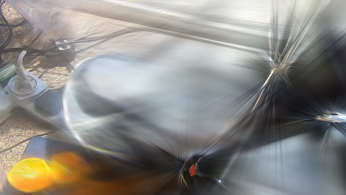} &
      \qualimg{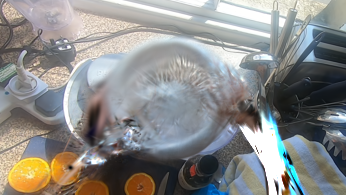} &
      \qualimg{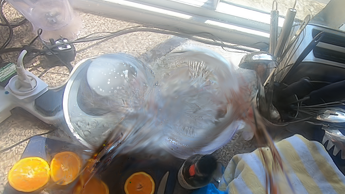} &
      \qualimg{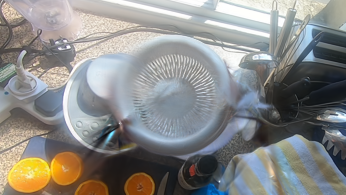}
      \\[-2.5pt]
      
      \rotatebox{90}{\textit{open}} &&
      \qualimg{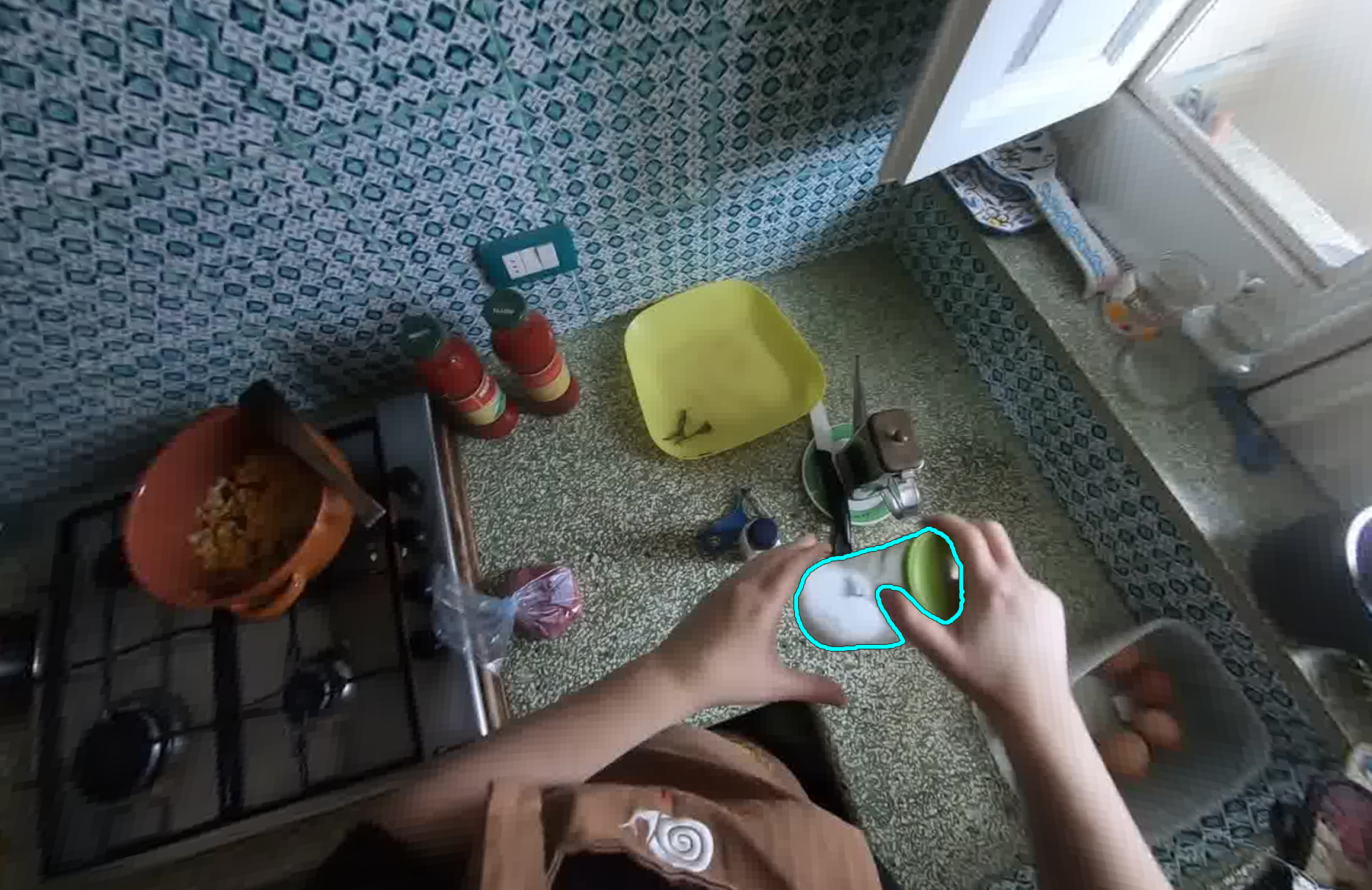} &
      \qualimg{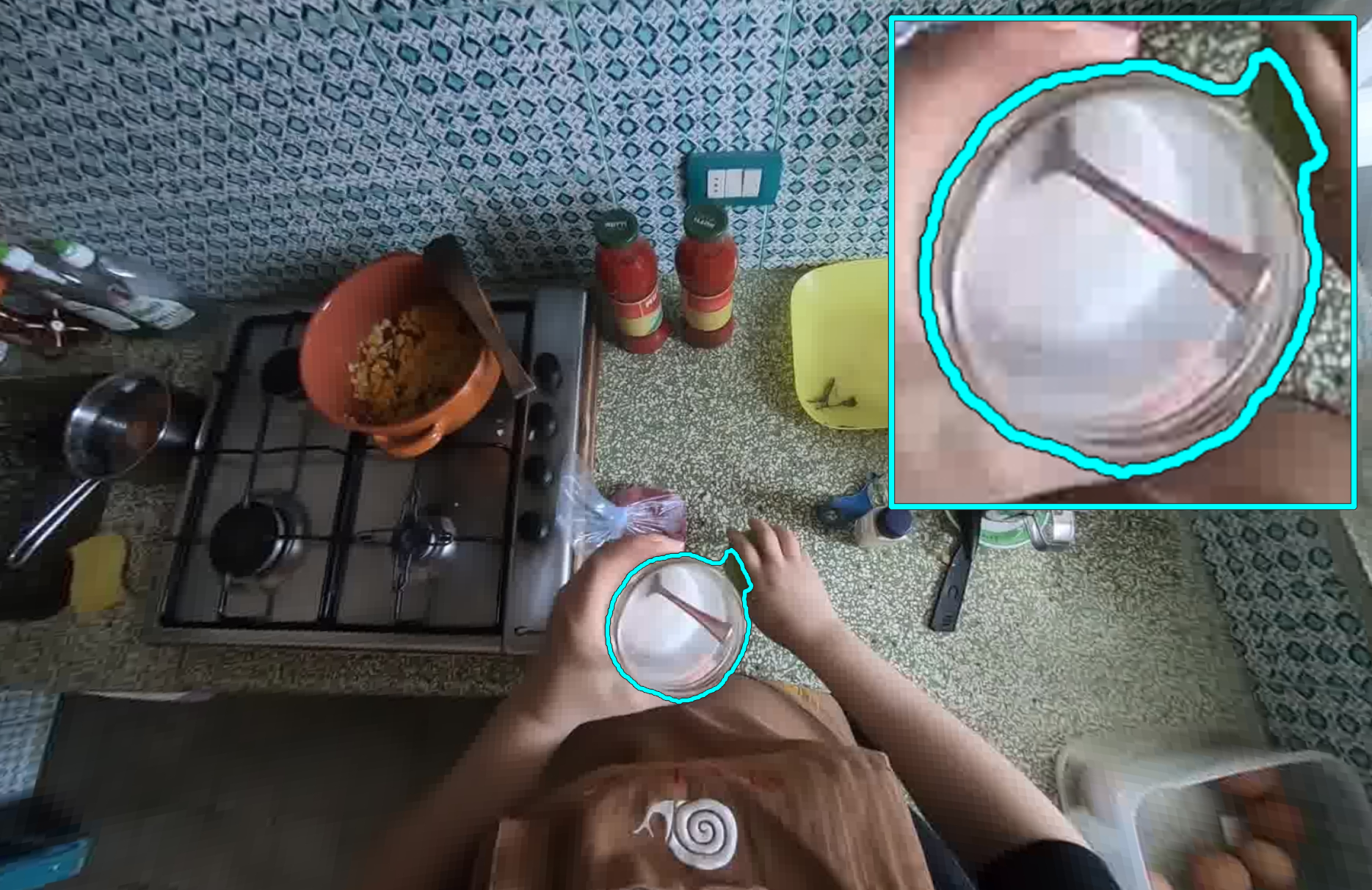} &
      \qualimg{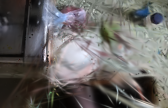} &
      \qualimg{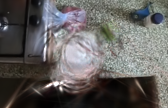} &
      \qualimg{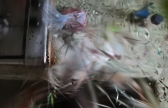} &
      \qualimg{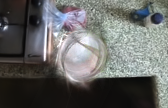}
      \\[-2.5pt]

      \rotatebox{90}{\textit{close}} &&
      \qualimg{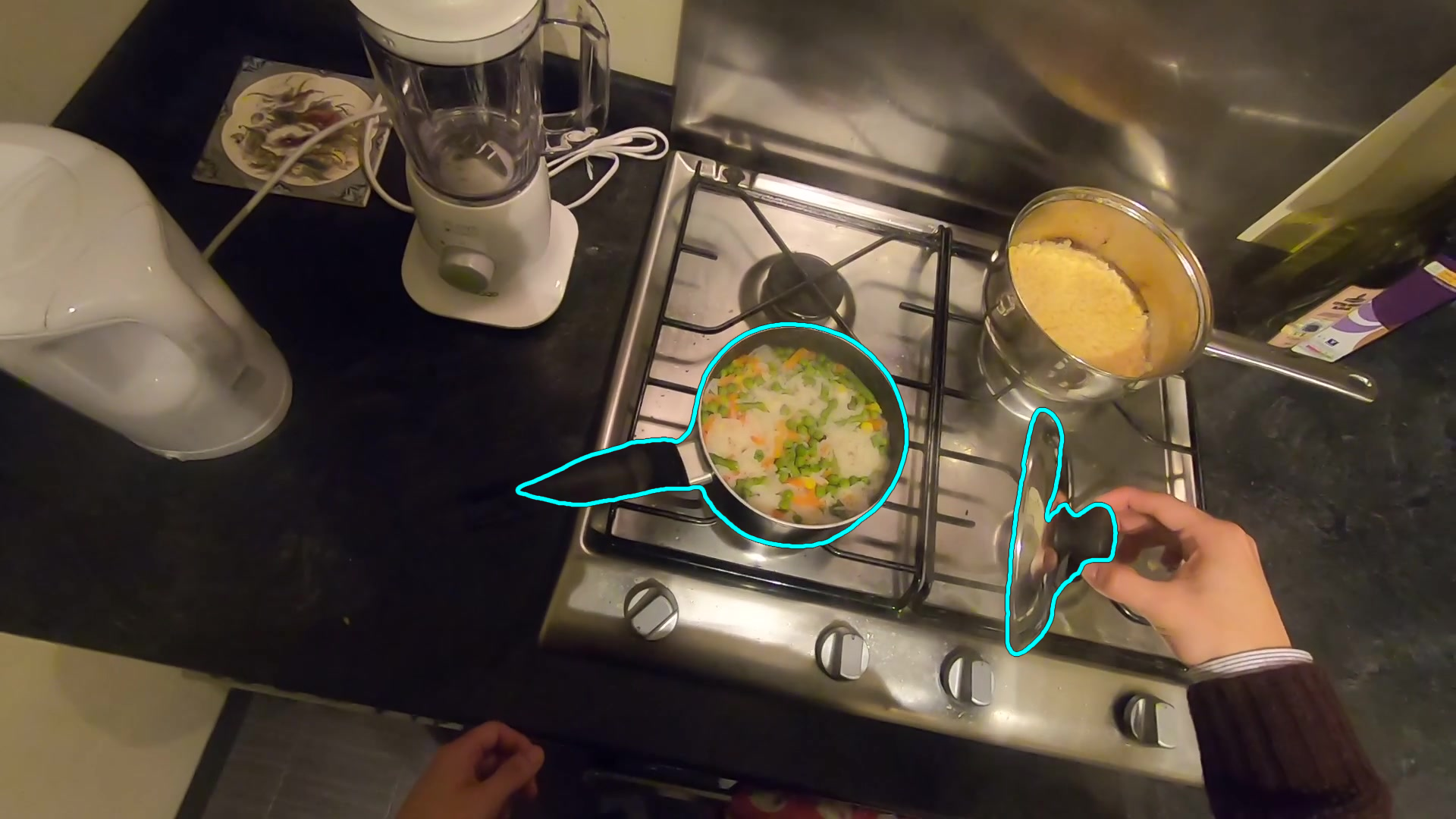} &
      \qualimg{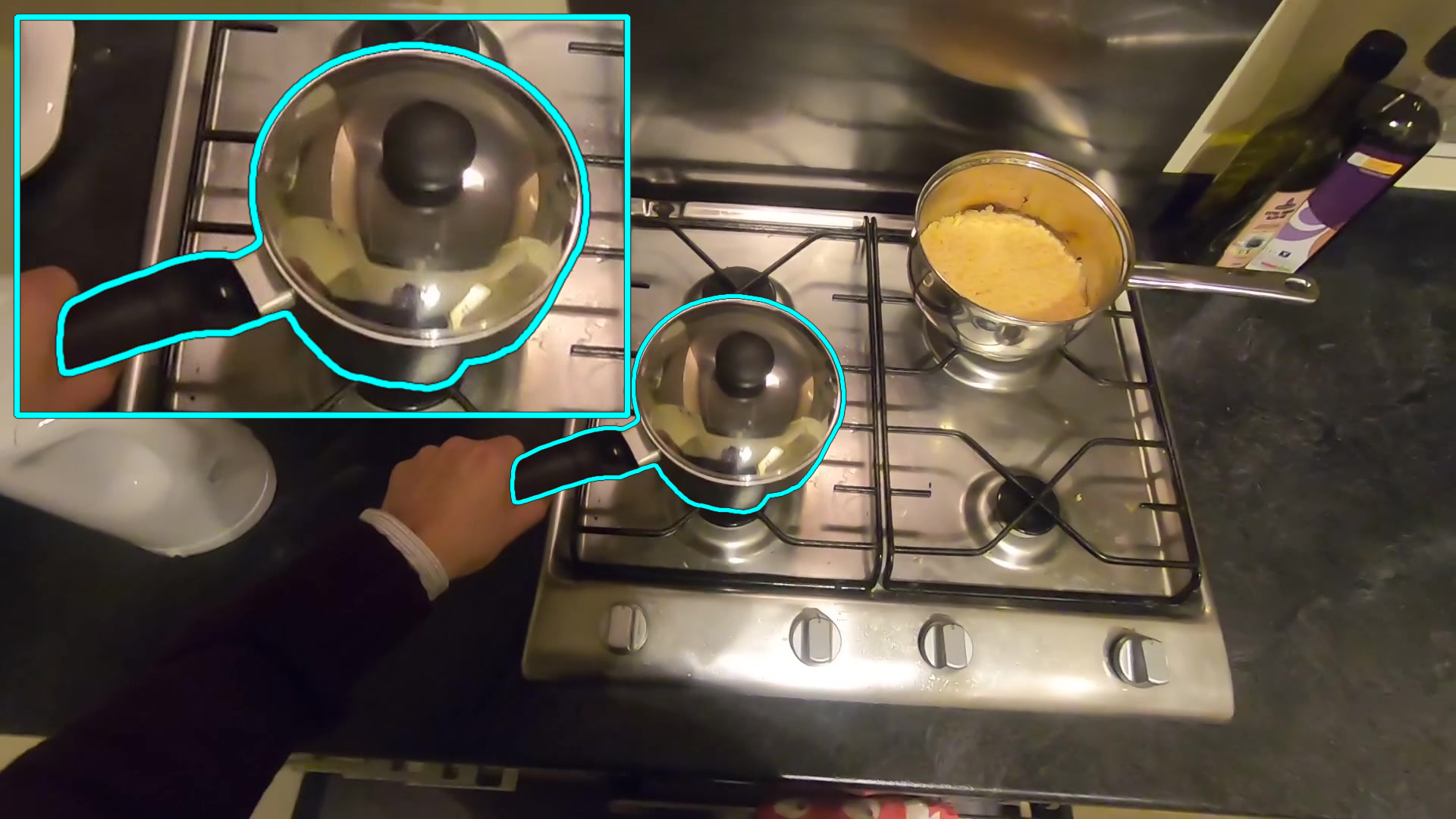} &
      \qualimg{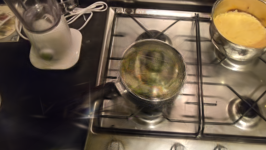} &
      \qualimg{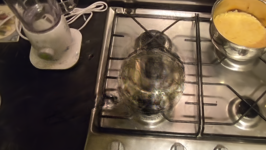} &
      \qualimg{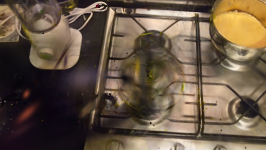} &
      \qualimg{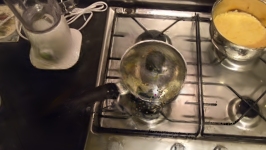}
      \\%
      \end{tabular}
    \caption{\textbf{Qualitative comparison} on TopoEgo. Different events shown (left) with changes highlighted in cyan in pre and post GT frames.
}
    \label{fig:qualitative_main}
\end{figure*}

\paragraph{Baselines.}
We compare Topo-4DGS with four Gaussian-splatting methods. EgoGaussian~\cite{zhang2024egogaussian}, our closest egocentric baseline, provides the rigid object-tracking backbone on which our method builds. DeGauss~\cite{wang2025degauss} combines grid-based deformation with static--dynamic decomposition. Def-3DGS~\cite{yang2024deformable3dgs} models motion using an MLP deformation field. Finally, 4DGS~\cite{wu2024fourdgs} uses a HexPlane-based representation. Topo-4DGS and EgoGaussian~\cite{zhang2024egogaussian} exclude hand pixels from the training loss. For fair comparisons, we additionally evaluate hand-masked variants of DeGauss~\cite{wang2025degauss}, Def-3DGS~\cite{yang2024deformable3dgs}, and 4DGS~\cite{wu2024fourdgs} by setting the photometric-loss weight to zero at hand pixels while keeping all other settings unchanged.
We additionally compare with HyperNeRF~\cite{park2021hypernerf}, the closest topology-varying neural-field baseline, on four short sequences.
We omit TED-4DGS~\cite{ho2025ted4dgs} and DP-DeGauss~\cite{chen2026dpdegaussdynamicprobabilisticgaussian} due to the lack of a public implementation.

\paragraph{Implementation details.}
Importantly, we use the same Topo-4DGS architecture, optimization schedule, and hyperparameter configuration across all 24 sequences and event families; only the input data and topology specification—including the number of objects, piece count $K$, event tree, event family, and event times—vary. %

\subsection{Main Results}
\label{sec:main_results}

\paragraph{Performance across topological changes.}
Table~\ref{tab:final_primary} reports object-region PSNR for each source--event-family group. Topo-4DGS ranks first in all eight groups, improving over the strongest baseline in each group by up to $1.26$\,dB. The baseline ordering helps separate the two sources of difficulty. EgoGaussian is the strongest baseline on the rigid objects without topological changes (rightmost, control columns), consistent with the benefits of its egocentric-specific object decomposition and rigid tracking~\cite{zhang2024egogaussian}. On topology-changing clips, however, its single-rigid-object assumption becomes restrictive. Def-3DGS~\cite{yang2024deformable3dgs} is the strongest competitor, especially its variant without actor hands. This suggests that flexible deformation fields can partially accommodate event-induced motion, despite the challenges of egocentric videos, but still provide neither explicit motion branching nor event-conditioned surface support. The largest gains of Topo-4DGS occur for separation events (\textit{open} and \textit{fission}), where independent part motion and newly created or exposed surfaces must be represented. Instead, \textit{fusion} and \textit{close} margins are consistent with surface concealment being easier to approximate than surface creation or exposure. By combining egocentric object modeling with explicit topology-aware mechanisms, Topo-4DGS remains best across the board.

On the 6 constant-topology control clips (rightmost columns in Tab.~\ref{tab:final_primary}), Topo-4DGS also obtains the best scores, showing that the topology-aware representation does not introduce a quality trade-off in standard settings.

\paragraph{Aggregate and post-event performance.}
Table~\ref{tab:final_complementary} evaluates four complementary regions: the complete scene with static background (\textit{Full Scene}), the object undergoing the topological event throughout the clip (\textit{Topo. Object}) and only after the event (\textit{Post-Event Topo. Object}), and other dynamic rigid objects (\textit{Other Dyn. Objects}). Topo-4DGS leads again across the board, with $+0.92$\,dB PSNR on the topological objects and a \textbf{$\textbf{+4.47}$\,dB PSNR} on the most challenging post-event frames, when new part motions and surface support must be represented. Topo-4DGS also outperforms on static and rigid areas, suggesting that resolving the topology-changing object improves joint scene composition without compromising the surrounding content.

Prior work shows that NeRF-based reconstruction degrades on egocentric footage with fast motion, dynamic hand--object interactions, and uneven viewpoint coverage~\cite{tschernezki2023epicfields,zhang2024egogaussian}.
Consistent with these findings, on four short egocentric sequences, Topo-4DGS outperforms HyperNeRF~\cite{park2021hypernerf} on every full-scene and object metric (e.g., 31.86 vs.\ 21.11\,dB full-scene PSNR); complete results in the Supp.

\begin{figure}[t]
    \centering
    \includegraphics[width=\columnwidth]
        {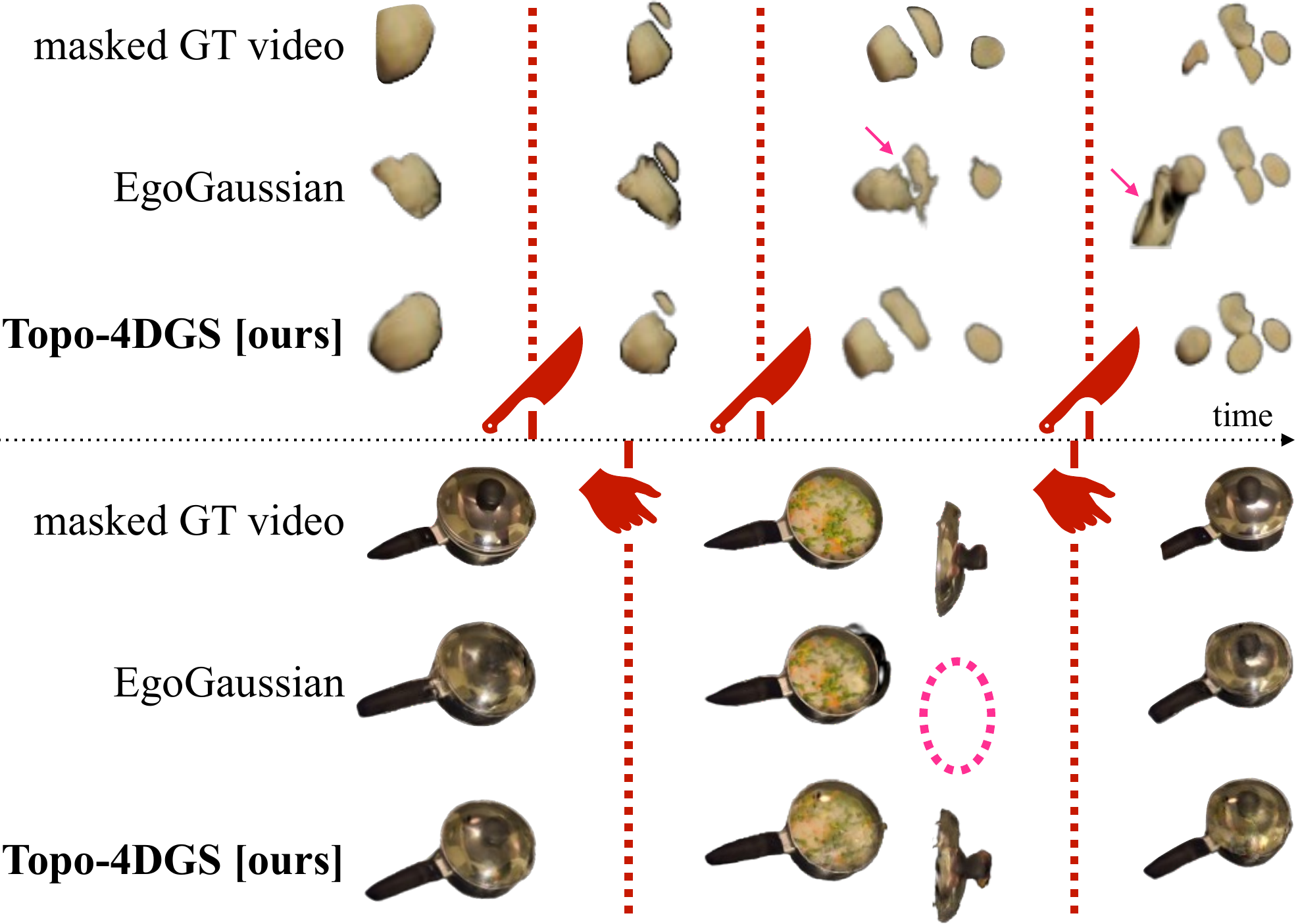}
    \caption{
        \textbf{Object-only reconstruction across repeated topological changes.} Comparison with EgoGaussian~\cite{zhang2024egogaussian}, the only baseline that also supports reconstructing objects alone. Above, three successive \textit{fission} events. Below, \textit{open} and \textit{close}.
    }
    \vspace{-10pt}
    \label{fig:potato_timeline}
\end{figure}

\paragraph{Qualitative results. }
\label{sec:qualitative_results}
Figure~\ref{fig:qualitative_main} shows that Topo-4DGS better preserves part geometry and changing surface support across all events, whereas baselines blur or merge independently moving parts and struggle with newly exposed or concealed surfaces. This is aligned with the results in Tabs.~\ref{tab:final_primary} and \ref{tab:final_complementary}. Figure~\ref{fig:potato_timeline} further demonstrates consistent reconstruction across repeated topological changes: Topo-4DGS preserves independently moving pieces through successive cuts and handles interior activation during opening and closing, while EgoGaussian produces connected, missing, or distorted geometry. Additional results are in Supp.

\subsection{Ablation Studies}
\label{sec:ablations}

\begin{table}[b]
\centering

\centering
\small
\setlength{\tabcolsep}{2pt}

\caption{\textbf{Topo-4DGS components.}
MO = Multiple Objects, MB = Motion Bifurcation,
BC = Per-piece Boundary Construction, and TA = Event-Conditioned Temporal Activation. Together, BC and TA constitute Topological Boundary Resolution.}

\begin{tabular}{cccc ccc ccc}

\toprule
\multicolumn{4}{c}{Config}
& \multicolumn{3}{c}{Topo. Object}
& \multicolumn{3}{c}{Post-Event} \\
\cmidrule(lr){1-4}
\cmidrule(lr){5-7}
\cmidrule(lr){8-10}

MO & MB & BC & TA
& PSNR & SSIM & LPIPS
& PSNR & SSIM & LPIPS \\
\midrule

\xmark & \xmark & \xmark & \xmark
& 21.11 & 0.664 & 0.278 & 18.45 & 0.579 & 0.286 \\

\cmark & \xmark & \xmark & \xmark
& 21.72 & 0.681 & 0.270 & 19.07 & 0.607 & 0.274 \\

\cmark & \cmark & \xmark & \xmark
& 23.84 & 0.745 & 0.190
& 23.05 & 0.698 & 0.222 \\

\cmark & \cmark & \cmark & \xmark
& 23.92 & 0.755 & 0.179
& 23.27 & 0.709 & 0.199 \\

\cmark & \cmark & \cmark & \cmark
& \textbf{25.32} & \textbf{0.786} & \textbf{0.158}
& \textbf{26.77} & \textbf{0.780} & \textbf{0.151} \\

\bottomrule
\end{tabular}%

\label{tab:abl_method}

\end{table}
\begin{table}[t]
\centering
\small
\setlength{\tabcolsep}{4.2pt}
\caption{\textbf{Topo-4DGS design choices.} Each variant changes one component of Topo-4DGS: UT-MLP replaces the event-gated temporal term with an MLP; Fixed-KM freezes the initial $k$-means partition in the spatial branch; and w/o TR, JSR, or SR removes Time Refinement, Joint Scene Refinement, or Spatial Regularization $\mathcal{L}_{sp}$, respectively.} 
\label{tab:abl_alternatives}
\begin{tabular}{l ccc ccc}
\toprule
& \multicolumn{3}{c}{Topo. Object}
& \multicolumn{3}{c}{Post-Event} \\
\cmidrule(lr){2-4}
\cmidrule(lr){5-7}
Variant
& PSNR & SSIM & LPIPS
& PSNR & SSIM & LPIPS \\
\midrule
UT-MLP
& 23.18 & 0.701 & 0.270
& 19.41 & 0.622 & 0.277 \\
Fixed-KM
& 22.95 & 0.703 & 0.267
& 19.59 & 0.633 & 0.263 \\
w/o TR
& 21.60 & 0.662 & 0.274
& 18.26 & 0.596 & 0.275 \\
w/o JSR
& 21.61 & 0.665 & 0.277
& 18.57 & 0.607 & 0.272 \\
w/o SR
& 23.25 & 0.693 & 0.293
& 20.80 & 0.628 & 0.293 \\
\midrule
Topo-4DGS
& \textbf{25.32} & \textbf{0.786} & \textbf{0.158}
& \textbf{26.77} & \textbf{0.780} & \textbf{0.151} \\
\bottomrule
\end{tabular}
\end{table}

\paragraph{Method components.}
Table~\ref{tab:abl_method} incrementally builds Topo-4DGS from a single-rigid-object baseline, equivalent to EgoGaussian. Handling multiple dynamic objects (MO) helps, e.g., rendering the knife in Fig.~\ref{fig:teaser}. By modeling independent part motions, Motion Bifurcation (MB, \cref{sec:branching}) substantially improves throughout, even by $+3.98$\,dB PSNR post-event. Per-piece Boundary Construction (BC, \cref{PC}) yields smaller but consistent improvements, likely because contact surfaces occupy a relatively small image region. Finally, Event-Conditioned Temporal Activation (TA, \cref{temporal_gate}) provides a further significant boost, demonstrating that surfaces must also appear and disappear at the appropriate times to yield high-quality renderings.

\paragraph{Alternative design choices.}
Table~\ref{tab:abl_alternatives} evaluates alternative designs for individual Topo-4DGS components. The impact of these changes is particularly noticeable on the more challenging post-event metrics, which reflect the inability to represent the transformation. Replacing the event-gated temporal term with an MLP (UT-MLP, \cref{UTMLP}) drastically reduces post-event PSNR by over 7\,dB, indicating that arbitrary frame-to-frame routing changes can introduce inconsistent motion. Freezing the initial $k$-means partition (Fixed-KM, \cref{Kmeans}) instead of learning the Spatial Branch yields a similar catastrophic drop, showcasing the impact of the introduced optimization. Removing Time Refinement (TR, \cref{sec:learned_timers}) or Joint Scene Refinement (JSR, \cref{sec:joint_scene}) also substantially drops the performance, highlighting the importance of correcting activation times and optimizing the composed scene altogether. Finally, removing Spatial Regularization (SR, \cref{eq:spatial_regularizer}) confirms that spatial coherence prevents fragmented routing.

Additional analyses of piece count, event-time error, input frame rate, runtime, and resource use are provided in Supp.

\section{Conclusion}
\label{sec:conclusion}
In this work, we introduced the problem of reconstructing egocentric scenes with topologically changing objects. For this, we introduced TopoEgo as the first attempt to benchmark it, and Topo-4DGS, a topology-aware Gaussian representation that jointly models event-conditioned part motion and changing surface support. Across various transformation types, Topo-4DGS consistently outperforms egocentric and dynamic reconstruction methods.
The method currently assumes reliable event specifications and approximately piecewise-rigid parts. Extending it to automatic topology discovery and strongly deformable objects remains important future work.

\section*{Acknowledgements}
We acknowledge the CINECA award under the ISCRA initiative and the Bocconi University HPC infrastructure for the availability of high performance computing resources and support.

\bibliographystyle{ieeenat_fullname}
\bibliography{references}

\clearpage
\maketitlesupplementary

\appendix

\section*{Overview}

This supplementary material provides the benchmark construction, method details, implementation specifications, and extended evaluation omitted from the main paper.  \Cref{app:dataset} documents TopoEgo construction and annotation, while \cref{app:auto_discovery} presents automatic topology discovery. The table below maps each supplementary pointer in the main paper to the corresponding section or table.
\newcolumntype{Y}{>{\raggedright\arraybackslash}X}

\begin{table}[h]
    \centering
    \caption{\textbf{Guide to the supplementary material.}
    Correspondence between sections of the main paper and the extended
    material provided in this document.}
    \label{tab:supplement_guide}

    \footnotesize
    \setlength{\tabcolsep}{2.5pt}
    \renewcommand{\arraystretch}{1.15}

    \begin{tabularx}{\columnwidth}{
        @{}
        >{\raggedright\arraybackslash}p{0.25\columnwidth}
        >{\raggedright\arraybackslash}X
        >{\raggedright\arraybackslash}p{0.17\columnwidth}
        @{}
    }
        \toprule
        Main paper section & Extended material & Sec. \\
        \midrule

        Sec.~5.1\newline
        \emph{Motion Bifurcation}
        & Spatial and Temporal Branch normalization and parameterization
        & \cref{app:event_state_parameterization} \\

        Sec.~5.1\newline
        \emph{Motion Bifurcation}
        & Spatial Branch initialization and learned partition
        & \cref{app:learned_partition} \\

        Sec.~5.3\newline
        \emph{Optimization}
        & Objectives, masking, and spatial regularization
        & \cref{optimization} \\

        Sec.~5.2\newline
        \emph{Topological Boundary Resolution}
        & Time Refinement
        & \cref{app:boundary_timers} \\

        Sec.~5.3\newline
        \emph{Optimization}
        & Missing-pose interpolation and its ablation
        & \cref{app:pose_interpolation,tab:interpolation_ablation} \\

        Sec.~5.2\newline
        \emph{Topological Boundary Resolution}
        & Boundary construction and branch binding
        & \cref{app:boundary_construction} \\

        Sec.~6.2\newline
        \emph{Ablation Studies}
        & Sensitivity to terminal-piece count, event time, and input frame rate
        & \cref{sec:supp_robustness} \\

        Sec.~6.2\newline
        \emph{Ablation Studies}
        & Runtime and resource usage
        & \cref{app:runtime} \\

        Sec.~6.1\newline
        \emph{Main Results}
        & Complete HyperNeRF comparison
        & \cref{tab:final_hypernerf} \\

        Sec.~6.1\newline
        \emph{Main Results}
        & Additional qualitative results
        & \cref{app:qualitative} \\

        \bottomrule
    \end{tabularx}
\end{table}

\section{Video Comparisons}
\label{app:video_comparisons}

Single-frame comparisons do not fully expose temporal artifacts such as flickering, inconsistent motion, premature or delayed structural changes, and parts that remain incorrectly attached after an event. The project page (\url{https://topoego.github.io}) therefore provides video comparisons for eleven TopoEgo sequences, six from EPIC-KITCHENS and five from Ego4D, covering all four event families. Each video shows, side by side and synchronized, the ground-truth observations and the reconstructions from Topo-4DGS, hand-masked Def-3DGS, EgoGaussian, hand-masked 4DGS, and hand-masked DeGauss. Within each sequence, all panels use the same frame range, camera trajectory, spatial resolution, and playback rate.

Topo-4DGS and EgoGaussian exclude hand pixels as part of their standard training protocols. For these video comparisons, we apply the same exclusion to Def-3DGS, 4DGS, and DeGauss, providing a controlled comparison in which reconstructed hand geometry does not obscure object motion and changing surface support. Because these transitions often occur within only a few
frames, playback slows around each annotated event and then returns to native speed. Sequences with particularly rapid native motion are instead shown at a reduced rate throughout.

Each video covers the event and its surrounding pre- and post-event intervals. The videos allow direct assessment of whether pieces begin moving independently after \textit{fission}, remain coherently assembled after \textit{fusion}, and correctly expose or conceal interior support during \textit{open} and \textit{close} events.

\paragraph{Representative observations.}
Sequences are referred to by the clip identifiers shown on the project page.

\textit{Fission.}
In \texttt{P06\_110}, Topo-4DGS maintains the two pizza portions as distinct bodies after the cut. EgoGaussian represents the separation only intermittently, while 4DGS produces elongated artifacts around the separated region. In \texttt{P02\_122}, EgoGaussian continues to render an intact egg after the shell has broken: the reconstruction remains stable but retains the pre-event topology.

\textit{Fusion.}
In \texttt{P02\_108}, Topo-4DGS maintains a fixed relative pose between the spinner basket and bowl after contact. Def-3DGS leaves a persistent artifact across their seam, while EgoGaussian remains blurred throughout. In \texttt{8c335aa9}, Topo-4DGS integrates the newly placed plank into the floor, whereas Def-3DGS, 4DGS, and DeGauss retain streaking or distorted geometry around the attachment region.

\textit{Open.}
In \texttt{P30\_110}, Topo-4DGS exposes the container interior and continues to represent the lid as a separate body. EgoGaussian retains the lid only as a faint translucent region, while 4DGS produces residual geometry that does not fully detach.

\textit{Close.}
In \texttt{200a8}, Topo-4DGS conceals the container interior at the annotated event. In contrast, 4DGS removes it gradually and leaves dark artifacts around the rim after closure.

The slowed intervals show that these differences persist over time rather than occurring only in selected frames. Their structural and temporal character can be obscured by image metrics averaged over the complete sequence.

\begin{table}[t]
\caption{\textbf{Prevalence of the target event families in the source annotations.} Counts are obtained from Ego4D state-transition annotations and EPIC-KITCHENS timestamped action segments. Coverage is the fraction of source videos containing at least one annotation associated with the corresponding event family; a video may contribute to multiple families. Retrieved interactions are manually screened for a visible topological event as described in \cref{sec:clip_selection}.}
\label{tab:prevalence}
\centering
\footnotesize
\setlength{\tabcolsep}{4.2pt}
\renewcommand{\arraystretch}{1.1}
\begin{tabular}{l ccc}
\toprule
Category & 
    \rotatebox{90}{\#Actions} & 
    \rotatebox{90}{\#Videos} & 
    \rotatebox{90}{Coverage} \\
\midrule
\multicolumn{4}{l}{\emph{\textbf{Ego4D} ($1{,}725$ videos)}}\\
Topological (four target families) & $16{,}132$ & $1{,}187$ & $68.8\%$ \\
\quad \textit{fission} ($1\!\to\!N$: cut/break/tear) & $3{,}417$ & $567$ & $32.9\%$ \\
\quad \textit{fusion} ($N\!\to\!1$: attach/sew/weld) & $4{,}210$ & $743$ & $43.1\%$ \\
\quad \textit{open} (separable containers) & $6{,}410$ & $972$ & $56.3\%$ \\
\quad \textit{close} (separable containers) & $2{,}095$ & $600$ & $34.8\%$ \\
\midrule
\multicolumn{4}{l}{\emph{\textbf{EPIC-KITCHENS-100} ($495$ annotated videos)}}\\
Topological (four target families) & $5{,}176$ & $396$ & $80.0\%$ \\
\quad \textit{fission} (cut/chop/peel/grate/disassemble) & $2{,}618$ & $213$ & $43.0\%$ \\
\quad \textit{fusion} (attach/screw/tie) & $88$ & $51$ & $10.3\%$ \\
\quad \textit{open} (separable containers) & $1{,}428$ & $349$ & $70.5\%$ \\
\quad \textit{close} (separable containers) & $1{,}042$ & $292$ & $59.0\%$ \\
\bottomrule
\end{tabular}
\end{table}
\section{TopoEgo Benchmark Construction}
\label{app:dataset}

TopoEgo is constructed from EPIC-KITCHENS~\cite{damen2022epickitchens} and Ego4D~\cite{grauman2022ego4d}. The benchmark distributes derived annotations and preprocessing assets rather than the source videos.

We define a \emph{topological event} as a temporally localized manipulation that changes an object's part connectivity, motion structure, or surface support. This operational definition includes separable opening and closing, even when the underlying material remains partially connected, because these events change which boundary surfaces must be represented.

\begin{table*}[t]
    \centering
    \caption{\textbf{TopoEgo event taxonomy.}
    Events are categorized by object-part motion and their effect on the
    surface support represented by the model. \emph{Open} and \emph{close}
    apply to separable assemblies and need not alter material connectivity.}
    \label{tab:dataset_taxonomy}

    \small
    \setlength{\tabcolsep}{7pt}
    \renewcommand{\arraystretch}{1.15}

    \begin{tabularx}{\textwidth}{
        @{}
        >{\bfseries}l
        >{\hsize=0.85\hsize}Y
        >{\hsize=1.15\hsize}Y
        @{}
    }
        \toprule
        Event family
        & Inclusion criterion
        & Effect on represented surface support \\
        \midrule

        Fission
        & A manipulated object separates into independently moving pieces
          through cutting, fracture, tearing, detachment, or disassembly.
        & Newly exposed cut or separation surfaces may become visible and
          require representation. \\

        Fusion
        & Independently moving pieces are joined into a single coupled object.
        & Contacting surfaces become sealed or occluded and may be removed
          from the active representation. \\

        Open
        & Parts of a separable assembly move apart, exposing an interior
          (e.g., removing a lid or opening a container).
        & Previously concealed interior surfaces become visible and enter the
          active representation. \\

        Close
        & Parts of a separable assembly are rejoined, concealing its interior.
        & Interior and contact surfaces become concealed and leave the active
          representation. \\

        \bottomrule
    \end{tabularx}
\end{table*}

\subsection{Corpus Analysis and Candidate Mining}

Table~\ref{tab:prevalence} estimates prevalence from the structured temporal annotations provided by EPIC-KITCHENS~\cite{damen2022epickitchens} and Ego4D~\cite{grauman2022ego4d}. The two datasets use different annotation formats.

EPIC-KITCHENS represents an action as a temporally bounded segment. Each annotation specifies the start and end of the action, both as timestamps and frame indices, together with a narration and parsed verb and noun labels. For example, an action such as \textit{open door} is associated with a specific temporal interval and the labels \textit{open} and \textit{door}. We use the annotated action interval as a retrieval window and the verb label to identify interactions potentially belonging to our taxonomy.

Ego4D Focused Hand--Object (FHO) annotations instead describe individual
hand--object interactions using critical frames before, during, and after an
object state change. In particular, they provide a pre-condition frame, a
point-of-no-return frame marking when the object state change becomes
irreversible, and a post-condition frame, together with the state-change type
and annotations of the manipulated objects, tools, and hands. We use these state-transition annotations to retrieve candidate interactions.

We map only annotations associated with the four target families to the taxonomy in Table~\ref{tab:dataset_taxonomy}. In Ego4D, potential \textit{fission} events are retrieved from FHO deconstruct transitions (including disassembly and fracture subcases), potential \textit{fusion} events from construct, and relevant activate/deactivate transitions provide \textit{open} and \textit{close} candidates. In EPIC-KITCHENS, we retrieve action segments labeled cut, chop, slice, peel, grate, crack, break, snap and disassemble for \textit{fission}; attach, screw, and tie for \textit{fusion}; open/uncover for \textit{open}; and close/cover for \textit{close}.

These source annotations are used only to retrieve potential interactions:
neither an EPIC-KITCHENS action label nor an Ego4D state-transition annotation
is treated directly as a TopoEgo ground-truth event. Every retrieved
interaction is subsequently inspected manually following the clip-selection and annotation procedure described in \cref{sec:clip_selection}.

\subsection{Clip Selection and Manual Annotation}
\label{sec:clip_selection}

We review every retrieved interaction at its native frame rate. We retain clips with a visible event, an identifiable manipulated object, and sufficient pre- and post-event context, excluding off-frame, fully occluded, unrealized, or non-topological interactions. The final manifest is balanced across source datasets and event families. Table~\ref{tab:topoego_stats} summarizes its composition; the released manifest provides the complete per-sequence breakdown.

\begin{table}[t]
\caption{\textbf{TopoEgo benchmark composition.} Each event family contains 3 clips from EPIC-KITCHENS and 3 from Ego4D. A clip may contain multiple annotated dynamic objects and topological events, so the object and event totals may exceed the number of clips.}
\label{tab:topoego_stats}
\centering
\footnotesize
\setlength{\tabcolsep}{3.5pt}
\renewcommand{\arraystretch}{1.1}

\begin{tabular}{lccccc}
\toprule
& \multicolumn{2}{c}{Clips} & & & \\
\cmidrule(lr){2-3}
Event family
& EPIC-K.
& Ego4D
& Frames
& Objects
& Events \\
\midrule

\textit{fission}
& 3 & 3 & 1,922 & 12 & 11 \\

\textit{fusion}
& 3 & 3 & 939 & 8 & 7 \\

\textit{open}
& 3 & 3 & 810 & 6 & 7 \\

\textit{close}
& 3 & 3 & 947 & 10 & 9 \\

\midrule

\textbf{Total}
& \textbf{12}
& \textbf{12}
& \textbf{4,618}
& \textbf{36}
& \textbf{34} \\

\bottomrule
\end{tabular}
\end{table}

\begin{figure*}[t]
    \centering
    \includegraphics[width=0.88\textwidth]{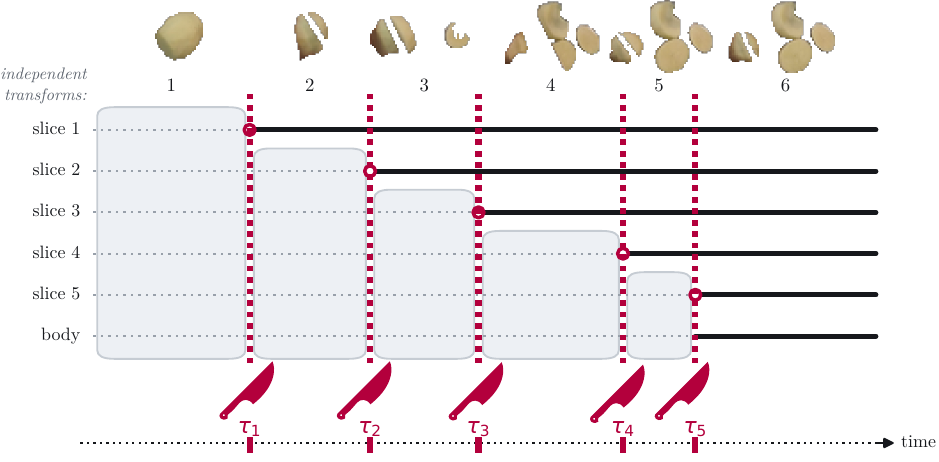}
    \caption{\textbf{Example event-tree annotation.} A potato undergoes 5 successive cuts, with its observed states shown above the timeline. At each event time $\tau_m$ (red dashed line), one slice separates from the remaining body and begins following an independent rigid transform. Gray blocks group parts that still share a transform. The event-tree leaves are the terminal pieces: the 5 slices and the remaining body (i.e., terminal-piece count K=6).}
    \label{fig:event_tree_example}
\end{figure*}

All final labels were set manually and constitute the ground truth for the reconstruction experiments. For each topological event, the annotations specify the event family and event time. For each manipulated object, they specify the terminal-piece count K and, whenever motion branching is required, an event tree linking that object’s events to its terminal pieces. Figure~\ref{fig:event_tree_example} illustrates this annotation for a sequence containing multiple successive events.
Each event connects a parent component to the child components released at that event. A child that later separates becomes an internal node of the tree; only nodes with no children after all events are resolved are leaves, which we call \emph{terminal pieces}.

For each selected interaction, we define a \emph{clip} as the contiguous sequence of source-video frames retained for the benchmark, including context before and after the topological event. The manifest preserves the original frame range in the source video. After extraction, we renumber the frames of every TopoEgo clip consecutively starting from 0, so that frame 0 is always the first frame of the extracted clip, regardless of the source dataset. The source dataset, original frame range, and native frame rate are retained in the manifest so that the clip can be mapped back to the original video unambiguously.

The event time is annotated as the first clip frame in which the post-event state is visible: independently moving parts for \textit{fission}, a revealed interior for \textit{open}, stable assembly contact for \textit{fusion}, or a concealed interior for \textit{close}.

\subsection{Distributed Assets and Use}

Each manifest entry provides a stable clip identifier, source-video identifier, dataset name, original start and stop frames, native frame rate, event annotations, object identifiers, and the source action used for retrieval. The distributed assets also include camera intrinsics and extrinsics, object and hand masks, topology-configuration files, preprocessing scripts, and evaluation scripts. We use camera trajectories from EPIC Fields~\cite{tschernezki2023epicfields} when their temporal coverage is sufficient and estimate the remaining trajectories using COLMAP~\cite{schoenberger2016colmap}. Object and hand masks are generated using SAM~3~\cite{carion2026sam3segmentconcepts} and the released segmentation preprocessing. All generated masks are manually inspected before use. These poses and masks provide standardized reconstruction inputs, while the manual event annotations define the benchmark ground truth.

Researchers should obtain raw EPIC-KITCHENS and Ego4D footage from the original providers using the source identifiers in the manifest. We will distribute only annotations, derived metadata, scripts, and benchmark manifests, without redistributing the raw videos. This preserves the source datasets' licensing, consent, and privacy conditions and avoids duplicating identifiable first-person footage outside their established access procedures.

\section{Automatic Topology Discovery}
\label{app:auto_discovery}

All reconstruction results in the main paper use the manually reviewed topology annotations provided by TopoEgo.
The main paper's conclusion identifies automatic topology discovery as future work, and this section does not change that: the module below is a standalone preprocessing tool, not a component of Topo-4DGS. For videos without annotations, we provide an optional preprocessing module that automatically estimates the inputs required by Topo-4DGS: the terminal-piece count $K$, event family, event tree, and ordered event times. The module takes as input the RGB video, per-frame object masks, and the frame range containing the interaction for each manipulated object

\paragraph{Semantic proposal.}
We uniformly sample 12 frames showing the masked object and provide them, together with their frame indices, to Qwen2.5-VL-7B-Instruct~\cite{bai2025qwen25vltechnicalreport}, a Vision-Language Model (VLM). A three-stage prompting procedure identifies the manipulated object and interacting tools, classifies the event as \emph{fission}, \emph{fusion}, \emph{open}, \emph{close}, \emph{mixed}, or \emph{none}, and estimates the initial and final piece counts and coarse event times. We use greedy decoding with a maximum of 800 generated tokens. The complete prompts will be released with the code.

For rigid tools, the prompt defaults to no topological event unless a structural change is clearly visible. For other objects, we set $K$ to the larger of the predicted initial and final piece counts and enforce $K\geq2$ whenever an event is predicted. Finally, we sort the predicted events by time and convert them into the parent--child representation used by Topo-4DGS. Predictions labeled as mixed, as well as predictions that cannot be converted into a consistent parent--child hierarchy, are flagged for manual review.

\paragraph{Event-time refinement.}
We refine each coarse VLM event time in two stages. First, we perform a coarse-to-fine VLM search around the predicted event. We sample frames within a temporal window centered on the current estimate, query the VLM again, and then repeat the procedure using a narrower window and more closely spaced frames. This progressively restricts the event to a short temporal interval.

We then refine the boundary using object motion. CoTracker3-offline~\cite{karaev2024cotracker3simplerbetterpoint} is run on the full, uncropped RGB frames surrounding the VLM estimate. Tracking points are initialized from pixels inside the manipulated-object mask; these are tracking seeds rather than separately detected semantic keypoints. We discard unreliable tracks and tracks dominated by regions labeled as hands.

For each frame, we measure how much the relative arrangement of the remaining object points has changed compared with the pre-event frames. We compute pairwise image-plane distances between tracked points, normalize them by object size, and compare them with their median pre-event values. For each point, we take the median change relative to all other points, and use the 85th percentile across points as a frame-level motion-divergence score. Using an upper percentile makes the score sensitive when only a small part of the object separates from the rest.

For \textit{fission} and \textit{open}, the first sustained increase in this score is taken as the refined event time. \textit{Fusion} and \textit{close} use the same procedure after reversing the sequence in time. This refinement changes only the event time: the predicted event family, terminal-piece count, and event tree remain unchanged. If tracking is unreliable or no sustained change is detected, we retain the VLM estimate.

\paragraph{Automatic-discovery accuracy.}
We evaluate the automatically generated topology specifications against the manually reviewed TopoEgo annotations. Of the 36 annotated dynamic objects, the module predicts the correct terminal-piece count in $26/36$ cases ($72.2\%$) and the correct object-level event-family annotation for 16/24 topological objects ($66.7\%$). Of the 34 ground-truth events, 24 receive a matched prediction, with a median absolute timing error of 16 frames. The remaining 10 events are unmatched because no event is predicted for their corresponding objects.

These results evaluate the topology-discovery outputs themselves. All benchmark reconstruction results use the manually reviewed annotations. We separately evaluate reconstruction sensitivity to automatically predicted event times in \cref{sec:supp_robustness}.

\section{Additional Details of Motion Bifurcation and the Spacetime Clustering Field}
\label{app:spacetime_clustering_math}

Sec.~5.1 of the main paper introduces \emph{Motion Bifurcation}, which assigns the Gaussians of a manipulated object to event-dependent piecewise-rigid motion branches. The corresponding \emph{Spacetime Clustering Field} computes these routing assignments by combining a time-independent Spatial Branch with a Temporal Branch driven by the annotated event times.

This section provides the formulation and implementation details omitted from the main paper. \Cref{app:multi_object_routing} extends the single-object notation of Sec.~5.1 to multiple manipulated objects, while \cref{app:routing_derivation} motivates restricting routing changes to event windows. \Cref{app:event_state_parameterization} specifies the Spatial and Temporal Branches, and \cref{app:learned_partition} details the $K$-means initialization used to warm-start the learned Spatial Branch. Finally, \Cref{optimization} reports the corresponding optimization details.

\paragraph{Notation recap.}
For one manipulated object, $\mathcal{G}_i$ denotes canonical Gaussian $i$ with
center $\boldsymbol{\mu}^{\mathrm{can}}_i$, and
$\mathcal{K}=\{1,\ldots,K\}$ indexes the terminal pieces. Piece $k$ follows
the rigid transform
$\mathbf{T}_k(t)=(\mathbf{R}_k(t),\mathbf{d}_k(t))$, while
$P_{i,k}(t)$ is the routing weight assigning Gaussian $i$ to that piece at
frame $t$. Annotated events are indexed by $m\in\{1,\ldots,M\}$ and occur at
times $\tau_m$. Object index $j$ is introduced only where the multi-object
formulation requires it. We use \emph{terminal piece} for a leaf of the event tree and
\emph{motion branch} for the rigid trajectory it follows; the two are in
one-to-one correspondence.

\subsection{Multi-Object Formulation and Notation}
\label{app:multi_object_routing}

For clarity, Sec.~5.1 of the main paper formulates Motion Bifurcation for a single manipulated object and suppresses the object index. Here we give the multi-object formulation used in the implementation.

Let the scene contain a static Gaussian set $\mathcal{G}_{\mathrm{bg}}$ and $J$ dynamic objects. We denote by $\mathcal{G}^{(j)}_{\mathrm{obj}}(t)$ the Gaussians belonging to dynamic object $j$ after applying its motion at frame $t$. The scene rendered at that frame is therefore
\begin{equation}
\mathcal{G}_{\mathrm{scene}}(t)=
\mathcal{G}_{\mathrm{bg}}\cup
\bigcup_{j=1}^{J}\mathcal{G}^{(j)}_{\mathrm{obj}}(t).
\end{equation}

For object $j$, let $K_j$ denote its number of terminal pieces, as defined in
Sec.~5.1 of the main paper. At each frame, the annotated event tree partitions
these pieces into groups that must share a rigid transform. We formalize this
constraint with a time-dependent representative map
$\rho_{j,t}:\{1,\ldots,K_j\}\rightarrow\{1,\ldots,K_j\}$.
The map assigns every terminal piece a common index for its current
event-tree group. Thus,
$\rho_{j,t}(k)=\rho_{j,t}(\ell)$ exactly when pieces $k$ and $\ell$ must share
a transform at time $t$; the numerical value used to identify the group has no
separate physical meaning. The grouping changes only at annotated events.

For canonical Gaussian $i$ of object $j$, let $\boldsymbol{\mu}^{\mathrm{can}}_{i,j}$ be its center and $\mathbf{L}^{\mathrm{can}}_{i,j}=\mathbf{R}^{\mathrm{can}}_{i,j} \operatorname{diag}(\mathbf{s}^{\mathrm{can}}_{i,j})$ its covariance factor. Terminal piece $k$ follows a motion branch with rigid transform $\mathbf{T}_{j,k}(t)=(\mathbf{R}_{j,k}(t),\mathbf{d}_{j,k}(t))\in\mathrm{SE}(3)$. Motion branches whose terminal pieces belong to the same group share transform
parameters. Let $P_{i,j,k}(t)$ denote the routing weight assigning Gaussian $i$ of object $j$ to terminal piece $k$ at frame $t$. We compute the rendered center and covariance factor as
\begin{align}
\boldsymbol{\mu}_{i,j}(t)
&=\sum_{k=1}^{K_j}P_{i,j,k}(t)
\left(
\mathbf{R}_{j,\rho_{j,t}(k)}(t)
\boldsymbol{\mu}^{\mathrm{can}}_{i,j}+
\mathbf{d}_{j,\rho_{j,t}(k)}(t)
\right), \\
\mathbf{L}_{i,j}(t)
&=\sum_{k=1}^{K_j}P_{i,j,k}(t)
\mathbf{R}_{j,\rho_{j,t}(k)}(t)
\mathbf{L}^{\mathrm{can}}_{i,j}, \\
\boldsymbol{\Sigma}_{i,j}(t)
&=\mathbf{L}_{i,j}(t)\mathbf{L}_{i,j}(t)^\top.
\end{align}
For a single object when every terminal piece is its own representative, $\rho_{j,t}$ is the identity and Eq.~(12) reduces to Eq.~(4) of the main paper. Eqs.~(13) and~(14) apply the same routing weights to the Gaussian covariance factor.
The static background and all dynamic objects are jointly rasterized. The same routing weights transport the center and covariance factor, so this operation softly mixes the piece-conditioned rigid motions without duplicating the canonical primitive. Below, we fix one object and omit its index $j$.

\subsection{Routing Changes Restricted to Event Windows}
\label{app:routing_derivation}

The event gates of Sec.~5.1 of the main paper (Eq.~(6)) restrict temporal changes in the Spacetime Clustering Field to neighborhoods of the annotated events. This design prevents the routing assignment itself from acting as an additional unconstrained motion model. Here we make this motivation explicit.

For one object, let $P_{i,k}(t)$ be the routing weight assigning Gaussian $i$ to terminal piece $k$, and let
\[
\mathbf{q}_{i,k}(t)
=
\mathbf{R}_k(t)\boldsymbol{\mu}^{\mathrm{can}}_i
+\mathbf{d}_k(t)
\]
be the position predicted for that Gaussian by the corresponding motion branch. As in Sec.~5.1 of the main paper, the rendered Gaussian center is the weighted combination \begin{equation}
\label{eq:softroute_app}
\begin{array}{c}
\boldsymbol{\mu}_i(t)=\sum_{k=1}^{K}P_{i,k}(t)\mathbf{q}_{i,k}(t).
\end{array}
\end{equation} 
Differentiating its routed center gives
\begin{align}
\dot{\boldsymbol{\mu}}_i(t)
&=
\underbrace{\sum_{k=1}^{K}P_{i,k}(t)\dot{\mathbf{q}}_{i,k}(t)}_{
\text{branch motion}}
+
\underbrace{\sum_{k=1}^{K}\dot{P}_{i,k}(t)\mathbf{q}_{i,k}(t)}_{
\text{assignment-induced motion}}.
\end{align}
Because the routing weights sum to one, $\sum_k\dot P_{i,k}(t)=0$. Choosing any branch $r$ as a reference, the second term can be written as
\begin{equation}
\label{eq:assignment_velocity}
\sum_{k=1}^{K}\dot P_{i,k}(t)
\bigl(\mathbf{q}_{i,k}(t)-\mathbf{q}_{i,r}(t)\bigr).
\end{equation}
The first term is motion produced by the motion branches themselves, whereas the second is motion caused solely by changing the routing weights. Once the corresponding motion branches have diverged, even a small change in $P_{i,k}(t)$ can therefore move a Gaussian substantially between branches. This motivates the event-gated Temporal Branch introduced in Sec.~5.1: routing is allowed to change around annotated events but remains approximately constant between them. The Unrestricted Temporal MLP (UT-MLP) ablation in Table~4 of the main paper tests the alternative of unrestricted time-dependent routing.

\subsection{Spatial and Temporal Branch Details}
\label{app:event_state_parameterization}

We now provide the implementation details of the Spatial and Temporal Branches introduced in Sec.~5.1 of the main paper. The Spatial Branch maps the canonical position of each object Gaussian to logits over the $K$ terminal pieces, while the Temporal Branch modifies these logits according to which annotated events have occurred.

We normalize the canonical centers using the object's bounding sphere:
\begin{equation}
\bar{\boldsymbol{\mu}}^{\mathrm{can}}_i
=
\frac{
\boldsymbol{\mu}^{\mathrm{can}}_i-\boldsymbol{\mu}_{\mathrm{obj}}
}{
r_{\mathrm{obj}}
},
\end{equation}
where
\begin{equation}
\boldsymbol{\mu}_{\mathrm{obj}}
=
\frac{1}{N_{\mathrm{obj}}}
\sum_{i=1}^{N_{\mathrm{obj}}}
\boldsymbol{\mu}^{\mathrm{can}}_i,
\ \ \
r_{\mathrm{obj}}
=
\max_{1\leq i\leq N_{\mathrm{obj}}}
\left\|
\boldsymbol{\mu}^{\mathrm{can}}_i-
\boldsymbol{\mu}_{\mathrm{obj}}
\right\|_2
\end{equation}
and $N_{\mathrm{obj}}$ is the number of canonical object Gaussians.
These normalization statistics are computed once before coarse object tracking and remain fixed thereafter. Here, $\bar{\boldsymbol{\mu}}^{\mathrm{can}}_i$ is the normalized canonical center used in Eq.~(5) of the main paper. It is passed through the Spatial Branch as
\begin{equation}
\mathbf{z}^{\mathrm{sp}}_i
=
h_{\phi}\!\left(
\gamma(\bar{\boldsymbol{\mu}}^{\mathrm{can}}_i)
\right)
\in\mathbb{R}^{K},
\end{equation}
where $\gamma$ is a four-band (L=4) sinusoidal encoding, $h_{\phi}$ is a three-layer MLP with two $64$-unit ReLU hidden layers, and the superscript $\mathrm{sp}$ indicates that $\mathbf{z}^{\mathrm{sp}}_i$ contains the $K$ logits produced by the Spatial Branch for Gaussian $i$.

Event gates are evaluated in an object-level normalized time coordinate. For a frame index $t$ and annotated event time $\tau_m$, we define
\begin{equation}
\bar t
=
\operatorname{clip}
\left(
\frac{t-c}{s},-4,4
\right),
\qquad
\bar\tau_m
=
\frac{\tau_m-c}{s},
\end{equation}
where \(c\) is the temporal center and \(s>0\) is the temporal scale. When they are not specified explicitly, $c$ is set to the first annotated event frame and $s$ to the duration of the annotated dynamic interval. The event gate is then
\begin{equation}
g_m(\bar t)
=
\sigma\!\left(
\frac{\bar t-\bar\tau_m}{\beta}
\right),
\qquad
\beta=0.25.
\end{equation}
where $\sigma(u)=(1+e^{-u})^{-1}$ is the logistic sigmoid.
We use the same fixed, dimensionless transition width $\beta$ for every event and object. Neither the event times nor the gate parameters are optimized. For $M$ annotated events, we collect these gates into
\[
\mathbf{g}(\bar t)
=
[g_1(\bar t),\ldots,g_M(\bar t)]^\top.
\]
The Spacetime Clustering Field combines the two branches as
\begin{equation}
P_{i,k}(t)
=
\left[
\operatorname{Softmax}
\left(
\frac{
\mathbf{z}^{\mathrm{sp}}_i
+
\mathbf{B}\mathbf{g}(\bar t)}
{\lambda}
\right)
\right]_k,
\end{equation}
where $\mathbf{B}\in\mathbb{R}^{K\times M}$ is the learned event-bias matrix, $\lambda>0$ is the routing temperature, and $P_{i,k}(t)$ is the resulting routing probability for branch $k$.

Thus, temporal information enters the routing only through the annotated-event gates. The event-bias matrix $\mathbf B$ is learned, but the event times and gate parameters are fixed. Consequently, routing can adapt near an event while remaining approximately stable between events, as motivated in \cref{app:routing_derivation}.

\paragraph{Parameter sensitivity.}
Our default configuration uses a Spatial Branch MLP width of $64$ and a fixed routing temperature of $0.01$ for both training and rendering. On the TopoEgo ablation subset, we evaluated MLP widths of $32$, $64$, and $128$ under three temperature schedules: fixed sharp ($0.01$ for training and rendering), annealed ($0.1\!\to\!0.01$ during training and $0.01$ for rendering), and fixed soft ($0.1$ for training and rendering). The observed differences were within run-to-run variation, and no alternative consistently outperformed the default. We therefore use this configuration across all sequences and event families, without family-specific tuning.

\subsection{$K$-Means Initialization}
\label{app:learned_partition}

Sec.~5.1 of the main paper uses $K$-means only as a warm start for the learned Spatial Branch; the resulting clusters are not kept as the final piece assignments. Table~4 in the main paper additionally compares this design with \emph{Fixed K-Means} (Fixed-KM), which freezes the initial $K$-means partition. Here we detail the initialization procedure and explain why the partition remains learnable.

A fixed $K$-means partition based only on Euclidean proximity cannot be corrected by the rendering objective and need not coincide with the parts that follow distinct rigid motions. We therefore cluster the raw canonical Gaussian centers only to initialize branch identities and break permutation symmetry. Let $c_i^{(0)}\in\{1,\ldots,K\}$ denote the initial $K$-means label of Gaussian $i$, and let $\mathbf{z}^{\mathrm{sp}}_i\in\mathbb{R}^{K}$ denote the Spatial Branch logits defined in Sec.~\ref{app:event_state_parameterization}. We warm-start the Spatial Branch with
\begin{equation}
\label{eq:routing_init}
\mathcal{L}_{\mathrm{init}}
=
-\mathbb{E}_{i}
\left[
\log
\left[
\operatorname{Softmax}
\left(\mathbf{z}_i^{\mathrm{sp}}\right)
\right]_{c_i^{(0)}}
\right].
\end{equation}
During this warm start, we optimize only the Spatial Branch MLP for $300$ full-batch iterations using Adam with learning rate $10^{-2}$; no image-space or regularization loss is active. After this warm start, the $K$-means labels are discarded and the
routing is optimized through the reconstruction objective. This is the learned-partition configuration used by Topo-4DGS; the
\emph{Fixed-KM} variant in Table~4 of the main paper instead keeps the initial partition fixed throughout optimization.

\subsection{Optimization Details}
\label{optimization}

Sec.~5.3 of the main paper describes the optimization of the canonical object representation, motion branch trajectories, and Spacetime Clustering Field, and defines the object-level objective in terms of a photometric reconstruction term $\mathcal{L}_{\mathrm{rgb}}$. Here we specify that photometric term, the treatment of hand-occluded pixels, and the optimization choices used to stabilize the canonical geometry.

Let $\mathbf C$ and $\mathbf C^{\mathrm{gt}}$ denote the rendered and ground-truth RGB images, respectively, and let $\lambda_{\mathrm{ssim}}$ weight the D-SSIM term. All optimization stages use the weighted $L_1$ and D-SSIM photometric term
\begin{equation}
\small
\mathcal{L}_{\mathrm{rgb}}(\mathbf{C},\mathbf{C}^{\mathrm{gt}})
=
(1-\lambda_{\mathrm{ssim}})
\lVert\mathbf{C}-\mathbf{C}^{\mathrm{gt}}\rVert_1
+
\lambda_{\mathrm{ssim}}
\bigl(1-\mathrm{SSIM}(\mathbf{C},\mathbf{C}^{\mathrm{gt}})\bigr).
\end{equation}
For any rendered image-space quantity $X$, let $V$ be a binary validity mask
that equals one at supervised pixels and zero at excluded pixels. Denoting
element-wise multiplication by $\odot$, we define
\begin{equation}
\widetilde{X}^{V}
=
V\odot X
+
(1-V)\odot\operatorname{sg}[X],
\end{equation}
where $\operatorname{sg}[\cdot]$ stops gradients
(e.g., \texttt{.detach()} in PyTorch). In the object stages, $V$ is the
non-hand supervision mask, and the operator is applied to both rendered RGB
and opacity.

\paragraph{Canonical-geometry stabilization.}
\label{app:canonical_tearing}
During the coarse object-tracking and object-fine-tuning stages of Sec.~5.3, we reduce the learning rates of canonical geometry and appearance by one order of magnitude relative to pose and routing, preventing the canonical object from stretching to explain event-induced motion. During Joint Scene Refinement, Gaussian centers, recovered poses, and routing remain fixed; only appearance, opacity, scale, and rotation are refined jointly with the static background.

\paragraph{Spatial-coherence regularization.}
For objects with $K>1$, the spatial-coherence term defined in Eq.~(10) of the main paper is evaluated on at most $4096$ canonical centers per optimization step. We perturb their normalized positions using
$\boldsymbol{\epsilon}\sim\mathcal{N}(\mathbf{0},0.02^2\mathbf{I}_3)$, where the standard deviation corresponds to $2\%$ of the object bounding-sphere radius and therefore defines a local neighborhood relative to the object scale. We detach the sampled positions, so the regularizer updates the routing field without moving the canonical geometry.

\section{Event-Aware Pose Interpolation}
\label{app:pose_interpolation}

Sec.~5.3 of the main paper follows EgoGaussian to estimate rigid object poses
at a subset of frames with successful pose estimates and refers to the supplementary for interpolation
of the remaining poses. Because a missing-pose interval may cross a topological
event, standard interpolation must additionally respect the event-tree
constraint that branches share motion before the event and may move
independently afterward. This section gives the event-aware interpolation procedure used by
Topo-4DGS. We evaluate its effect in \cref{sec:supp_robustness}.

\paragraph{Interpolation setup.}
Object tracking estimates rigid poses only at a subset of frames. Following EgoGaussian~\cite{zhang2024egogaussian}, we complete each trajectory by interpolating between consecutive frames for which pose estimation succeeds. Let $t_a<t_b$ be two such frames, with $N=t_b-t_a-1$ missing poses between them and $n=N+1$ relative-transform steps. Writing their relative endpoint transform as
\begin{equation}
\mathbf{T}_{b}\mathbf{T}_{a}^{-1}
=
(\mathbf{R},\mathbf{d})
\in\mathrm{SE}(3),
\end{equation}
we seek a constant transform step $\boldsymbol{\Delta}=(\mathbf{R}_{\Delta},\mathbf{d}_{\Delta})$ whose repeated application exactly reaches the endpoint.

\paragraph{Endpoint-exact transform interpolation.}
\label{endpoint_exact_slerp}
We compute the rotational step as the geodesic $n$-th root of $\mathbf{R}$ on
$\mathrm{SO}(3)$:
\begin{equation}
\mathbf{R}_{\Delta}
=
\exp\!\left(
\frac{\log(\mathbf{R})}{n}
\right).
\end{equation}
This is equivalent to constant-step Slerp and satisfies
$\mathbf{R}_{\Delta}^{n}=\mathbf{R}$ while keeping every interpolated rotation
in $\mathrm{SO}(3)$. Under the left-composition convention
$\mathbf{T}_{r+1}=\boldsymbol{\Delta}\cdot\mathbf{T}_r$, the translation
accumulated over $n$ steps is
$\sum_{r=0}^{n-1}\mathbf{R}_{\Delta}^{r}\mathbf{d}_{\Delta}$.
Defining
\begin{equation}
\mathbf{S}_{\Delta}
=
\sum_{r=0}^{n-1}\mathbf{R}_{\Delta}^{r},
\end{equation}
the required translation step is
\begin{equation}
\label{eq:interp_translation_root}
\mathbf{d}_{\Delta}
=
\mathbf{S}_{\Delta}^{-1}\mathbf{d}.
\end{equation}
Repeating $\boldsymbol{\Delta}$ exactly recovers the endpoint pose. If
$\mathbf{S}_{\Delta}$ is singular or sufficiently \textit{close} to singular that its
inverse cannot be computed reliably, we use the linear fallback
$\mathbf{d}_{\Delta}=\mathbf{d}/n$.

\paragraph{Gaps that straddle events.}
When a gap straddles an event, the branches must share their pre-event motion and interpolate independently afterward. We divide such gaps into contiguous segments. Let
\begin{equation}
\mathbf T_k^{ab}
=
\mathbf T_k(t_b)\mathbf T_k(t_a)^{-1}
\end{equation}
be the relative endpoint transform required by branch $k$ across the complete gap. Within each pre-boundary segment, branches use the interpolation step of their shared representative. Let $\mathbf T_{\mathrm{acc},k}$ be the transform accumulated for branch $k$ over these segments. Before the final segment, its remaining endpoint transform is
\begin{equation}
\mathbf T_{\mathrm{res},k}
=
\mathbf T_k^{ab}\mathbf T_{\mathrm{acc},k}^{-1}
=
(\mathbf R_{\mathrm{res},k},\mathbf d_{\mathrm{res},k})
\in\mathrm{SE}(3).
\end{equation}
If the final segment contains $n_S$ steps, we compute
\begin{align}
\mathbf R_{\mathrm{last},k}
&=
\exp\!\left(
\frac{\log(\mathbf R_{\mathrm{res},k})}{n_S}
\right),\\
\mathbf d_{\mathrm{last},k}
&=
\left(
\sum_{r=0}^{n_S-1}
\mathbf R_{\mathrm{last},k}^{r}
\right)^{-1}
\mathbf d_{\mathrm{res},k}.
\end{align}
If the matrix
$\sum_{r=0}^{n_S-1}\mathbf{R}_{\mathrm{last},k}^{r}$ is singular or cannot be
inverted reliably, we use
$\mathbf d_{\mathrm{last},k}=\mathbf d_{\mathrm{res},k}/n_S$. This preserves shared motion before each event while reaching every branch endpoint exactly. With $K=1$ or no event inside the gap, the construction reduces to the single geodesic root above.

\section{Topological Boundary Resolution}
\label{app:per_piece}

Sec.~5.2 of the main paper introduces \emph{Topological Boundary Resolution} to update surface support at topological events. Here we provide the implementation details of its boundary construction, temporal activation, branch binding, and support retirement.

For terminal pieces $p$ and $q$, let $t_{\mathrm{sep}}(p,q)$ be the first event time after which they belong to different event-tree components, with $t_{\mathrm{sep}}(p,q)=\infty$ if they never separate. We process every finite pair separately in each piece's canonical frame using only the recovered Gaussians, learned Spacetime Clustering Field, and event tree. Pairs without reliable contact evidence are skipped; no ground-truth boundary geometry or per-Gaussian piece labels are used. For \textit{fusion} and \textit{close}, we temporarily reverse the input sequence before this stage so that the same construction applies to every event family.

\subsection{Event-Conditioned Temporal Activation and Time Refinement}
\label{app:boundary_timers}

For completeness, we restate the general temporal activation parameterization
from Sec.~5.2 of the main paper before specifying the constrained form used in
our implementation. Each Gaussian $i$ has an activation boundary $a_{s,i}$, a
deactivation boundary $a_{f,i}$, and transition widths
$b_{s,i},b_{f,i}>0$. Its time-dependent opacity is
\begin{equation}
\alpha_i^t=\alpha_i\mathcal{W}_i(t),
\end{equation}
where
\begin{equation}
\mathcal{W}_i(t)=
\begin{cases}
\exp\!\left[
-\left((t-a_{s,i})/b_{s,i}\right)^2
\right],
& t<a_{s,i}, \\[6pt]
1,
& a_{s,i}\leq t\leq a_{f,i}, \\[6pt]
\exp\!\left[
-\left((t-a_{f,i})/b_{f,i}\right)^2
\right],
& t>a_{f,i}.
\end{cases}
\label{eq:temporal_activation_supp}
\end{equation}
Here, $a_{s,i}$ marks when Gaussian $i$ becomes fully active,
$a_{f,i}$ marks when it begins to deactivate, and $b_{s,i}$ and $b_{f,i}$
control the corresponding transition widths. All temporal quantities are
expressed in frame units.

In our implementation, we use the tied-width specialization
\begin{equation}
b_{s,i}=b_s,
\qquad
b_{f,i}=b_f,
\qquad \forall i,
\end{equation}
so all Gaussians share the same activation and deactivation transition widths.
This is a constrained instance of the general parameterization in
Eq.~\eqref{eq:temporal_activation_supp}; the activation and deactivation
boundaries remain specific to each Gaussian.

Newly created or exposed support is initialized to activate at its associated
event, whereas obsolete or concealed support is initialized to deactivate at
the event. Gaussians unaffected by an event remain active throughout the
sequence.

\paragraph{Time Refinement.}
Time Refinement photometrically optimizes the event-initialized activation and
deactivation boundaries during Joint Scene Refinement. Only boundaries
initialized at an event are optimized; the annotated event times remain fixed,
and Gaussians unaffected by any event retain their original activation
windows. The shared widths $b_s$ and $b_f$ follow a fixed narrowing schedule
and are not optimized independently for each Gaussian. We enforce
$a_{f,i}\geq a_{s,i}$ after every update. Densification and pruning are
disabled during this stage to preserve Gaussian identities and optimizer
state. This procedure permits small corrections to the activation timing
without modifying the supplied topology.

\subsection{Per-Piece Construction and Branch Binding}
\label{app:boundary_construction}

\paragraph{Piece-specific canonical frames.}
Canonical geometry and branch motion are recovered jointly. Consequently, each
piece's canonical frame is determined only up to a rigid transformation:
rigidly transforming its canonical geometry and applying the inverse change to
its trajectory leaves all rendered observations unchanged. Two pieces can
therefore be separated in canonical space even when they touch in the observed
frames. A single boundary face fitted between their canonical point sets may
then appear detached from both pieces after rendering. We avoid this failure by
constructing a separate boundary face in each piece's canonical frame.

\paragraph{Boundary-face construction.}
At the last frame with an available interpolated pose, we evaluate the Spacetime Clustering Field and assign each existing Gaussian to its maximum-probability terminal piece, producing one canonical point set per piece. We implement the mutual contact search summarized in Sec.~5.2 of the main paper by evaluating nearest-neighbor proximity in both directions. On each piece, the points closest to the opposing piece define its facing contact region. This procedure does not require per-Gaussian piece supervision.

For each contact region, we fit a local plane, project the selected points, and fill their two-dimensional convex hull with a regular grid. Grid spacing is derived from the local contact extent, and the grid is offset toward the interior of its piece. Each spawned Gaussian inherits its initial color, scale, and rotation from the nearest existing Gaussian in that piece. We skip pairs for which the contact evidence is insufficient or a reliable plane cannot be estimated.

\paragraph{Branch binding.}
Each spawned Gaussian is bound to the motion branch of the terminal piece for which its boundary face was constructed, as stated in Sec.~5.2 of the main paper. This binding is retained during Joint Scene Refinement and rendering. It applies only to spawned support; existing Gaussians retain their learned soft routing distributions.

\subsection{Retiring Obsolete and Concealed Support}
\label{app:boundary_retirement}

Existing Gaussians are retired when the event geometry indicates that their
support has become obsolete or concealed. For each piece, we orient its contact
plane toward the opposing contact region and compute the signed distance of
every Gaussian center from that plane. For each terminal piece $p$, let $\delta_p$ be the median over the three
principal scale components of the Gaussians assigned to that piece. We select
Gaussians whose centers lie on the contact-facing side of the plane and within
distance $\delta_p$ of it. These
Gaussians receive a deactivation boundary at the associated event time. If
multiple event regions select the same Gaussian, we retain the earliest
deactivation time; a selection is rejected if it would retire an entire terminal piece.

Before Joint Scene Refinement, we additionally initialize deactivation
boundaries using an explicit depth-based visibility test. For each event that
may conceal or retire support, we render the composed scene---the background
and all posed objects---from up to twelve training views with available camera poses before and after
the event. A Gaussian is considered visible when its posed center projects
inside the camera frustum and lies no more than $5\%$ behind the rendered scene
depth at that pixel. We assign $a_{f,i}$ to the event time when Gaussian $i$ is
visible in at least one pre-event view but in no sampled post-event view; a
single post-event observation is therefore sufficient to keep it active.
Hands cannot trigger deactivation because they are not represented in the
rendered depth buffer. If the proposed deactivations would remove at least
$90\%$ of a terminal piece, they are canceled for that branch to avoid
retiring an entire piece because of a depth-estimation error. This depth-based
test and the contact-plane criterion provide alternative initialization paths;
if multiple events or procedures select the same Gaussian, we retain the
earliest deactivation time. Only $a_{f,i}$ is initialized, and it may
subsequently be adjusted through photometric Time Refinement.

\section{Additional Ablations}
\label{sec:supp_robustness}

The experiments in the main paper use manually annotated event specifications, the ground-truth number of terminal pieces, and the native input frame rate. Here, we further analyze Event-Aware Pose Interpolation and sensitivity to the terminal-piece count, event time, and input frame rate. These experiments are controlled diagnostics complementing the component-level ablations in Sec.~6 of the main paper.

\paragraph{Scope and evaluation subsets.}
The component ablations in the main paper use eight clips: one
EPIC-KITCHENS clip and one Ego4D clip for each of the four event families.
The supplementary experiments in
\cref{tab:interpolation_ablation,tab:k_sensitivity,tab:abl_t,tab:abl_fps}
instead study sensitivity to pose interpolation, terminal-piece count, event
time, and temporal input density. Because each setting requires retraining the
complete model, we evaluate these analyses on subsets of the same eight clips.

Event-Aware Pose Interpolation (\cref{tab:interpolation_ablation}) and
terminal-piece-count sensitivity (\cref{tab:k_sensitivity}) use the same four
clips: one fission, one fusion, one open, and one close sequence. This selection
covers all event families, including both events that separate or expose object
parts and events that join or conceal them.

Event-time sensitivity (\cref{tab:abl_t}) and input-frame-rate sensitivity
(\cref{tab:abl_fps}) use the fission and open clips from this four-clip set.
These experiments perturb temporal information, whose effect is most directly
observable in separation events: after the event, the reconstruction must
recover independently moving pieces and newly exposed surface support. Using
the same two clips for both analyses also limits the cost of retraining every
temporal perturbation.

Within each table, all rows use the same clips and evaluation protocol.
Experiments reported in different tables were trained independently, so their
default Topo-4DGS results may differ slightly;
comparisons should therefore be made within each table. The full 24-clip
benchmark and the eight-clip component ablations in the main paper provide the
primary evaluation, while the supplementary experiments characterize
sensitivity to specific input and interpolation choices.

\paragraph{Event-Aware Pose Interpolation.}
The reconstruction-baseline rows measure complete pipeline performance under the same sparse-frame supervision. They are included as end-to-end comparisons, not as interpolation variants; only the four Topo-4DGS rows isolate the effect of the pose-completion strategy.
We evaluate the interpolation formulation from \cref{app:pose_interpolation} in \cref{tab:interpolation_ablation}. We consider held-out gap widths $g\in\{1,5,10\}$, alternating $g$ training and $g$ held-out frames while maintaining a training fraction of $1/2$. Within each gap width, all four Topo-4DGS interpolation variants share the
same reconstruction up to the interpolation stage and differ only in how missing poses are completed. We compare holding the last valid pose, EgoGaussian's unconstrained rotation update~\cite{zhang2024egogaussian}, endpoint-exact Slerp without event boundaries (\cref{endpoint_exact_slerp}), and the complete event-aware formulation (\cref{app:pose_interpolation}). PSNR and LPIPS are computed inside the ground-truth mask of the topology-changing object and only on evaluation frames without keyframe pose estimates; hand pixels are excluded following the main evaluation protocol.

Both geodesic variants outperform the two non-geodesic controls across all gap widths. At the standard $g{=}1$ split, Event-Aware Pose Interpolation reaches $22.67$\,dB PSNR and $0.207$ LPIPS, compared with $22.27$\,dB and $0.212$ for ordinary Slerp and $21.15$\,dB and $0.225$ for the unconstrained update. Accounting for event boundaries improves PSNR over ordinary Slerp by $0.40$\,dB at $g{=}1$ and $0.32$\,dB at both $g{=}5$ and $g{=}10$, while also reducing LPIPS. The complete formulation also outperforms all reconstruction baselines across all three gap widths. These results validate both endpoint-exact geodesic interpolation and the explicit division of pose gaps at topological events.
\begin{table}[t]
\caption{\textbf{Event-Aware Pose Interpolation.}
Reconstruction baselines and interpolation variants on the four-clip subset containing one sequence from each event family. Each split alternates $g$ training and $g$ evaluation frames, preserving the $1{:}1$ ratio; larger $g$ creates longer intervals without directly estimated poses. Metrics are computed only on those evaluation frames. PSNR uses topology-changing-object mask pixels, LPIPS uses the corresponding bounding-box crop, and hand pixels are zeroed only during scoring.}
\label{tab:interpolation_ablation}
\centering
\footnotesize
\setlength{\tabcolsep}{3pt}
\renewcommand{\arraystretch}{1.1}
\begin{tabular}{l ccc ccc}
\toprule
& \multicolumn{3}{c}{PSNR $\uparrow$} & \multicolumn{3}{c}{LPIPS $\downarrow$} \\
\cmidrule(lr){2-4}\cmidrule(lr){5-7}
Method & $g{=}1$ & $g{=}5$ & $g{=}10$ & $g{=}1$ & $g{=}5$ & $g{=}10$ \\
\midrule
\multicolumn{7}{@{}l}{\emph{Baselines}}\\
Def-3DGS~\cite{yang2024deformable3dgs}                 & 20.27 & 16.43 & 12.46 & 0.252 & 0.267 & 0.389 \\
DeGauss~\cite{wang2025degauss}                  & 19.62 & 16.12 & 11.34 & 0.265 & 0.276 & 0.432 \\
EgoGaussian~\cite{zhang2024egogaussian}              & 20.12 & 16.53 & 15.71 & 0.238 & 0.263 & 0.392 \\
4DGS~\cite{wu2024fourdgs}              & 17.25 & 13.20 & 9.45  & 0.385 & 0.491 & 0.546 \\
\midrule
\multicolumn{7}{@{}l}{\emph{Ours -- interpolation variants}}\\
Hold last valid pose   & 19.58 & 16.91 & 15.91 & 0.268 & 0.257 & 0.312 \\
Unconstrained update     & 21.15 & 17.33 & 15.69 & 0.225 & 0.243 & 0.317 \\
Closed-form Slerp        & 22.27 & 18.30 & 16.54 & 0.212 & 0.230 & 0.300 \\
Event-aware Slerp (full) & \textbf{22.67} & \textbf{18.62} & \textbf{16.86} & \textbf{0.207} & \textbf{0.225} & \textbf{0.294} \\
\bottomrule
\end{tabular}
\end{table}

\paragraph{Sensitivity to terminal-piece count.}
For all benchmark results, we set $K$ to the manually annotated ground-truth terminal-piece count, i.e., the actual number of terminal pieces of the object. Here we test how reconstruction changes when this input is under- or over-specified. \Cref{tab:k_sensitivity} evaluates $K\in\{1,2,4,8,16,100\}$ on four clips, one from each event family, all with ground-truth $K{=}2$. The table reports topology-changing-object reconstruction and optimization throughput; $K{=}1$ is under-specified, $K\in\{4,8,16\}$ represents moderate over-specification, and $K{=}100$ is a deliberately extreme stress test.

Relative to the ground-truth $K{=}2$ setting, $K{=}1$ reduces PSNR by $2.44$\,dB because the model cannot represent both terminal-piece motions. Moderate over-specification is well tolerated, with $K\in\{4,8,16\}$ remaining within $0.37$\,dB of the ground-truth $K{=}2$ setting. The small non-monotonic differences within this range do not indicate that a larger $K$ is preferable; rather, they show that unused capacity does not systematically degrade reconstruction. Extreme over-specification at $K{=}100$ reduces PSNR by $1.57$\,dB and lowers the optimization rate from $37.24$ to $23.42$ iterations per second. The results therefore show that under-specification is harmful, moderate count errors are tolerated, and only extreme over-specification substantially affects quality and cost.
\begin{table}[t]
\caption{\textbf{Sensitivity to terminal-piece count $K$.}
The benchmark uses the ground-truth count; here we vary the input $K$ on the
4-clip, all-family diagnostic subset, whose clips all have ground-truth
$K{=}2$. $K{=}1$ is under-specified, $K\in\{4,8,16\}$ is moderately
over-specified, and $K{=}100$ is an extreme stress test. Rate denotes
optimization throughput.}
\label{tab:k_sensitivity}
\centering
\small
\setlength{\tabcolsep}{4pt}
\renewcommand{\arraystretch}{1.08}

\begin{tabular}{l ccc c}
\toprule
& \multicolumn{3}{c}{\textbf{Topo. Object (region)}} & \\
\cmidrule(lr){2-4}
$K$
& PSNR $\uparrow$
& SSIM $\uparrow$
& LPIPS $\downarrow$
& Rate (it/s) $\uparrow$ \\
\midrule
1   & 20.62 & 0.741 & 0.198 & 43.83 \\
2 (default)   & 23.06 & 0.781 & 0.173 & 37.24 \\
4   & 22.86 & 0.773 & 0.176 & 34.13 \\
8   & 23.43 & 0.786 & 0.176 & 33.62 \\
16  & 23.31 & 0.778 & 0.173 & 33.48 \\
100 & 21.49 & 0.758 & 0.196 & 23.42 \\
\bottomrule
\end{tabular}
\end{table}

\paragraph{Event-time sensitivity.}
All benchmark results use the manually annotated event times. Here, we measure robustness when these inputs are replaced by automatically predicted times or offset from the annotations. \Cref{tab:abl_t} reports full-scene and topology-changing-object reconstruction on the 2-clip \textit{fission}--\textit{open} diagnostic subset. The ground-truth row represents the benchmark setting.

First, we replace the annotations with the event times predicted by the automatic topology-discovery module. These predictions reduce topology-changing-object PSNR by only $0.07$\,dB relative to the ground-truth configuration, indicating that the predicted times remain sufficiently accurate for reconstruction on these clips.

Second, we apply controlled offsets of $\pm8$ and $\pm16$ frames to the annotated times. Negative offsets place the event earlier, and positive offsets place it later; the two magnitudes correspond to half and the full median timing
error of the automatic topology-discovery module. The $\pm8$-frame perturbations remain within $0.14$\,dB of the ground-truth configuration. Moving the event 16 frames earlier reduces topology-changing-object PSNR by $0.47$\,dB, whereas delaying it by 16 frames produces the largest reduction, $1.26$\,dB. A delayed
event forces early post-event frames to retain the pre-event shared-motion and
surface-support configuration; an early event only makes the post-event degrees
of freedom available before they are needed, allowing optimization to keep the
branches aligned.

Full-scene SSIM increases slightly under the offsets, but these changes are
not accompanied by consistent improvements in PSNR, LPIPS, or
topology-changing-object reconstruction. Full-scene SSIM is strongly influenced
by the mostly unchanged background and can favor smoother local structure. We
therefore interpret the metrics jointly. Overall, the method accommodates the
automatically predicted times and moderate controlled offsets, but becomes less
reliable when the event is substantially delayed.

\begin{table}[t]
\caption{\textbf{Event-time sensitivity.}
The ground-truth row uses the manually annotated event times; the remaining rows use automatically predicted times or controlled offsets from the annotations. Negative offsets place the event earlier and positive offsets place it later. Results are averaged over the two-clip \textit{fission}--\textit{open} subset. Object-region metrics use mask PSNR, padded-crop SSIM, and bounding-box LPIPS; hand pixels are zeroed only during scoring.}
\centering
\scriptsize
\setlength{\tabcolsep}{2.2pt}
\renewcommand{\arraystretch}{1.1}
\begin{tabular}{l ccc ccc}
\toprule
& \multicolumn{3}{c}{Full scene} & \multicolumn{3}{c}{Topo. Object (region)} \\
\cmidrule(lr){2-4}\cmidrule(lr){5-7}
Event times
& PSNR$\uparrow$ & SSIM$\uparrow$ & LPIPS$\downarrow$
& PSNR$\uparrow$ & SSIM$\uparrow$ & LPIPS$\downarrow$ \\
\midrule
Ground-truth   & 26.92 & 0.842 & 0.249 & 26.66 & 0.815 & 0.125 \\

Automatically predicted      & 26.87 & 0.838 & 0.253 & 26.59 & 0.805 & 0.139 \\

\hline

Offset $-16$ frames     & 26.15 & 0.859 & 0.272 & 26.19 & 0.789 & 0.178 \\

Offset $-8$ frames      & 26.17 & 0.859 & 0.261 & 26.52 & 0.796 & 0.141 \\

Offset $+8$ frames      & 26.19 & 0.860 & 0.263 & 26.58 & 0.798 & 0.149 \\

Offset $+16$ frames     & 26.19 & 0.859 & 0.281 & 25.40 & 0.813 & 0.199 \\
\bottomrule
\end{tabular}
\label{tab:abl_t}
\end{table}

\paragraph{Input frame rate.}
The benchmark uses each clip at its native frame rate, with the alternating
$1{:}1$ training/evaluation split defined in the main paper. Here, we reduce
the temporal density of the training input while leaving the evaluation frames
unchanged. Starting from the ordered default training-frame set
$\mathcal{T}_{\mathrm{train}}=(t_0,t_1,\ldots)$, the full, half, and quarter
settings retain
$\{t_m\}$, $\{t_{2m}\}$, and $\{t_{4m}\}$, respectively, always beginning
with the first default training frame. No frames are added or repeated at the
end of a clip. Thus, ``half'' and ``quarter'' refer to fractions of the
default training frames, not fractions of all native video frames.

\Cref{tab:abl_fps} evaluates these settings on the same 2-clip
\textit{fission}--\textit{open} subset used for event-time sensitivity. Relative to the default
training set, retaining one-half and one-quarter of its frames reduces
topology-changing-object PSNR by $3.92$ and $4.35$\,dB, respectively, while
full-scene PSNR decreases by $0.99$ and $2.08$\,dB. All three
topology-changing-object metrics degrade under both reductions, even though
some full-scene SSIM and LPIPS values vary non-monotonically. The
topology-changing object is therefore more sensitive to reduced temporal
training coverage than the complete scene.
\begin{table}[t]
\caption{\textbf{Input-frame-rate sensitivity.}
Starting from the benchmark's alternating $1{:}1$ split, the reduced settings
retain every second or fourth frame of the default training set while leaving
all evaluation frames unchanged. The fractions therefore refer to the number
of default training frames retained. Results are averaged over the 2-clip
fission--open diagnostic subset. PSNR/SSIM~$\uparrow$, LPIPS~$\downarrow$;
hand pixels are zeroed in prediction and ground truth only during scoring.}
\label{tab:abl_fps}
\centering
\footnotesize
\setlength{\tabcolsep}{2pt}
\renewcommand{\arraystretch}{1.1}
\begin{tabular}{l ccc ccc}
\toprule
& \multicolumn{3}{c}{Full scene} & \multicolumn{3}{c}{Topo. Object (region)} \\
\cmidrule(lr){2-4}\cmidrule(lr){5-7}
Training input
& PSNR$\uparrow$ & SSIM$\uparrow$ & LPIPS$\downarrow$
& PSNR$\uparrow$ & SSIM$\uparrow$ & LPIPS$\downarrow$ \\
\midrule
Default training set      & 27.42 & 0.817 & 0.205 & 26.78 & 0.812 & 0.132 \\
$1/2$ training frames              & 26.43 & 0.842 & 0.269 & 22.86 & 0.760 & 0.202 \\
$1/4$ training frames            & 25.34 & 0.758 & 0.234 & 22.43 & 0.657 & 0.183 \\
\bottomrule
\end{tabular}
\end{table}

\section{Runtime and Resource Analysis}
\label{app:runtime}

The main paper focuses on reconstruction quality. Here we complement those results with the computational cost of the complete Topo-4DGS pipeline, including training time, the additional cost of Topological Boundary Resolution and Joint Scene Refinement, checkpoint size, and rendering speed.

\paragraph{Training runtime.}
We report total training time, excluding data preprocessing. Across all 24 TopoEgo clips, Topo-4DGS requires a median of 10.9\,h per clip, with a range of 3.6--22.8\,h on a single A100 GPU. On 8 clips with directly comparable EgoGaussian runs under matched hardware and iteration budgets, Topo-4DGS requires a median of 1.28$\times$ the EgoGaussian training time, with a range of 0.80--1.68$\times$. Runtime also increases with the number of reconstructed objects: the median is 9.3\,h for one-object scenes and 12.8\,h for two-object scenes.

\paragraph{Cost of Topological Boundary Resolution and Joint Scene Refinement.}
On the 8 ablation clips, removing Topological Boundary Resolution and the subsequent Joint Scene Refinement reduces training time by a median of 5.8\,h. Together, these stages account for a median of 34\% of total training time, with a range of 9--59\% across clips.

\paragraph{Model size and rendering.}
Across the 24 benchmark clips, the final checkpoint has a median storage footprint of 55.5\,MB and a range of 27--399\,MB. At the benchmark evaluation resolution, composed multi-object rendering runs at approximately $5.2$--$7.4$\,FPS on a single A100 GPU.

\section{Additional Quantitative Comparisons}
\label{app:additional_quantitative}

Sec.~6 of the main paper reports the aggregate benchmark comparisons. Here, we provide additional object-only and per-sequence results using the same evaluation protocol.

\paragraph{Object-only comparison with EgoGaussian.}
The experiments in the main paper (Tab. 1 and 2) evaluate object regions within the full composite render because every baseline supports this protocol. Topo-4DGS and EgoGaussian additionally maintain explicit object--background decompositions, which permit isolated object rendering. In \cref{tab:final_egoobjonly}, we render the object on black, zero the ground truth outside its mask, and exclude hand pixels from both images during scoring. We also report $\alpha$-IoU between the rendered opacity thresholded at $0.5$ and the ground-truth object mask.
\begin{table}[t!]
\caption{\textbf{Object-only topology-changing-object reconstruction on TopoEgo.}
Topo-4DGS and EgoGaussian support isolated object rendering, unlike baselines
without an explicit object--background decomposition. Values are averaged over
the 3 clips in each source--event cell and over all 24 TopoEgo clips in the
overall row. PSNR/SSIM/$\alpha$-IoU $\uparrow$, LPIPS $\downarrow$;
$\alpha$-IoU thresholds both rendered opacity and the ground-truth mask at
$0.5$. Hand pixels are zeroed in both prediction and ground truth only during scoring.}
\label{tab:final_egoobjonly}
\centering
\footnotesize
\setlength{\tabcolsep}{3.2pt}
\renewcommand{\arraystretch}{1.05}
\begin{tabular}{lllcccc}
\toprule
Source & Event & Method &
PSNR$\uparrow$ & SSIM$\uparrow$ & LPIPS$\downarrow$ & $\alpha$-IoU$\uparrow$ \\
\midrule

\multirow{9}{*}{EPIC-K.}
& \multirow{2}{*}{\textit{fission}}
& EgoGaussian & 19.46 & 0.597 & 0.364 & 0.465 \\
& & Topo-4DGS   & \textbf{21.92} & \textbf{0.650} & \textbf{0.355} & \textbf{0.488} \\

\cmidrule(lr){2-7}
& \multirow{2}{*}{\textit{fusion}}
& EgoGaussian & 13.75 & 0.218 & 0.441 & 0.066 \\
& & Topo-4DGS   & \textbf{25.21} & \textbf{0.734} & \textbf{0.309} & \textbf{0.507} \\

\cmidrule(lr){2-7}
& \multirow{2}{*}{\textit{open}}
& EgoGaussian & 18.60 & 0.608 & 0.309 & 0.514 \\
& & Topo-4DGS   & \textbf{21.68} & \textbf{0.643} & \textbf{0.293} & \textbf{0.574} \\

\cmidrule(lr){2-7}
& \multirow{2}{*}{\textit{close}}
& EgoGaussian & 21.57 & 0.647 & 0.375 & 0.353 \\
& & Topo-4DGS   & \textbf{24.06} & \textbf{0.709} & \textbf{0.342} & \textbf{0.479} \\

\midrule

\multirow{9}{*}{Ego4D}
& \multirow{2}{*}{\textit{fission}}
& EgoGaussian & 20.61 & 0.688 & 0.337 & 0.518 \\
& & Topo-4DGS   & \textbf{22.03} & \textbf{0.736} & \textbf{0.280} & \textbf{0.591} \\

\cmidrule(lr){2-7}
& \multirow{2}{*}{\textit{fusion}}
& EgoGaussian & 18.75 & 0.606 & 0.265 & 0.498 \\
& & Topo-4DGS   & \textbf{20.95} & \textbf{0.637} & \textbf{0.228} & \textbf{0.606} \\

\cmidrule(lr){2-7}
& \multirow{2}{*}{\textit{open}}
& EgoGaussian & 18.97 & \textbf{0.623} & \textbf{0.355} & 0.528 \\
& & Topo-4DGS   & \textbf{19.13} & 0.593 & 0.366 & \textbf{0.561} \\

\cmidrule(lr){2-7}
& \multirow{2}{*}{\textit{close}}
& EgoGaussian & 16.69 & 0.562 & 0.356 & 0.603 \\
& & Topo-4DGS   & \textbf{18.90} & \textbf{0.599} & \textbf{0.340} & \textbf{0.619} \\

\midrule

\multicolumn{2}{l}{\multirow{2}{*}{Overall}}
& EgoGaussian & 18.55 & 0.569 & 0.350 & 0.443 \\
& & Topo-4DGS   & \textbf{21.74} & \textbf{0.663} & \textbf{0.314} & \textbf{0.553} \\

\bottomrule
\end{tabular}
\end{table}

Across the 24 TopoEgo clips, Topo-4DGS improves over EgoGaussian by $+3.19$\,dB PSNR, $+0.094$ SSIM, $-0.036$ LPIPS, and $+0.110$ $\alpha$-IoU. The only metric-level exceptions among the source--event groups occur for Ego4D \textit{open}, where EgoGaussian obtains higher SSIM ($0.623$ versus $0.593$) and lower LPIPS ($0.355$ versus $0.366$); the two methods are nearly tied in PSNR ($18.97$ versus $19.13$\,dB), while Topo-4DGS remains better in $\alpha$-IoU. EgoGaussian obtains a mean $\alpha$-IoU of only $0.066$ on EPIC-KITCHENS \textit{fusion}, reflecting severe tracking failures around the \textit{fusion} events. This comparison isolates the reconstructed object rather than its appearance within the full-scene composite.

\paragraph{Per-sequence comparison with HyperNeRF.}
\Cref{tab:final_hypernerf} provides the per-sequence breakdown of the same
four-sequence HyperNeRF comparison summarized in Sec.~6.1 of the main paper.
HyperNeRF~\cite{park2021hypernerf} is substantially more expensive to train
than the Gaussian baselines, making a complete 24-clip evaluation impractical.
One sequence belongs to TopoEgo; the other three, marked with a dagger
($\dagger$), satisfy the same clip-selection criteria (Sec.~3 of the main
paper) but were selected for their tractable duration.

Unconstrained egocentric footage is particularly challenging for neural-field reconstruction. Large camera motion and parallax can dominate the comparatively sporadic motion of manipulated objects, while the actor and hands repeatedly occlude the scene~\cite{tschernezki2021neuraldiff}. Rapid motion, limited overlapping observations, and scene inconsistencies can additionally produce blurred geometry and reconstruction artifacts~\cite{zhang2024egogaussian}. HyperNeRF was not designed around these egocentric-specific decomposition and
occlusion challenges. Its low absolute performance should therefore be interpreted as evidence that a general topology-varying neural field does not directly transfer to this demanding setting, rather than as a general comparison between neural fields and Gaussian representations.

Across this subset, Topo-4DGS improves mean full-scene PSNR from $21.11$ to $31.86$\,dB and topology-changing-object PSNR from $15.73$ to $25.79$\,dB, while reducing the corresponding LPIPS values by $0.205$ and $0.218$.
Topo-4DGS also improves every reported full-scene and object metric on every sequence. Although this targeted comparison is not a substitute for a full benchmark evaluation, it shows that the advantage extends to a topology-varying neural-field baseline under matched egocentric evaluation.

\begin{table*}[t]
\centering
\caption{\textbf{Per-sequence comparison with HyperNeRF.}
Full-scene, topology-changing-object, and other-dynamic-object reconstruction
on four short egocentric sequences using matched held-out frames and scoring.
Hand pixels are zeroed in prediction and ground truth only during scoring. A
dagger ($\dagger$) marks the three sequences outside TopoEgo selected to keep
HyperNeRF~\cite{park2021hypernerf} training tractable. Other Dyn.\ Objects
entries are omitted when a sequence contains no evaluated additional dynamic
object.}
\label{tab:final_hypernerf}

\footnotesize
\begin{tabular}{@{}ll ccc ccc ccc@{}}
\toprule
& &
\multicolumn{3}{c}{Full scene} &
\multicolumn{3}{c}{Topo. Object (region)} &
\multicolumn{3}{c}{Other Dyn. Objects (region)} \\
\cmidrule(lr){3-5}
\cmidrule(lr){6-8}
\cmidrule(lr){9-11}

Clip & Method
& PSNR$\uparrow$ & SSIM$\uparrow$ & LPIPS$\downarrow$
& PSNR$\uparrow$ & SSIM$\uparrow$ & LPIPS$\downarrow$
& PSNR$\uparrow$ & SSIM$\uparrow$ & LPIPS$\downarrow$ \\
\midrule

\multirow{2}{*}{P01\_105$^\dagger$}
& HyperNeRF
& 19.46 & 0.707 & 0.337
& 12.78 & 0.414 & 0.357
& 11.09 & 0.224 & 0.364 \\

& Topo-4DGS
& \textbf{30.37} & \textbf{0.946} & \textbf{0.102}
& \textbf{24.17} & \textbf{0.888} & \textbf{0.108}
& \textbf{21.07} & \textbf{0.815} & \textbf{0.120} \\
\midrule

\multirow{2}{*}{P28\_105$^\dagger$}
& HyperNeRF
& 19.28 & 0.550 & 0.340
& 16.92 & 0.513 & 0.450
& -- & -- & -- \\

& Topo-4DGS
& \textbf{35.37} & \textbf{0.966} & \textbf{0.071}
& \textbf{31.50} & \textbf{0.907} & \textbf{0.137}
& -- & -- & -- \\
\midrule

\multirow{2}{*}{P03\_120$^\dagger$}
& HyperNeRF
& 23.11 & 0.771 & 0.244
& 16.51 & 0.508 & 0.392
& 14.35 & 0.176 & 0.471 \\

& Topo-4DGS
& \textbf{29.56} & \textbf{0.946} & \textbf{0.087}
& \textbf{21.71} & \textbf{0.770} & \textbf{0.224}
& \textbf{17.66} & \textbf{0.504} & \textbf{0.291} \\
\midrule

\multirow{2}{*}{P04\_04}
& HyperNeRF
& 22.60 & 0.728 & 0.247
& 16.72 & 0.466 & 0.360
& 16.66 & 0.248 & 0.346 \\

& Topo-4DGS
& \textbf{32.15} & \textbf{0.914} & \textbf{0.086}
& \textbf{25.78} & \textbf{0.862} & \textbf{0.219}
& \textbf{27.82} & \textbf{0.834} & \textbf{0.193} \\
\midrule

\multirow{2}{*}{Mean}
& HyperNeRF
& 21.11 & 0.689 & 0.292
& 15.73 & 0.475 & 0.390
& 14.03 & 0.216 & 0.394 \\

& Topo-4DGS
& \textbf{31.86} & \textbf{0.943} & \textbf{0.087}
& \textbf{25.79} & \textbf{0.857} & \textbf{0.172}
& \textbf{22.18} & \textbf{0.718} & \textbf{0.201} \\

\bottomrule
\end{tabular}
\end{table*}

\paragraph{Per-sequence results.}
Tables~\ref{tab:full_scene}, \ref{tab:primary}, \ref{tab:secondary}, and~\ref{tab:primary_postevent} provide the per-sequence results underlying the aggregate TopoEgo comparisons in the main paper. They report full-scene
reconstruction, topology-changing-object reconstruction, Other Dyn.\ Objects, and post-event topology-changing-object reconstruction, respectively. The Other Dyn.\ Objects results include only clips containing at least one
annotated dynamic object besides the topology-changing object and average all such objects within each eligible clip.

For the post-event evaluation in Table~\ref{tab:primary_postevent}, frames before the first annotated event are excluded. In clips containing multiple events, each subsequent frame is weighted by the number of annotated events
that have occurred by that frame, so frames reflecting multiple completed transitions receive proportionally greater weight. The same weighting is applied to every method. We compute each clip's event-weighted score first and
then report the unweighted mean over the 24 clips.

For readability, the per-sequence tables report each baseline in its original configuration; aggregate comparisons with the hand-masked variants are provided in Tables~1 and~2 of the main paper. All tables use the same held-out split, resolution, masks, and metric definitions as the corresponding aggregate comparisons. Topo-4DGS obtains the best mean PSNR, SSIM, and LPIPS in all four evaluations. Relative to the strongest original baseline, its mean PSNR gain is $0.80$\,dB for the full scene, $1.19$\,dB for the topology-changing object, $1.93$\,dB for Other Dyn.\ Objects, and $4.99$\,dB after the event. The larger post-event margin indicates that the improvement is concentrated in frames requiring the changed motion structure and surface support.

We also report per-sequence results for the Rigid Objects Control. The control columns in Table~1 of the main paper evaluate the ground-truth region of the manipulated rigid object, and Table~\ref{tab:noevent_primary} provides the corresponding per-sequence values. Topo-4DGS obtains the best mean object-region PSNR, SSIM, and LPIPS.

Table~\ref{tab:noevent_full_scene} additionally evaluates the complete rendered scene. Under this broader protocol, Def-3DGS obtains the best mean full-scene PSNR, SSIM, and LPIPS, while DeGauss also obtains lower LPIPS than Topo-4DGS and nearly identical SSIM. The object-region results show that the topology-aware representation does not impose a systematic penalty when no structural transition occurs.

\begin{table*}[t]
\caption{\textbf{Per-sequence full-scene reconstruction on TopoEgo.} Full-frame PSNR, SSIM, and LPIPS are evaluated on held-out frames. Hand pixels are zeroed in prediction and ground truth only during scoring. PSNR/SSIM $\uparrow$, LPIPS $\downarrow$.}
\label{tab:full_scene}
\centering
\footnotesize
\setlength{\tabcolsep}{4pt}
\renewcommand{\arraystretch}{1.1}
\begin{tabular}{l l ccc ccc ccc ccc ccc}
\toprule
& & \multicolumn{3}{c}{\textbf{Topo-4DGS [Ours]}} & \multicolumn{3}{c}{EgoGaussian~\cite{zhang2024egogaussian}} & \multicolumn{3}{c}{4DGS~\cite{wu2024fourdgs}} & \multicolumn{3}{c}{Def-3DGS\cite{yang2024deformable3dgs}} & \multicolumn{3}{c}{DeGauss~\cite{wang2025degauss}} \\
Clip & Type & PSNR & SSIM & LPIPS & PSNR & SSIM & LPIPS & PSNR & SSIM & LPIPS & PSNR & SSIM & LPIPS & PSNR & SSIM & LPIPS \\
\midrule
\multicolumn{17}{l}{\itshape \textbf{EPIC-KITCHENS}} \\

P02\_108 & \textit{fusion} & \bestcell{33.27} & \bestcell{0.964} & \bestcell{0.045} & 29.05 & 0.748 & \secondcell{0.119} & 31.19 & 0.807 & 0.196 & \secondcell{32.34} & \secondcell{0.893} & \thirdcell{0.125} & \thirdcell{31.70} & \thirdcell{0.860} & 0.139 \\
P02\_122 & \textit{fission} & \bestcell{32.67} & \bestcell{0.961} & \bestcell{0.057} & 28.76 & 0.759 & \secondcell{0.136} & 29.15 & 0.753 & 0.280 & \thirdcell{30.83} & \thirdcell{0.843} & 0.169 & \secondcell{31.24} & \secondcell{0.865} & \thirdcell{0.160} \\
P04\_04 & \textit{fission} & \bestcell{33.85} & \bestcell{0.964} & \bestcell{0.086} & 22.02 & 0.641 & \thirdcell{0.094} & \thirdcell{32.17} & \thirdcell{0.939} & 0.097 & \secondcell{32.19} & \secondcell{0.945} & \secondcell{0.090} & 30.29 & 0.915 & 0.127 \\
P04\_118 & \textit{close} & \bestcell{26.08} & \bestcell{0.889} & \bestcell{0.302} & 20.91 & 0.696 & \secondcell{0.447} & 22.94 & 0.736 & 0.531 & \secondcell{23.96} & \secondcell{0.821} & \thirdcell{0.458} & \thirdcell{23.39} & \thirdcell{0.787} & 0.479 \\
P06\_110 & \textit{fission} & \secondcell{26.84} & \secondcell{0.838} & \bestcell{0.161} & 26.51 & 0.829 & 0.164 & 26.63 & 0.833 & 0.168 & \bestcell{26.88} & \thirdcell{0.836} & \thirdcell{0.163} & \thirdcell{26.79} & \bestcell{0.841} & \secondcell{0.162} \\
P07\_116 & \textit{open} & \bestcell{28.71} & \bestcell{0.942} & \bestcell{0.119} & 23.41 & \secondcell{0.918} & \secondcell{0.146} & 25.85 & 0.769 & 0.306 & \secondcell{27.24} & \thirdcell{0.879} & \thirdcell{0.184} & \thirdcell{26.15} & 0.845 & 0.193 \\
P12\_103 & \textit{open} & \bestcell{29.04} & \bestcell{0.922} & \bestcell{0.116} & 24.40 & 0.716 & \secondcell{0.218} & 26.35 & 0.775 & 0.335 & \secondcell{27.45} & \thirdcell{0.846} & \thirdcell{0.230} & \thirdcell{27.15} & \secondcell{0.855} & 0.238 \\
P22\_14 & \textit{fusion} & \secondcell{25.51} & \secondcell{0.809} & \bestcell{0.137} & 22.35 & 0.666 & \secondcell{0.312} & 24.35 & 0.717 & 0.449 & \bestcell{25.70} & \bestcell{0.823} & \thirdcell{0.325} & \thirdcell{24.79} & \thirdcell{0.768} & 0.399 \\
P28\_13 & \textit{close} & \bestcell{33.72} & \bestcell{0.963} & \bestcell{0.107} & 29.37 & 0.741 & \thirdcell{0.337} & 31.51 & 0.812 & 0.395 & \secondcell{32.48} & \secondcell{0.874} & \secondcell{0.326} & \thirdcell{32.01} & \thirdcell{0.857} & 0.352 \\
P29\_04 & \textit{fusion} & \thirdcell{20.63} & \secondcell{0.744} & \secondcell{0.488} & 19.01 & \thirdcell{0.703} & \thirdcell{0.516} & \secondcell{21.09} & 0.673 & 0.524 & \bestcell{21.91} & \bestcell{0.783} & \bestcell{0.455} & 20.01 & 0.663 & 0.555 \\
P30\_02 & \textit{close} & \thirdcell{19.57} & \thirdcell{0.636} & \bestcell{0.379} & 16.20 & 0.500 & \thirdcell{0.392} & 17.35 & 0.545 & 0.445 & \bestcell{20.10} & \secondcell{0.655} & 0.397 & \secondcell{19.86} & \bestcell{0.663} & \secondcell{0.389} \\
P30\_110 & \textit{open} & \bestcell{28.76} & \bestcell{0.929} & \bestcell{0.134} & 24.16 & 0.711 & \secondcell{0.303} & 26.10 & 0.773 & 0.400 & \secondcell{27.02} & \secondcell{0.844} & \thirdcell{0.315} & \thirdcell{26.57} & \thirdcell{0.823} & 0.350 \\

\midrule
\multicolumn{17}{l}{\itshape \textbf{Ego4D}} \\

04994ce8 & \textit{fission} & \thirdcell{26.89} & \thirdcell{0.901} & \thirdcell{0.208} & 26.42 & 0.887 & 0.213 & 26.61 & 0.893 & 0.217 & \bestcell{27.89} & \bestcell{0.951} & \bestcell{0.158} & \secondcell{27.39} & \secondcell{0.926} & \secondcell{0.183} \\
200a8 & \textit{close} & \secondcell{27.30} & \thirdcell{0.937} & \thirdcell{0.098} & 25.14 & 0.917 & 0.112 & 24.95 & 0.805 & 0.257 & \bestcell{27.58} & \secondcell{0.939} & \secondcell{0.095} & \thirdcell{27.15} & \bestcell{0.943} & \bestcell{0.092} \\
2d29b45a & \textit{fission} & \bestcell{29.58} & \bestcell{0.891} & \bestcell{0.188} & 25.15 & 0.718 & \secondcell{0.279} & 27.25 & 0.735 & 0.395 & \secondcell{28.18} & \thirdcell{0.830} & 0.301 & \thirdcell{27.85} & \secondcell{0.842} & \thirdcell{0.290} \\
477feff8 & \textit{fusion} & \bestcell{30.40} & \bestcell{0.946} & \bestcell{0.085} & 25.90 & 0.770 & \secondcell{0.168} & 27.70 & 0.792 & 0.290 & \secondcell{28.95} & \thirdcell{0.880} & 0.190 & \thirdcell{28.55} & \secondcell{0.892} & \thirdcell{0.175} \\
47bb1fd4 & \textit{open} & \secondcell{29.04} & \secondcell{0.923} & \secondcell{0.109} & 24.70 & 0.750 & 0.118 & 26.60 & 0.775 & 0.300 & \bestcell{29.10} & \thirdcell{0.920} & \thirdcell{0.112} & \thirdcell{28.98} & \bestcell{0.925} & \bestcell{0.107} \\
672c902a & \textit{close} & \bestcell{25.19} & \bestcell{0.838} & \bestcell{0.170} & 20.90 & 0.660 & \secondcell{0.260} & 22.80 & 0.690 & 0.370 & \secondcell{23.85} & \thirdcell{0.775} & 0.285 & \thirdcell{23.55} & \secondcell{0.785} & \thirdcell{0.270} \\
68b90b42 & \textit{fission} & \secondcell{32.71} & \secondcell{0.961} & \secondcell{0.085} & 28.15 & 0.780 & \thirdcell{0.087} & 30.15 & 0.805 & 0.280 & \bestcell{32.96} & \thirdcell{0.958} & 0.090 & \thirdcell{32.60} & \bestcell{0.965} & \bestcell{0.082} \\
6c03be74 & \textit{fusion} & \bestcell{28.37} & \bestcell{0.916} & \bestcell{0.313} & 23.95 & 0.745 & \secondcell{0.397} & 25.95 & 0.765 & 0.510 & \secondcell{27.00} & \thirdcell{0.850} & 0.425 & \thirdcell{26.70} & \secondcell{0.860} & \thirdcell{0.410} \\
6c56ee6e & \textit{close} & \secondcell{25.90} & \secondcell{0.878} & \secondcell{0.145} & 21.70 & 0.706 & \thirdcell{0.147} & 23.35 & 0.725 & 0.350 & \bestcell{25.96} & \thirdcell{0.875} & 0.148 & \thirdcell{25.72} & \bestcell{0.881} & \bestcell{0.143} \\
6c56ee6e & \textit{open} & \bestcell{15.16} & \bestcell{0.595} & \bestcell{0.489} & 10.90 & 0.420 & \secondcell{0.568} & 12.55 & 0.445 & 0.690 & \secondcell{13.75} & \thirdcell{0.530} & 0.600 & \thirdcell{13.40} & \secondcell{0.545} & \thirdcell{0.580} \\
8c335aa9 & \textit{fusion} & \bestcell{33.14} & \bestcell{0.954} & \bestcell{0.145} & 28.55 & 0.770 & \secondcell{0.226} & 30.35 & 0.798 & 0.355 & \secondcell{31.00} & \thirdcell{0.884} & 0.250 & \thirdcell{30.72} & \secondcell{0.895} & \thirdcell{0.235} \\
b4707bc6 & \textit{open} & \bestcell{28.55} & \bestcell{0.931} & \bestcell{0.140} & 24.00 & 0.755 & \secondcell{0.222} & 26.10 & 0.780 & 0.330 & \secondcell{27.20} & \thirdcell{0.865} & 0.245 & \thirdcell{26.90} & \secondcell{0.875} & \thirdcell{0.230} \\

\midrule
\textbf{Mean} &  & \bestcell{27.95} & \bestcell{0.885} & \bestcell{0.179} & 23.82 & 0.729 & \secondcell{0.249} & 25.79 & 0.756 & 0.353 & \secondcell{27.15} & \secondcell{0.846} & \thirdcell{0.256} & \thirdcell{26.64} & \thirdcell{0.837} & 0.264 \\
\bottomrule
\end{tabular}

\end{table*}

\begin{table*}[t]
\caption{\textbf{Per-sequence topology-changing-object reconstruction on TopoEgo.} PSNR is computed over ground-truth mask pixels, SSIM over the masked pixels of a padded image crop, and LPIPS on the corresponding bounding-box crop. Hand pixels are zeroed in prediction and ground truth only during scoring. Results use held-out evaluation frames.}
\label{tab:primary}
\centering
\footnotesize
\setlength{\tabcolsep}{3pt}
\renewcommand{\arraystretch}{1.1}
\resizebox{\textwidth}{!}{%
\begin{tabular}{l l ccc ccc ccc ccc ccc}
\toprule
& & \multicolumn{3}{c}{\textbf{Topo-4DGS [Ours]}} & \multicolumn{3}{c}{EgoGaussian~\cite{zhang2024egogaussian}} & \multicolumn{3}{c}{4DGS~\cite{wu2024fourdgs}} & \multicolumn{3}{c}{Def-3DGS~\cite{yang2024deformable3dgs}} & \multicolumn{3}{c}{DeGauss~\cite{wang2025degauss}} \\
Clip & Type & PSNR & SSIM & LPIPS & PSNR & SSIM & LPIPS & PSNR & SSIM & LPIPS & PSNR & SSIM & LPIPS & PSNR & SSIM & LPIPS \\
\midrule
\multicolumn{17}{l}{\itshape \textbf{EPIC-KITCHENS}} \\

P02\_108 & \textit{fusion} & \bestcell{28.17} & \bestcell{0.866} & \bestcell{0.075} & 25.85 & 0.764 & \secondcell{0.077} & 26.00 & 0.752 & 0.083 & \secondcell{27.12} & \secondcell{0.828} & \thirdcell{0.078} & \thirdcell{26.65} & \thirdcell{0.796} & 0.080 \\
P02\_122 & \textit{fission} & \bestcell{20.91} & \bestcell{0.769} & \bestcell{0.113} & 19.10 & 0.681 & \secondcell{0.123} & 18.85 & 0.668 & 0.149 & \secondcell{20.02} & \secondcell{0.737} & \thirdcell{0.128} & \thirdcell{19.55} & \thirdcell{0.708} & 0.136 \\
P04\_04 & \textit{fission} & \bestcell{25.78} & \bestcell{0.862} & \bestcell{0.219} & 18.50 & 0.560 & \secondcell{0.280} & 18.20 & 0.550 & 0.410 & \secondcell{21.20} & \secondcell{0.760} & \thirdcell{0.292} & \thirdcell{20.20} & \thirdcell{0.680} & 0.335 \\
P04\_118 & \textit{close} & \bestcell{21.96} & \bestcell{0.701} & \bestcell{0.301} & 19.20 & 0.611 & \secondcell{0.418} & 19.35 & 0.620 & 0.451 & \secondcell{20.18} & \secondcell{0.665} & \thirdcell{0.429} & \thirdcell{19.82} & \thirdcell{0.640} & 0.440 \\
P06\_110 & \textit{fission} & \bestcell{24.82} & \bestcell{0.844} & \bestcell{0.110} & 23.85 & 0.811 & \secondcell{0.114} & 24.02 & 0.818 & 0.132 & \secondcell{24.56} & \thirdcell{0.838} & 0.117 & \thirdcell{24.43} & \secondcell{0.840} & \thirdcell{0.115} \\
P07\_116 & \textit{open} & \bestcell{20.76} & \bestcell{0.631} & \bestcell{0.209} & \secondcell{20.12} & \thirdcell{0.592} & \secondcell{0.216} & 18.85 & 0.548 & 0.232 & \thirdcell{19.70} & \secondcell{0.594} & \thirdcell{0.220} & 19.55 & 0.575 & 0.224 \\
P12\_103 & \textit{open} & \bestcell{19.66} & \bestcell{0.614} & \bestcell{0.173} & 17.45 & 0.520 & \thirdcell{0.194} & 17.25 & 0.505 & 0.260 & \secondcell{18.75} & \thirdcell{0.590} & \secondcell{0.191} & \thirdcell{18.45} & \secondcell{0.596} & 0.197 \\
P22\_14 & \textit{fusion} & \bestcell{26.65} & \bestcell{0.812} & \bestcell{0.125} & 23.15 & 0.716 & \secondcell{0.244} & 22.95 & 0.704 & 0.271 & \secondcell{24.25} & \secondcell{0.774} & \thirdcell{0.255} & \thirdcell{23.75} & \thirdcell{0.744} & 0.262 \\
P28\_13 & \textit{close} & \bestcell{29.29} & \bestcell{0.840} & \bestcell{0.285} & 27.15 & 0.744 & \thirdcell{0.299} & 26.95 & 0.731 & 0.315 & \secondcell{28.24} & \secondcell{0.801} & \secondcell{0.297} & \thirdcell{27.85} & \thirdcell{0.774} & 0.303 \\
P29\_04 & \textit{fusion} & \bestcell{27.16} & \bestcell{0.805} & \bestcell{0.274} & \thirdcell{26.50} & \thirdcell{0.780} & \secondcell{0.277} & 25.95 & 0.755 & 0.289 & \secondcell{26.72} & \secondcell{0.790} & \thirdcell{0.279} & 26.40 & 0.775 & 0.283 \\
P30\_02 & \textit{close} & \secondcell{22.19} & \bestcell{0.668} & \bestcell{0.343} & 20.25 & 0.585 & \secondcell{0.347} & 20.55 & 0.600 & 0.410 & \bestcell{22.31} & \thirdcell{0.657} & 0.352 & \thirdcell{22.05} & \secondcell{0.662} & \thirdcell{0.349} \\
P30\_110 & \textit{open} & \bestcell{20.51} & \bestcell{0.605} & \bestcell{0.244} & 17.40 & 0.466 & \secondcell{0.419} & 17.15 & 0.451 & 0.471 & \secondcell{18.34} & \secondcell{0.520} & \thirdcell{0.433} & \thirdcell{17.75} & \thirdcell{0.492} & 0.450 \\

\midrule
\multicolumn{17}{l}{\itshape \textbf{Ego4D}} \\

04994ce8 & \textit{fission} & \bestcell{25.67} & \bestcell{0.861} & \bestcell{0.274} & 24.46 & 0.822 & \secondcell{0.281} & 24.71 & 0.814 & 0.306 & \secondcell{25.41} & \thirdcell{0.851} & \thirdcell{0.284} & \thirdcell{25.17} & \secondcell{0.855} & 0.291 \\
200a8 & \textit{close} & \secondcell{21.49} & \thirdcell{0.758} & \thirdcell{0.228} & 19.71 & 0.706 & 0.235 & 18.45 & 0.624 & 0.315 & \bestcell{21.65} & \secondcell{0.762} & \secondcell{0.224} & \thirdcell{21.35} & \bestcell{0.764} & \bestcell{0.222} \\
2d29b45a & \textit{fission} & \bestcell{28.86} & \bestcell{0.852} & \bestcell{0.215} & 25.73 & 0.732 & \secondcell{0.253} & 25.27 & 0.713 & 0.339 & \secondcell{27.42} & \thirdcell{0.810} & 0.274 & \thirdcell{26.96} & \secondcell{0.821} & \thirdcell{0.261} \\
477feff8 & \textit{fusion} & \bestcell{20.48} & \bestcell{0.746} & \bestcell{0.187} & 17.45 & 0.628 & \secondcell{0.228} & 17.13 & 0.608 & 0.314 & \secondcell{19.01} & \thirdcell{0.699} & 0.249 & \thirdcell{18.69} & \secondcell{0.710} & \thirdcell{0.236} \\
47bb1fd4 & \textit{open} & \bestcell{25.24} & \bestcell{0.781} & \bestcell{0.128} & 22.13 & 0.663 & \secondcell{0.131} & 21.81 & 0.650 & 0.250 & \thirdcell{24.86} & \secondcell{0.769} & 0.136 & \secondcell{24.95} & \thirdcell{0.765} & \thirdcell{0.133} \\
672c902a & \textit{close} & \bestcell{22.89} & \bestcell{0.718} & \bestcell{0.200} & 19.98 & 0.604 & \secondcell{0.244} & 19.59 & 0.584 & 0.323 & \secondcell{21.41} & \thirdcell{0.669} & 0.265 & \thirdcell{21.08} & \secondcell{0.679} & \thirdcell{0.252} \\
68b90b42 & \textit{fission} & \bestcell{21.46} & \bestcell{0.762} & \bestcell{0.116} & 18.72 & 0.649 & \secondcell{0.120} & 18.39 & 0.629 & 0.232 & \secondcell{21.10} & \thirdcell{0.749} & 0.125 & \thirdcell{20.97} & \secondcell{0.754} & \thirdcell{0.122} \\
6c03be74& \textit{fusion} & \secondcell{17.18} & \secondcell{0.489} & \secondcell{0.164} & 14.41 & 0.380 & \thirdcell{0.166} & 14.02 & 0.360 & 0.282 & \bestcell{17.21} & \thirdcell{0.485} & 0.167 & \thirdcell{17.01} & \bestcell{0.491} & \bestcell{0.163} \\
6c56ee6e & \textit{close} & \bestcell{22.31} & \bestcell{0.608} & \bestcell{0.344} & 19.37 & 0.494 & \secondcell{0.389} & 19.05 & 0.474 & 0.469 & \secondcell{20.87} & \thirdcell{0.559} & 0.410 & \thirdcell{20.54} & \secondcell{0.569} & \thirdcell{0.397} \\
6c56ee6e & \textit{open} & \bestcell{13.40} & \bestcell{0.416} & \bestcell{0.494} & 10.35 & 0.304 & \secondcell{0.535} & 10.08 & 0.291 & 0.619 & \secondcell{11.84} & \thirdcell{0.369} & 0.560 & \thirdcell{11.58} & \secondcell{0.380} & \thirdcell{0.547} \\
8c335aa9 & \textit{fusion} & \bestcell{33.80} & \bestcell{0.920} & \bestcell{0.137} & 30.42 & 0.790 & \secondcell{0.185} & 30.09 & 0.771 & 0.271 & \secondcell{31.39} & \thirdcell{0.858} & 0.206 & \thirdcell{31.00} & \secondcell{0.868} & \thirdcell{0.193} \\
b4707bc6 & \textit{open} & \bestcell{23.76} & \bestcell{0.839} & \bestcell{0.119} & 20.63 & 0.721 & \secondcell{0.158} & 20.30 & 0.704 & 0.244 & \secondcell{22.32} & \thirdcell{0.794} & 0.179 & \thirdcell{21.99} & \secondcell{0.804} & \thirdcell{0.166} \\

\midrule
\textbf{Mean} &  & \bestcell{23.52} & \bestcell{0.740} & \bestcell{0.212} & 20.91 & 0.638 & \secondcell{0.247} & 20.62 & 0.622 & 0.310 & \secondcell{22.33} & \secondcell{0.705} & \thirdcell{0.256} & \thirdcell{21.99} & \thirdcell{0.698} & 0.257 \\
\bottomrule
\end{tabular}
}%

\end{table*}

\begin{table*}[t]
\caption{\textbf{Per-sequence other-dynamic-object reconstruction on TopoEgo.}
For each Other Dyn.\ Object, PSNR is computed over its ground-truth mask
pixels, SSIM over the masked pixels of its padded image crop, and LPIPS on its
corresponding bounding-box crop. Hand pixels are zeroed in both prediction and
ground truth only during scoring. Object scores are averaged within each
eligible clip.}
\label{tab:secondary}
\centering
\footnotesize
\setlength{\tabcolsep}{3pt}
\renewcommand{\arraystretch}{1.1}
\resizebox{\textwidth}{!}{%
\begin{tabular}{l l l ccc ccc ccc ccc ccc}
\toprule
 &  &  & \multicolumn{3}{c}{\textbf{Topo-4DGS [Ours]}} & \multicolumn{3}{c}{EgoGaussian~\cite{zhang2024egogaussian}} & \multicolumn{3}{c}{4DGS~\cite{wu2024fourdgs}} & \multicolumn{3}{c}{Def-3DGS~\cite{yang2024deformable3dgs}} & \multicolumn{3}{c}{DeGauss~\cite{wang2025degauss}} \\
Clip & Type & Object & PSNR & SSIM & LPIPS & PSNR & SSIM & LPIPS & PSNR & SSIM & LPIPS & PSNR & SSIM & LPIPS & PSNR & SSIM & LPIPS \\
\midrule
\multicolumn{18}{l}{\itshape \textbf{EPIC-KITCHENS}} \\

P02\_122 & \textit{fission} & compost\_bin & \bestcell{36.58} & \bestcell{0.945} & \bestcell{0.074} & 29.20 & 0.820 & 0.145 & 29.75 & 0.842 & 0.132 & \secondcell{32.95} & \secondcell{0.905} & \secondcell{0.092} & \thirdcell{31.90} & \thirdcell{0.885} & \thirdcell{0.105} \\
P04\_04 & \textit{fission} & knife & \bestcell{27.82} & \bestcell{0.834} & \bestcell{0.193} & 17.10 & 0.340 & 0.430 & 18.05 & 0.455 & 0.380 & \secondcell{21.45} & \secondcell{0.676} & \secondcell{0.285} & \thirdcell{21.20} & \thirdcell{0.661} & \thirdcell{0.295} \\
P04\_118 & \textit{close} & cutting\_board & \bestcell{24.20} & \thirdcell{0.798} & \bestcell{0.257} & 19.85 & 0.662 & 0.477 & 20.55 & 0.690 & 0.464 & \secondcell{24.08} & \bestcell{0.842} & \secondcell{0.400} & \thirdcell{23.82} & \secondcell{0.834} & \thirdcell{0.403} \\
P06\_110 & \textit{fission} & knife & \bestcell{17.83} & \bestcell{0.705} & \bestcell{0.151} & 15.85 & 0.595 & 0.250 & 16.15 & 0.625 & 0.225 & \secondcell{17.05} & \thirdcell{0.668} & \thirdcell{0.178} & \thirdcell{16.88} & \secondcell{0.690} & \secondcell{0.171} \\
P28\_13 & \textit{close} & egg & \secondcell{30.20} & \secondcell{0.872} & \secondcell{0.232} & 26.10 & 0.675 & 0.274 & 26.55 & 0.704 & 0.268 & \thirdcell{29.92} & \thirdcell{0.865} & \thirdcell{0.234} & \bestcell{30.28} & \bestcell{0.878} & \bestcell{0.231} \\

\multicolumn{18}{l}{\itshape \textbf{Ego4D}} \\

04994ce8 & \textit{fission} & quartz\_cutter & \secondcell{24.28} & \secondcell{0.806} & \secondcell{0.201} & 22.84 & 0.748 & 0.263 & 23.12 & 0.761 & 0.251 & \bestcell{24.76} & \bestcell{0.827} & \bestcell{0.187} & \thirdcell{24.03} & \thirdcell{0.797} & \thirdcell{0.209} \\
200a8 & \textit{close} & broom & \bestcell{28.06} & \bestcell{0.947} & \bestcell{0.041} & 21.00 & 0.720 & 0.110 & 21.60 & 0.750 & 0.095 & \secondcell{24.40} & \thirdcell{0.860} & \thirdcell{0.070} & \thirdcell{24.05} & \secondcell{0.885} & \secondcell{0.058} \\
2d29b45a & \textit{fission} & knife & \bestcell{20.62} & \bestcell{0.557} & \bestcell{0.243} & 15.60 & 0.405 & 0.345 & 16.10 & 0.430 & 0.330 & \secondcell{19.35} & \secondcell{0.545} & \secondcell{0.260} & \thirdcell{19.05} & \thirdcell{0.538} & \thirdcell{0.265} \\
672c902a & \textit{close} & wooden\_spatula & \bestcell{21.22} & \bestcell{0.557} & \bestcell{0.224} & 16.15 & 0.410 & 0.325 & 16.70 & 0.435 & 0.310 & \secondcell{20.15} & \secondcell{0.548} & \secondcell{0.240} & \thirdcell{19.90} & \thirdcell{0.542} & \thirdcell{0.245} \\
68b90b42 & \textit{fission} & egg1 & \bestcell{25.59} & \bestcell{0.820} & \bestcell{0.103} & 21.10 & 0.670 & 0.200 & 21.60 & 0.695 & 0.185 & \secondcell{25.18} & \thirdcell{0.808} & \thirdcell{0.110} & \thirdcell{25.05} & \secondcell{0.818} & \secondcell{0.105} \\
6c56ee6e & \textit{close} & container & \bestcell{23.29} & \secondcell{0.687} & \bestcell{0.418} & 18.10 & 0.530 & 0.520 & 18.65 & 0.555 & 0.505 & \secondcell{22.20} & \thirdcell{0.678} & \thirdcell{0.433} & \thirdcell{21.95} & \bestcell{0.690} & \secondcell{0.422} \\
8c335aa9 & \textit{fusion} & hammer, carpet & \bestcell{31.09} & \bestcell{0.901} & \bestcell{0.127} & 23.20 & 0.650 & 0.280 & 23.85 & 0.680 & 0.260 & \secondcell{26.15} & \thirdcell{0.800} & \thirdcell{0.190} & \thirdcell{25.70} & \secondcell{0.825} & \secondcell{0.175} \\

\midrule
\textbf{Mean} &  &  & \bestcell{25.90} & \bestcell{0.786} & \bestcell{0.189} & 20.51 & 0.602 & 0.302 & 21.06 & 0.635 & 0.284 & \secondcell{23.97} & \thirdcell{0.752} & \secondcell{0.223} & \thirdcell{23.65} & \secondcell{0.754} & \thirdcell{0.224} \\
\bottomrule
\end{tabular}
}%
\end{table*}
\begin{table*}[t]
\caption{\textbf{Per-sequence post-event reconstruction on TopoEgo.}
Object-region PSNR is computed over ground-truth mask pixels, SSIM over the
masked pixels of a padded image crop, and LPIPS on the corresponding
bounding-box crop. Hand pixels are zeroed in both prediction and ground truth
only during scoring. A frame is weighted by the number of annotated
events it is at or after, so frames following multiple events contribute to
each corresponding post-event interval; pre-event frames are excluded.}

\label{tab:primary_postevent}
\centering
\footnotesize
\setlength{\tabcolsep}{3pt}
\renewcommand{\arraystretch}{1.1}
\resizebox{\textwidth}{!}{%
\begin{tabular}{l l ccc ccc ccc ccc ccc}
\toprule
Clip & Type & \multicolumn{3}{c}{\textbf{Topo-4DGS [Ours]}} & \multicolumn{3}{c}{EgoGaussian~\cite{zhang2024egogaussian}} & \multicolumn{3}{c}{4DGS~\cite{wu2024fourdgs}} & \multicolumn{3}{c}{Def-3DGS~\cite{yang2024deformable3dgs}} & \multicolumn{3}{c}{DeGauss~\cite{wang2025degauss}} \\
 &  & PSNR & SSIM & LPIPS & PSNR & SSIM & LPIPS & PSNR & SSIM & LPIPS & PSNR & SSIM & LPIPS & PSNR & SSIM & LPIPS \\
\midrule
\multicolumn{17}{l}{\itshape \textbf{EPIC-KITCHENS}} \\

P02\_108 & \textit{fusion} & \bestcell{26.04} & \bestcell{0.809} & \bestcell{0.105} & 21.16 & 0.571 & \secondcell{0.132} & 21.70 & 0.597 & 0.160 & \secondcell{23.34} & \secondcell{0.737} & \thirdcell{0.137} & \thirdcell{22.38} & \thirdcell{0.661} & 0.144 \\
P02\_122 & \textit{fission} & \bestcell{26.94} & \bestcell{0.824} & \bestcell{0.083} & 21.14 & 0.578 & \thirdcell{0.146} & 20.75 & 0.564 & 0.171 & \secondcell{23.19} & \secondcell{0.737} & \secondcell{0.143} & \thirdcell{22.07} & \thirdcell{0.655} & 0.151 \\
P04\_04 & \textit{fission} & \bestcell{26.28} & \bestcell{0.872} & \bestcell{0.351} & 13.52 & 0.356 & \secondcell{0.388} & 13.24 & 0.426 & 0.553 & \secondcell{20.78} & \secondcell{0.580} & \thirdcell{0.417} & \thirdcell{20.20} & \thirdcell{0.535} & 0.426 \\
P04\_118 & \textit{close} & \bestcell{22.10} & \bestcell{0.718} & \bestcell{0.388} & 17.47 & 0.487 & \secondcell{0.462} & 17.57 & 0.503 & 0.546 & \secondcell{19.20} & \thirdcell{0.562} & \thirdcell{0.474} & \thirdcell{18.59} & \secondcell{0.569} & 0.477 \\
P06\_110 & \textit{fission} & \bestcell{23.34} & \bestcell{0.813} & \bestcell{0.142} & 21.95 & 0.760 & \secondcell{0.150} & 22.18 & 0.772 & 0.171 & \secondcell{22.82} & \thirdcell{0.797} & 0.157 & \thirdcell{22.66} & \secondcell{0.801} & \thirdcell{0.153} \\
P07\_116 & \textit{open}
& \bestcell{22.17} & \bestcell{0.681} & \bestcell{0.213}
& 16.79 & 0.464 & 0.300
& 16.46 & 0.421 & \thirdcell{0.293}
& \thirdcell{18.09} & \thirdcell{0.561} & \secondcell{0.253}
& \secondcell{18.76} & \secondcell{0.591} & 0.310 \\
P12\_103 & \textit{open} & \bestcell{17.52} & \bestcell{0.577} & \bestcell{0.214} & 12.05 & 0.350 & \thirdcell{0.239} & 12.30 & 0.365 & 0.330 & \secondcell{14.25} & \thirdcell{0.470} & 0.245 & \thirdcell{13.95} & \secondcell{0.485} & \secondcell{0.235} \\
P22\_14 & \textit{fusion} & \bestcell{23.43} & \bestcell{0.776} & \bestcell{0.265} & 18.90 & 0.552 & \secondcell{0.308} & 18.56 & 0.535 & 0.368 & \secondcell{21.21} & \secondcell{0.709} & \thirdcell{0.336} & \thirdcell{19.75} & \thirdcell{0.619} & 0.337 \\
P28\_13 & \textit{close} & \bestcell{28.16} & \bestcell{0.823} & \bestcell{0.292} & 23.69 & 0.604 & \thirdcell{0.355} & 23.43 & 0.586 & 0.389 & \secondcell{25.71} & \thirdcell{0.664} & \secondcell{0.353} & \thirdcell{24.57} & \secondcell{0.677} & 0.357 \\
P29\_04 & \textit{fusion} & \bestcell{26.96} & \bestcell{0.789} & \bestcell{0.286} & \secondcell{25.93} & \thirdcell{0.747} & \secondcell{0.307} & \thirdcell{25.49} & 0.711 & 0.329 & 25.06 & \secondcell{0.753} & \thirdcell{0.313} & 23.61 & 0.661 & 0.338 \\
P30\_02 & \textit{close} & \bestcell{20.13} & \bestcell{0.642} & \bestcell{0.398} & 15.80 & 0.405 & \thirdcell{0.438} & 16.10 & 0.425 & 0.525 & \secondcell{18.65} & \thirdcell{0.585} & 0.445 & \thirdcell{18.35} & \secondcell{0.600} & \secondcell{0.432} \\
P30\_110 & \textit{open} & \bestcell{19.21} & \bestcell{0.527} & \bestcell{0.395} & 14.79 & 0.308 & \thirdcell{0.506} & 14.58 & 0.301 & 0.585 & \secondcell{16.16} & \thirdcell{0.358} & \secondcell{0.504} & \thirdcell{15.41} & \secondcell{0.370} & 0.507 \\

\midrule
\multicolumn{17}{l}{\itshape \textbf{Ego4D}} \\

04994ce8 & \textit{fission} & \bestcell{22.81} & \bestcell{0.831} & \bestcell{0.275} & 21.71 & 0.774 & \thirdcell{0.289} & 21.89 & 0.785 & 0.310 & \secondcell{22.57} & \secondcell{0.820} & \secondcell{0.285} & \thirdcell{22.32} & \thirdcell{0.816} & 0.293 \\
200a8 & \textit{close} & \bestcell{22.74} & \bestcell{0.776} & \bestcell{0.234} & 14.12 & 0.543 & \secondcell{0.316} & 13.80 & 0.524 & 0.439 & \secondcell{17.52} & \thirdcell{0.681} & \thirdcell{0.331} & \thirdcell{17.40} & \secondcell{0.683} & 0.331 \\
2d29b45a & \textit{fission} & \bestcell{28.76} & \bestcell{0.766} & \bestcell{0.242} & 16.63 & 0.399 & \secondcell{0.392} & 16.31 & 0.379 & 0.481 & \secondcell{19.34} & \secondcell{0.564} & 0.441 & \thirdcell{18.32} & \thirdcell{0.561} & \thirdcell{0.406} \\
477feff8 & \textit{fusion} & \bestcell{25.92} & \bestcell{0.798} & \bestcell{0.153} & 14.49 & 0.443 & \secondcell{0.259} & 14.23 & 0.423 & 0.344 & \secondcell{16.77} & \thirdcell{0.586} & 0.286 & \thirdcell{16.18} & \secondcell{0.599} & \thirdcell{0.270} \\
47bb1fd4 & \textit{open} & \bestcell{24.71} & \bestcell{0.720} & \bestcell{0.132} & 13.67 & 0.334 & \secondcell{0.188} & 13.34 & 0.314 & 0.321 & \thirdcell{18.66} & \secondcell{0.587} & 0.200 & \secondcell{18.76} & \thirdcell{0.583} & \thirdcell{0.196} \\
672c902a & \textit{close} & \bestcell{27.17} & \bestcell{0.770} & \bestcell{0.155} & 15.51 & 0.409 & \secondcell{0.268} & 15.18 & 0.390 & 0.356 & \secondcell{17.71} & \thirdcell{0.552} & 0.295 & \thirdcell{17.13} & \secondcell{0.565} & \thirdcell{0.280} \\
68b90b42 & \textit{fission} & \bestcell{21.54} & \bestcell{0.749} & \bestcell{0.094} & 11.28 & 0.380 & \thirdcell{0.144} & 11.02 & 0.360 & 0.265 & \secondcell{16.15} & \thirdcell{0.624} & 0.146 & \thirdcell{15.99} & \secondcell{0.629} & \secondcell{0.142} \\
6c03be74 & \textit{fusion} & \bestcell{27.89} & \bestcell{0.723} & \bestcell{0.251} & 16.05 & 0.339 & \secondcell{0.402} & 15.73 & 0.319 & 0.490 & \secondcell{18.27} & \thirdcell{0.482} & 0.432 & \thirdcell{17.68} & \secondcell{0.495} & \thirdcell{0.415} \\
6c56ee6e & \textit{close} & \bestcell{20.06} & \bestcell{0.567} & \bestcell{0.139} & 10.26 & 0.126 & \secondcell{0.190} & 9.93 & 0.106 & 0.331 & \secondcell{15.46} & \thirdcell{0.378} & 0.202 & \thirdcell{15.34} & \secondcell{0.384} & \thirdcell{0.197} \\
6c56ee6e & \textit{open} & \bestcell{16.90} & \bestcell{0.581} & \bestcell{0.344} & 7.93 & 0.146 & \secondcell{0.535} & 7.60 & 0.126 & 0.626 & \secondcell{9.88} & \thirdcell{0.276} & 0.568 & \thirdcell{9.36} & \secondcell{0.289} & \thirdcell{0.551} \\
8c335aa9 & \textit{fusion} & \bestcell{38.54} & \bestcell{0.889} & \bestcell{0.147} & 25.12 & 0.623 & \secondcell{0.214} & 24.80 & 0.603 & 0.296 & \secondcell{27.33} & \thirdcell{0.766} & 0.238 & \thirdcell{26.75} & \secondcell{0.779} & \thirdcell{0.222} \\
b4707bc6 & \textit{open} & \bestcell{23.88} & \bestcell{0.787} & \bestcell{0.150} & 12.95 & 0.428 & \secondcell{0.263} & 12.69 & 0.409 & 0.345 & \secondcell{15.22} & \thirdcell{0.571} & 0.287 & \thirdcell{14.70} & \secondcell{0.584} & \thirdcell{0.271} \\

\midrule
\textbf{Mean} &  & \bestcell{24.30} & \bestcell{0.742} & \bestcell{0.227} & 16.79 & 0.464 & \secondcell{0.300} & 16.62 & 0.456 & 0.376 & \secondcell{19.31} & \secondcell{0.600} & 0.312 & \thirdcell{18.76} & \thirdcell{0.591} & \thirdcell{0.310} \\
\bottomrule

\end{tabular}
}%
\end{table*}
\begin{table*}[t]
\caption{\textbf{Full-scene reconstruction on the Rigid Objects Control.} Full-frame PSNR, SSIM, and LPIPS are evaluated on held-out frames from the six control sequences outside TopoEgo. Hand pixels are zeroed in prediction and ground truth only during scoring. PSNR/SSIM $\uparrow$, LPIPS $\downarrow$.}

\label{tab:noevent_full_scene}
\centering
\scriptsize
\setlength{\tabcolsep}{3pt}
\renewcommand{\arraystretch}{1.1}
\resizebox{\textwidth}{!}{%
\begin{tabular}{@{}l ccc ccc ccc ccc ccc@{}}
\toprule
& \multicolumn{3}{c}{\textbf{Topo-4DGS [Ours]}}
& \multicolumn{3}{c}{EgoGaussian~\cite{zhang2024egogaussian}}
& \multicolumn{3}{c}{Def-3DGS~\cite{yang2024deformable3dgs}}
& \multicolumn{3}{c}{DeGauss~\cite{wang2025degauss}}
& \multicolumn{3}{c}{4DGS~\cite{wu2024fourdgs}} \\
Clip & PSNR & SSIM & LPIPS
& PSNR & SSIM & LPIPS
& PSNR & SSIM & LPIPS
& PSNR & SSIM & LPIPS
& PSNR & SSIM & LPIPS \\
\midrule

P03\_03
& \secondcell{32.24} & \secondcell{0.939} & \secondcell{0.139}
& \bestcell{32.73} & \bestcell{0.940} & \bestcell{0.125}
& \thirdcell{31.92} & \thirdcell{0.931} & \thirdcell{0.146}
& 31.55 & 0.925 & 0.153
& 31.70 & 0.928 & 0.150 \\

P17\_01
& \bestcell{22.06} & \bestcell{0.767} & \bestcell{0.372}
& 18.67 & 0.685 & 0.431
& \secondcell{21.74} & \secondcell{0.752} & \secondcell{0.384}
& \thirdcell{21.48} & \thirdcell{0.743} & \thirdcell{0.392}
& 21.31 & 0.738 & 0.398 \\

35fa3
& \thirdcell{30.11} & 0.938 & 0.176
& \bestcell{31.53} & \bestcell{0.946} & \thirdcell{0.169}
& 29.86 & \thirdcell{0.941} & \secondcell{0.099}
& \secondcell{30.71} & \bestcell{0.946} & \bestcell{0.097}
& 22.49 & 0.755 & 0.316 \\

P18\_06
& \bestcell{38.21} & \bestcell{0.976} & \secondcell{0.107}
& \secondcell{37.58} & \secondcell{0.975} & \bestcell{0.106}
& \thirdcell{35.30} & \thirdcell{0.973} & \thirdcell{0.133}
& 33.63 & 0.963 & 0.144
& 32.37 & 0.916 & 0.284 \\

P32\_01
& 29.57 & 0.890 & 0.197
& \thirdcell{30.49} & \thirdcell{0.922} & 0.172
& \secondcell{32.33} & \bestcell{0.947} & \bestcell{0.132}
& 29.30 & \thirdcell{0.922} & \thirdcell{0.163}
& \bestcell{33.92} & \secondcell{0.944} & \secondcell{0.151} \\

ab23f
& \thirdcell{31.65} & \thirdcell{0.955} & \thirdcell{0.129}
& 23.26 & 0.812 & 0.292
& \secondcell{33.61} & \bestcell{0.960} & \secondcell{0.104}
& \bestcell{33.99} & \secondcell{0.959} & \bestcell{0.097}
& 24.19 & 0.749 & 0.303 \\

\midrule
\textbf{Mean}
& \secondcell{30.64} & \secondcell{0.911} & \thirdcell{0.187}
& 29.04 & 0.880 & 0.216
& \bestcell{30.79} & \bestcell{0.917} & \bestcell{0.166}
& \thirdcell{30.11} & \thirdcell{0.910} & \secondcell{0.174}
& 27.66 & 0.838 & 0.267 \\
\bottomrule
\end{tabular}
}%

\end{table*}
\begin{table*}[t]
\caption{\textbf{Evaluated-object reconstruction on the Rigid Objects Control.} PSNR is computed over ground-truth object-mask pixels, SSIM over the masked pixels of a padded image crop, and LPIPS on the corresponding bounding-box crop. Hand pixels are zeroed in prediction and ground truth only during scoring. Results use held-out frames from the six control sequences.}

\label{tab:noevent_primary}
\centering
\scriptsize
\setlength{\tabcolsep}{3pt}
\renewcommand{\arraystretch}{1.1}
\resizebox{\textwidth}{!}{%
\begin{tabular}{@{}l ccc ccc ccc ccc ccc@{}}
\toprule
& \multicolumn{3}{c}{\textbf{Topo-4DGS [Ours]}}
& \multicolumn{3}{c}{EgoGaussian~\cite{zhang2024egogaussian}}
& \multicolumn{3}{c}{Def-3DGS~\cite{yang2024deformable3dgs}}
& \multicolumn{3}{c}{DeGauss~\cite{wang2025degauss}}
& \multicolumn{3}{c}{4DGS~\cite{wu2024fourdgs}} \\
Clip & PSNR & SSIM & LPIPS
& PSNR & SSIM & LPIPS
& PSNR & SSIM & LPIPS
& PSNR & SSIM & LPIPS
& PSNR & SSIM & LPIPS \\
\midrule

P03\_03
& \secondcell{29.13} & \secondcell{0.900} & \secondcell{0.120}
& \bestcell{29.71} & \bestcell{0.902} & \bestcell{0.109}
& 28.31 & 0.881 & 0.139
& \thirdcell{28.82} & \thirdcell{0.892} & \thirdcell{0.128}
& 28.55 & 0.886 & 0.134 \\

P17\_01
& \bestcell{21.79} & \bestcell{0.657} & \bestcell{0.269}
& \thirdcell{21.58} & 0.583 & \thirdcell{0.285}
& 21.18 & \thirdcell{0.631} & 0.289
& \secondcell{21.42} & \secondcell{0.642} & \secondcell{0.280}
& 20.96 & 0.623 & 0.296 \\

35fa3
& \secondcell{25.36} & \secondcell{0.765} & \secondcell{0.214}
& \bestcell{26.24} & \bestcell{0.773} & \bestcell{0.188}
& \thirdcell{23.87} & \thirdcell{0.658} & 0.253
& 23.67 & 0.634 & \thirdcell{0.237}
& 20.29 & 0.486 & 0.378 \\

P18\_06
& \bestcell{33.68} & \bestcell{0.934} & \bestcell{0.079}
& \secondcell{32.72} & \secondcell{0.928} & \secondcell{0.084}
& 19.20 & 0.589 & \thirdcell{0.318}
& 19.37 & \thirdcell{0.599} & 0.320
& \thirdcell{20.63} & 0.590 & 0.369 \\

P32\_01
& 27.93 & 0.851 & \thirdcell{0.240}
& \thirdcell{28.82} & \secondcell{0.879} & \secondcell{0.218}
& 25.11 & 0.823 & 0.283
& \secondcell{29.00} & \thirdcell{0.878} & 0.241
& \bestcell{31.05} & \bestcell{0.897} & \bestcell{0.214} \\

ab23f
& \bestcell{29.65} & \bestcell{0.892} & \bestcell{0.124}
& 24.98 & 0.760 & 0.231
& \secondcell{28.54} & \secondcell{0.811} & \secondcell{0.178}
& \thirdcell{28.34} & \thirdcell{0.809} & \thirdcell{0.184}
& 21.97 & 0.572 & 0.338 \\

\midrule
\textbf{Mean}
& \bestcell{27.92} & \bestcell{0.833} & \bestcell{0.174}
& \secondcell{27.34} & \secondcell{0.804} & \secondcell{0.186}
& 24.37 & 0.732 & 0.243
& \thirdcell{25.10} & \thirdcell{0.742} & \thirdcell{0.232}
& 23.91 & 0.676 & 0.288 \\
\bottomrule
\end{tabular}
}%

\end{table*}

\section{Extended Qualitative Results}
\label{app:qualitative}

The following qualitative comparisons show the hand-masked variants of DeGauss, Def-3DGS, and 4DGS. Excluding hand pixels during their training prevents reconstructed hand geometry from obscuring differences in object structure and surface support. 

\begin{figure*}[t]
    \centering
    \footnotesize
    \renewcommand{\arraystretch}{1.02}

\newcommand{\suppqualimg}[1]{%
  \includegraphics[
    width=0.140\textwidth,
    trim=0.35pt 0 0.35pt 0,
    clip
  ]{#1}%
}

    \newcommand{\suppqualrow}[2]{%
      \rotatebox{90}{\textit{#1}} &
      \suppqualimg{figures/#2/GT_pre_highlight.png} &
      \suppqualimg{figures/#2/GT_zoom.png} &
      \suppqualimg{figures/#2/EgoGaussian_crop.png} &
      \suppqualimg{figures/#2/DeGauss_nohands_crop.png} &
      \suppqualimg{figures/#2/Def3DGS_nohands_crop.png} &
      \suppqualimg{figures/#2/4DGS_nohands_crop.png} &
      \suppqualimg{figures/#2/Topo4DGS_crop.png}%
    }

\begin{tabular}{
  @{}c@{\hspace{1pt}}
  c@{\kern-0.4pt}
  c@{\kern-0.4pt}
  c@{\kern-0.4pt}
  c@{\kern-0.4pt}
  c@{\kern-0.4pt}
  c@{\kern-0.4pt}
  c@{}
}
      &
       &
       &
       &
      DeGauss &
      Def-3DGS &
      4DGS &
      \\[0.2pt]
      
      &
      GT pre-event &
      GT post-event &
      EgoGaussian &
      (w/o hands) &
      (w/o hands) &
      (w/o hands) &
      \textbf{Topo-4DGS [ours]}
      \\[0.2pt]

      \suppqualrow{fission}{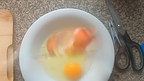}
      \\[-2.5pt]

      \suppqualrow{fusion}{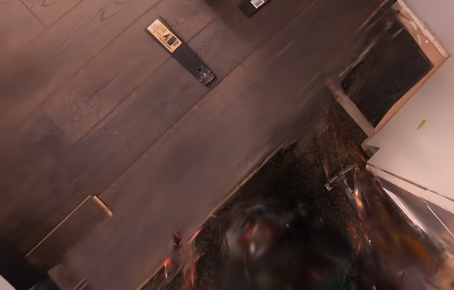}
      \\[-2.5pt]

      \suppqualrow{fusion}{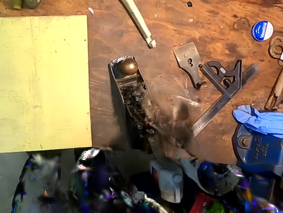}
      \\[-2.5pt]

      \suppqualrow{fusion}{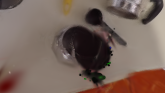}
      \\[-2.5pt]

      \suppqualrow{fusion}{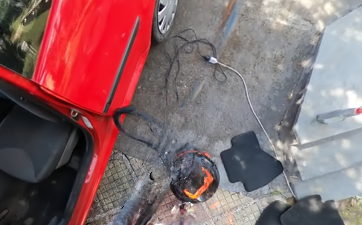}
      \\[-2.5pt]

      \suppqualrow{open}{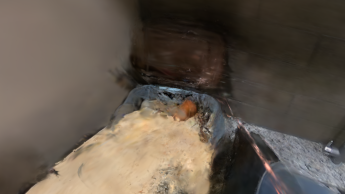}
      \\[-2.5pt]

      \suppqualrow{close}{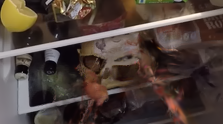}
      \\
    \end{tabular}

    \caption{\textbf{Extended qualitative comparison} on TopoEgo. Each row shows pre- and post-event ground truth followed by post-event renders from EgoGaussian, the hand-masked variants of DeGauss, Def-3DGS, and 4DGS, and Topo-4DGS. Changes are highlighted in cyan in the ground-truth frames.}
    \label{fig:supp-qualitative}
\end{figure*}

\Cref{fig:supp-qualitative} extends the qualitative comparison from the main paper (Fig.~4) with additional \textit{fission}, \textit{fusion}, \textit{open}, and \textit{close} events from both source datasets. Across these examples, Topo-4DGS more consistently preserves independently moving parts after \textit{fission} and the assembled structure after \textit{fusion}. For \textit{open} and \textit{close} events, it better represents the corresponding exposure or
concealment of interior surfaces. Fixed-topology baselines instead frequently merge separated parts, blur event regions, or omit newly visible geometry.

\begin{figure*}[t]
    \centering
    \scriptsize
    \setlength{\tabcolsep}{0.35pt}
    \renewcommand{\arraystretch}{1.02}

    \newcommand{\supprigidimg}[1]{%
      \includegraphics[width=0.161\textwidth]{#1}%
    }

    \newcommand{\supprigidrow}[2]{%
      \rotatebox{90}{\textit{#1}} &&
      \supprigidimg{figures/#2/GT_highlight.png} &
      \supprigidimg{figures/#2/EgoGaussian_crop.png} &
      \supprigidimg{figures/#2/DeGauss_nohands_crop.png} &
      \supprigidimg{figures/#2/Def3DGS_nohands_crop.png} &
      \supprigidimg{figures/#2/4DGS_nohands_crop.png} &
      \supprigidimg{figures/#2/Topo4DGS_crop.png}%
    }

    \begin{tabular}{cc c c c c c c}
      &&
       &
       &
      DeGauss &
      Def-3DGS &
      4DGS &
      \\[0.2pt]
      
      &&
      GT &
      EgoGaussian &
      (w/o hands) &
      (w/o hands) &
      (w/o hands) &
      \textbf{Topo-4DGS [ours]}
      \\[0.2pt]

      \supprigidrow{P03\_03}{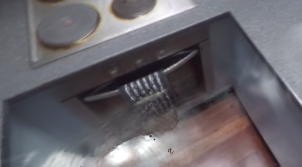}
      \\[-2.5pt]

      \supprigidrow{ab23f}{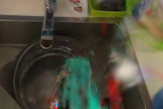}
      \\[-2.5pt]

      \supprigidrow{P18\_06}{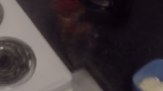}
      \\[-2.5pt]

      \supprigidrow{P32\_01}{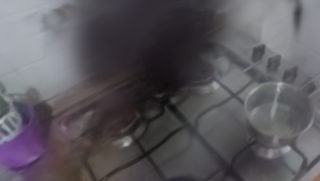}
      \\[-2.5pt]

      \supprigidrow{35fa3}{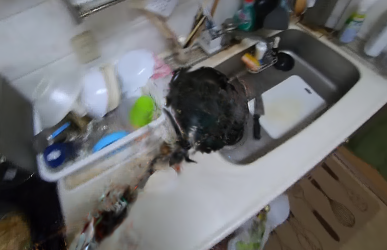}
      \\
    \end{tabular}

    \caption{\textbf{Qualitative comparison on Rigid Objects Control sequences ($K{=}1$).} The first column shows the ground-truth frame with the evaluated object outlined; the remaining columns show the corresponding object-region renders from EgoGaussian, the hand-masked variants of DeGauss, Def-3DGS, and 4DGS, and Topo-4DGS. These sequences belong to the Rigid Objects Control and contain no target topological event.}
    \label{fig:supp-rigid-controls}
\end{figure*}

\Cref{fig:supp-rigid-controls} considers fixed-topology $K{=}1$ sequences without a target topological event. Topo-4DGS preserves coherent object geometry and produces reconstructions comparable to the strongest baselines, showing that its event-conditioned components do not introduce visible artifacts when motion bifurcation or changes in surface support are not required. This observation is consistent with \cref{tab:noevent_full_scene,tab:noevent_primary}.

\section{Limitations}
\label{app:limitations}

\begin{figure*}[t]
    \centering
    \footnotesize
    \setlength{\tabcolsep}{1.0pt}
    \renewcommand{\arraystretch}{1.02}

    \newcommand{\limimg}[1]{%
      \includegraphics[width=0.235\textwidth]{#1}%
    }

    \begin{tabular}{@{}c c c c@{}}
      GT &
      \textbf{Topo-4DGS [ours]} &
      GT &
      \textbf{Topo-4DGS [ours]}
      \\[0.2pt]

      \limimg{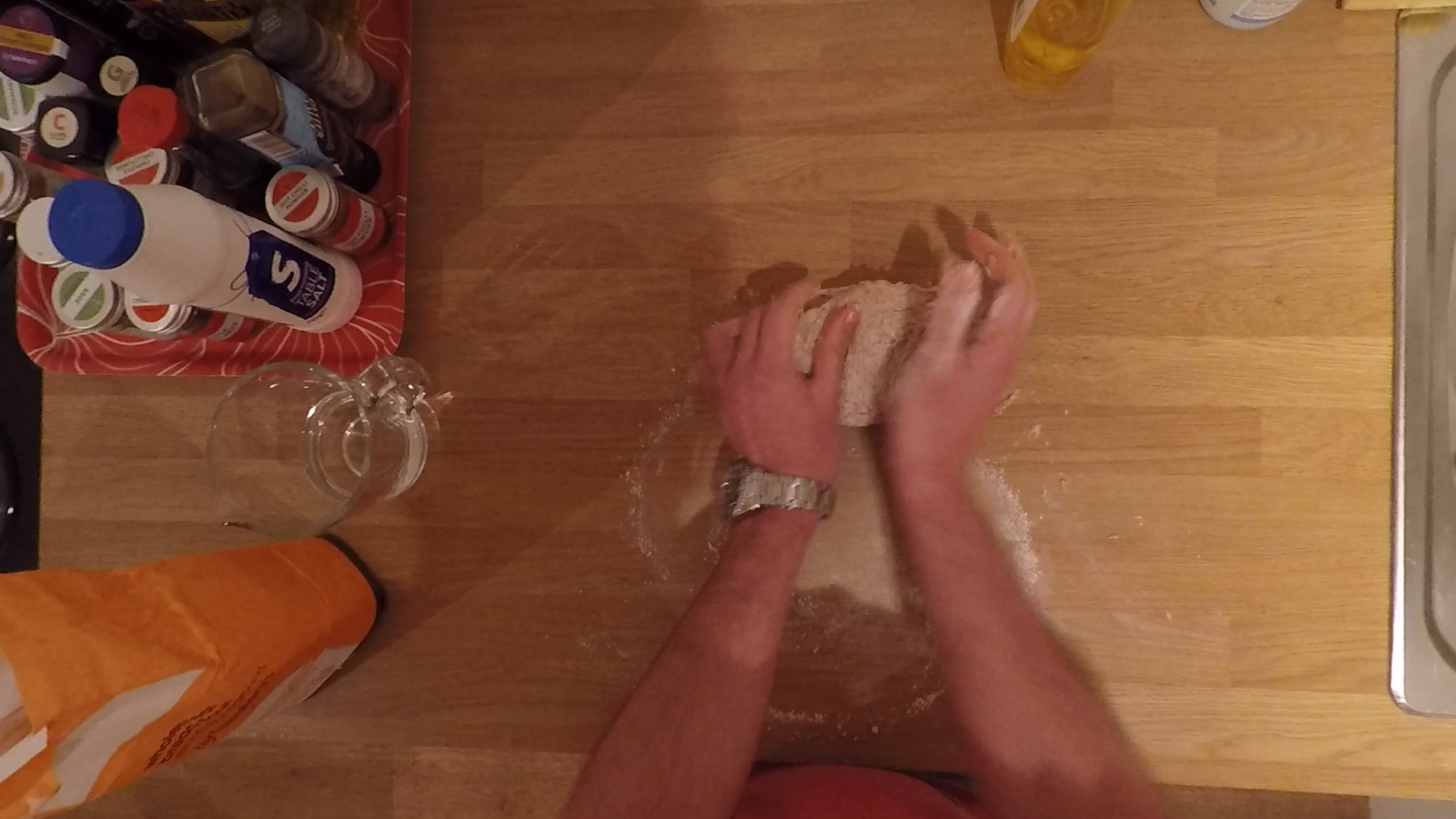} &
      \limimg{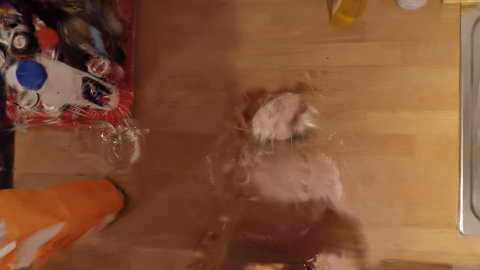} &
      \limimg{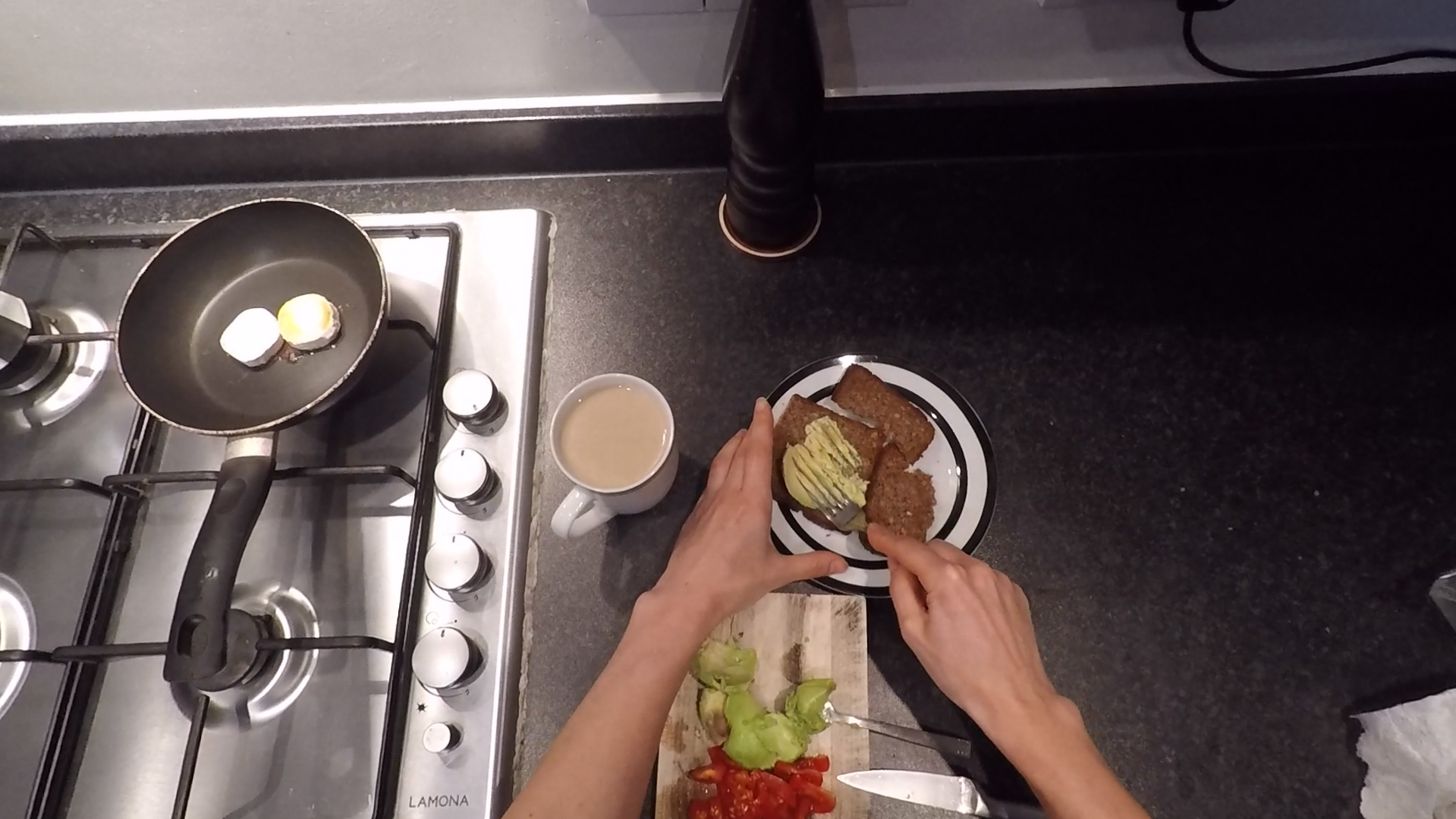} &
      \limimg{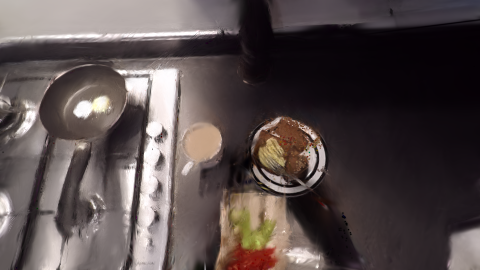}
    \end{tabular}

    \caption{\textbf{Qualitative limitations.} Left: strongly deformable dough violates the piecewise-rigid motion assumption. Right: a small, heavily occluded object part provides insufficient visual evidence for reliable reconstruction.}
    \label{fig:supp-limitations}
\end{figure*}

Topo-4DGS focuses on temporally localized topology changes whose resulting object parts can be approximated by piecewise-rigid motion. Strongly deformable interactions, such as stretching or kneading soft materials, fall outside this assumption and remain challenging (Fig.~\ref{fig:supp-limitations}). More generally, reconstruction is limited by the reduced visual evidence available in monocular egocentric video. Small parts, prolonged hand occlusions, or newly exposed surfaces that are observed only briefly can make their geometry and motion ambiguous. Challenging appearance effects, such as transparent, translucent, or strongly specular materials, can further reduce the reliability of photometric reconstruction. These difficulties are shared with related monocular dynamic and egocentric reconstruction methods, and extending the representation to more general deformation and more challenging visibility and appearance conditions is a natural direction for future work.

The current pipeline also relies on external estimates of camera intrinsics and trajectories, object and hand masks, and a specification of the topological events. Inaccurate camera calibration or pose estimation can propagate into the recovered object geometry and motion, as in other monocular reconstruction pipelines. Our benchmark experiments use manually reviewed event annotations, while Sec.~\ref{app:auto_discovery} shows that the required topology specification can also be estimated automatically from video. The automatic module is currently a preprocessing stage rather than an end-to-end component of the reconstruction system.

Finally, the present formulation targets the four discrete event families considered by TopoEgo (\textit{fission}, \textit{fusion}, \textit{open}, and \textit{close}) rather than arbitrary gradual or compound structural transformations. Broader event models and more tightly integrated automatic topology discovery provide natural extensions of the current framework.

\end{document}